\documentclass{article}

\PassOptionsToPackage{numbers, compress}{natbib}

\usepackage[preprint]{neurips_2026}

\usepackage[utf8]{inputenc} 
\usepackage[T1]{fontenc}    
\usepackage{hyperref}       
\usepackage{url}            
\usepackage{booktabs}       
\usepackage{amsfonts}       
\usepackage{nicefrac}       
\usepackage{microtype}      
\usepackage{xcolor}         

\usepackage{amsmath}
\usepackage{graphicx}
\usepackage{pifont}
\usepackage{tikz}
\usetikzlibrary{arrows.meta, calc, positioning}
\usepackage{pgf-pie}
\usepackage{multirow}
\usepackage{subcaption}
\usepackage{pgfplots}
\pgfplotsset{compat=1.17}
\usepackage{cleveref}
\usepackage{listings}
\usepgfplotslibrary{groupplots}

\definecolor{qcolor}{HTML}{2C3E50}
\definecolor{leafgreen}{HTML}{27AE60}
\definecolor{leafblue}{HTML}{2980B9}
\definecolor{yescolor}{HTML}{27AE60}
\definecolor{nocolor}{HTML}{C0392B}

\newcommand{\cmark}{\ding{51}}
\newcommand{\xmark}{\ding{55}}
\usepackage{array}
\usepackage{mathrsfs}
\usepackage{amsxtra}

\newcommand{\eat}[1]{} 

\newcommand{\td}[2]{\if*#1\else^{#1}\fi\if*#2\else_{#2}\fi} 

\newcommand\join\Join 

\DeclareSymbolFont{txsymbolsC}{U}{txsyc}{m}{n}
\SetSymbolFont{txsymbolsC}{bold}{U}{txsyc}{bx}{n}
\DeclareFontSubstitution{U}{txsyc}{m}{n}
\DeclareMathSymbol{\ljoin}{\mathrel}{txsymbolsC}{88}
\DeclareMathSymbol{\rjoin}{\mathrel}{txsymbolsC}{89}

\newcommand\sel\sigma
\newcommand\proj\pi
\newcommand\cross\times

\newcommand\LRA\Leftrightarrow

\newsavebox\setminusbox
\newlength\setminuslen

\newcolumntype{C}{>{$\displaystyle}c<{$}} 
\newcolumntype{L}{>{$\displaystyle}l<{$}} 
\newcolumntype{R}{>{$\displaystyle}r<{$}} 

\newcommand{\B}[3]{B\if*#1\else_{#1}\fi(#2,#3)} 
\newcommand{\I}[3]{I\if*#1\else_{#1}\fi(#2,#3)} 

\makeatletter
\def\imod#1{\allowbreak\mkern10mu({\operator@font mod}\,\,#1)}
\makeatother

\newlength\hspaceoflen

\newcommand\vect[1]{{\boldsymbol{#1}}}
\newcommand\va{\vect{a}}
\newcommand\vb{\vect{b}}
\newcommand\vc{\vect{c}}
\newcommand\vd{\vect{d}}
\newcommand\ve{\vect{e}}
\newcommand\vf{\vect{f}}
\newcommand\vg{\vect{g}}
\newcommand\vh{\vect{h}}
\newcommand\vi{\vect{i}}
\newcommand\vj{\vect{j}}
\newcommand\vk{\vect{k}}
\newcommand\vl{\vect{l}}
\newcommand\vm{\vect{m}}
\newcommand\vn{\vect{n}}
\newcommand\vo{\vect{o}}
\newcommand\vp{\vect{p}}
\newcommand\vq{\vect{q}}
\newcommand\vr{\vect{r}}
\newcommand\vs{\vect{s}}
\newcommand\vt{\vect{t}}
\newcommand\vu{\vect{u}}
\newcommand\vw{\vect{w}}
\newcommand\vx{\vect{x}}
\newcommand\vy{\vect{y}}
\newcommand\vz{\vect{z}}

\newcommand\mA{\vect{A}}
\newcommand\mB{\vect{B}}
\newcommand\mC{\vect{C}}
\newcommand\mD{\vect{D}}
\newcommand\mE{\vect{E}}
\newcommand\mF{\vect{F}}
\newcommand\mG{\vect{G}}
\newcommand\mH{\vect{H}}
\newcommand\mI{\vect{I}}
\newcommand\mJ{\vect{J}}
\newcommand\mK{\vect{K}}
\newcommand\mL{\vect{L}}
\newcommand\mM{\vect{M}}
\newcommand\mN{\vect{N}}
\newcommand\mO{\vect{O}}
\newcommand\mP{\vect{P}}
\newcommand\mQ{\vect{Q}}
\newcommand\mR{\vect{R}}
\newcommand\mS{\vect{S}}
\newcommand\mT{\vect{T}}
\newcommand\mU{\vect{U}}
\newcommand\mV{\vect{V}}
\newcommand\mW{\vect{W}}
\newcommand\mX{\vect{X}}
\newcommand\mY{\vect{Y}}
\newcommand\mZ{\vect{Z}}

\DeclareMathAlphabet{\mathcal}{OMS}{cmsy}{m}{n}

\newcommand\cD{\mathcal{D}}
\newcommand\cE{\mathcal{E}}

\newcommand\cG{\mathcal{G}}

\newcommand\cV{\mathcal{V}}

\accentedsymbol\Ahat{{\hat A}}
\accentedsymbol\Bhat{{\hat B}}
\accentedsymbol\Chat{{\hat C}}
\accentedsymbol\Dhat{{\hat D}}
\accentedsymbol\Ehat{{\hat E}}
\accentedsymbol\Fhat{{\hat F}}
\accentedsymbol\Ghat{{\hat G}}
\accentedsymbol\Hhat{{\hat H}}
\accentedsymbol\Ihat{{\hat I}}
\accentedsymbol\Jhat{{\hat J}}
\accentedsymbol\Khat{{\hat K}}
\accentedsymbol\Lhat{{\hat L}}
\accentedsymbol\Mhat{{\hat M}}
\accentedsymbol\Nhat{{\hat N}}
\accentedsymbol\Ohat{{\hat O}}
\accentedsymbol\Phat{{\hat P}}
\accentedsymbol\Qhat{{\hat Q}}
\accentedsymbol\Rhat{{\hat R}}
\accentedsymbol\Shat{{\hat S}}
\accentedsymbol\That{{\hat T}}
\accentedsymbol\Uhat{{\hat U}}
\accentedsymbol\Vhat{{\hat V}}
\accentedsymbol\What{{\hat W}}
\accentedsymbol\Xhat{{\hat X}}
\accentedsymbol\Yhat{{\hat Y}}
\accentedsymbol\Zhat{{\hat Z}}

\accentedsymbol\ahat{{\hat a}}
\accentedsymbol\bhat{{\hat b}}
\accentedsymbol\chat{{\hat c}}
\accentedsymbol\dhat{{\hat d}}
\accentedsymbol\ehat{{\hat e}}
\accentedsymbol\fhat{{\hat f}}
\accentedsymbol\ghat{{\hat g}}
\accentedsymbol\hhat{{\hat h}}
\accentedsymbol\ihat{{\hat i}}
\accentedsymbol\jhat{{\hat j}}
\accentedsymbol\khat{{\hat k}}
\accentedsymbol\lhat{{\hat l}}
\accentedsymbol\mhat{{\hat m}}
\accentedsymbol\nhat{{\hat n}}
\accentedsymbol\ohat{{\hat o}}
\accentedsymbol\phat{{\hat p}}
\accentedsymbol\qhat{{\hat q}}
\accentedsymbol\rhat{{\hat r}}
\accentedsymbol\shat{{\hat s}}
\accentedsymbol\that{{\hat t}}
\accentedsymbol\uhat{{\hat u}}
\accentedsymbol\vhat{{\hat v}}
\accentedsymbol\what{{\hat w}}
\accentedsymbol\xhat{{\hat x}}
\accentedsymbol\yhat{{\hat y}}
\accentedsymbol\zhat{{\hat z}}

\accentedsymbol\rhohat{{\hat\rho}}

\accentedsymbol\Abar{{\bar A}}
\accentedsymbol\Bbar{{\bar B}}
\accentedsymbol\Cbar{{\bar C}}
\accentedsymbol\Dbar{{\bar D}}
\accentedsymbol\Ebar{{\bar E}}
\accentedsymbol\Fbar{{\bar F}}
\accentedsymbol\Gbar{{\bar G}}
\accentedsymbol\Hbar{{\bar H}}
\accentedsymbol\Ibar{{\bar I}}
\accentedsymbol\Jbar{{\bar J}}
\accentedsymbol\Kbar{{\bar K}}
\accentedsymbol\Lbar{{\bar L}}
\accentedsymbol\Mbar{{\bar M}}
\accentedsymbol\Nbar{{\bar N}}
\accentedsymbol\Obar{{\bar O}}
\accentedsymbol\Pbar{{\bar P}}
\accentedsymbol\Qbar{{\bar Q}}
\accentedsymbol\Rbar{{\bar R}}
\accentedsymbol\Sbar{{\bar S}}
\accentedsymbol\Tbar{{\bar T}}
\accentedsymbol\Ubar{{\bar U}}
\accentedsymbol\Vbar{{\bar V}}
\accentedsymbol\Wbar{{\bar W}}
\accentedsymbol\Xbar{{\bar X}}
\accentedsymbol\Ybar{{\bar Y}}

\accentedsymbol\abar{{\bar a}}
\accentedsymbol\bbar{{\bar b}}
\accentedsymbol\cbar{{\bar c}}
\accentedsymbol\dbar{{\bar d}}
\accentedsymbol\ebar{{\bar e}}
\accentedsymbol\fbar{{\bar f}}
\accentedsymbol\gbar{{\bar g}}
\makeatletter
\@ifundefined{hbar}{}{
        \let\hbar\@undefined
}
\makeatother
\accentedsymbol\hbar{{\bar h}}
\accentedsymbol\ibar{{\bar i}}
\accentedsymbol\jbar{{\bar j}}
\accentedsymbol\kbar{{\bar k}}
\accentedsymbol\lbar{{\bar l}}
\accentedsymbol\mbar{{\bar m}}
\accentedsymbol\nbar{{\bar n}}
\makeatletter
\@ifundefined{obar}{}{
        \let\obar\@undefined
}
\makeatother
\accentedsymbol{\obar}{{\bar o}}
\accentedsymbol\pbar{{\bar p}}
\accentedsymbol\qbar{{\bar q}}
\accentedsymbol\rbar{{\bar r}}
\accentedsymbol\sbar{{\bar s}}
\accentedsymbol\tbar{{\bar t}}
\accentedsymbol\ubar{{\bar u}}
\accentedsymbol\vbar{{\bar v}}
\accentedsymbol\wbar{{\bar w}}
\accentedsymbol\xbar{{\bar x}}
\accentedsymbol\ybar{{\bar y}}
\accentedsymbol\zbar{{\bar z}}

\renewcommand{\epsilon}{\varepsilon}

\accentedsymbol\mAhat{{\hat\mA}}
\accentedsymbol\mBhat{{\hat\mB}}
\accentedsymbol\mChat{{\hat\mC}}
\accentedsymbol\mDhat{{\hat\mD}}
\accentedsymbol\mEhat{{\hat\mE}}
\accentedsymbol\mFhat{{\hat\mF}}
\accentedsymbol\mGhat{{\hat\mG}}
\accentedsymbol\mHhat{{\hat\mH}}
\accentedsymbol\mIhat{{\hat\mI}}
\accentedsymbol\mJhat{{\hat\mJ}}
\accentedsymbol\mKhat{{\hat\mK}}
\accentedsymbol\mLhat{{\hat\mL}}
\accentedsymbol\mMhat{{\hat\mM}}
\accentedsymbol\mNhat{{\hat\mN}}
\accentedsymbol\mOhat{{\hat\mO}}
\accentedsymbol\mPhat{{\hat\mP}}
\accentedsymbol\mQhat{{\hat\mQ}}
\accentedsymbol\mRhat{{\hat\mR}}
\accentedsymbol\mShat{{\hat\mS}}
\accentedsymbol\mThat{{\hat\mT}}
\accentedsymbol\mUhat{{\hat\mU}}
\accentedsymbol\mVhat{{\hat\mV}}
\accentedsymbol\mWhat{{\hat\mW}}
\accentedsymbol\mXhat{{\hat\mX}}
\accentedsymbol\mYhat{{\hat\mY}}
\accentedsymbol\mZhat{{\hat\mZ}}

\accentedsymbol\vahat{{\hat\va}}
\accentedsymbol\vbhat{{\hat\vb}}
\accentedsymbol\vchat{{\hat\vc}}
\accentedsymbol\vdhat{{\hat\vd}}
\accentedsymbol\vehat{{\hat\ve}}
\accentedsymbol\vfhat{{\hat\vf}}
\accentedsymbol\vghat{{\hat\vg}}
\accentedsymbol\vhhat{{\hat\vh}}
\accentedsymbol\vihat{{\hat\vi}}
\accentedsymbol\vjhat{{\hat\vj}}
\accentedsymbol\vkhat{{\hat\vk}}
\accentedsymbol\vlhat{{\hat\vl}}
\accentedsymbol\vmhat{{\hat\vm}}
\accentedsymbol\vnhat{{\hat\vn}}
\accentedsymbol\vohat{{\hat\vo}}
\accentedsymbol\vphat{{\hat\vp}}
\accentedsymbol\vqhat{{\hat\vq}}
\accentedsymbol\vrhat{{\hat\vr}}
\accentedsymbol\vshat{{\hat\vs}}
\accentedsymbol\vthat{{\hat\vt}}
\accentedsymbol\vuhat{{\hat\vu}}
\accentedsymbol\vwhat{{\hat\vw}}
\accentedsymbol\vxhat{{\hat\vx}}
\accentedsymbol\vyhat{{\hat\vy}}
\accentedsymbol\vzhat{{\hat\vz}}

\newcounter{kbNOC}
\newcommand{\kb}[1]{\textcolor{blue}{\small  [KB\#\arabic{kbNOC}\stepcounter{kbNOC}: #1]}}

\usepackage{xspace}
\newcommand{\cdb}{CausalDriveBench\xspace}

\title{CausalDriveBench: Evaluating Causal Reasoning in Vision-Language-Action Models for Autonomous Driving}

\author{%
  Narendiran Chembu\thanks{Equal contribution.} \\
  Fastcode AI\\
  \texttt{naren@fastcode.ai} \\
  \And
  Navvrat Rao\footnotemark[1] \\
  Fastcode AI\\
  \texttt{navvrat@fastcode.ai} \\
  \And
  Shreedhar Shreeshail Kodate \\
  Renesas Electronics \\
  shreedhar.kodate.vz@renesas.com \\
  \And
  Gayatri Srujana Banda \\
  Renesas Electronics \\
  gayatri.banda.jy@renesas.com \\
  \And
  Arko Sarkar \\
  Renesas Electronics \\
  arko.sarkar.hx@renesas.com \\
  \And
  Abhinav Khanna \\
  Indian Institue of Technology Delhi \\
  cs1221964@iitd.ac.in \\
  \And
  Rajarshee Das \\
  Indian Institue of Technology Delhi \\
  cs1221124@iitd.ac.in \\
  \And
  Umesh Kanala \\
  Indian Institue of Technology Delhi \\
  cs1221111@iitd.ac.in \\
  \And
  Siddarth Khandelwal \\
  Fastcode AI \\
  siddarth@fastcode.ai \\
  \And
  Kumar Aman \\
  Fastcode AI \\
  aman@fastcode.ai \\
  \And
  Aish Dubey \\
  Renesas Electronics \\
  aish.dubey.xj@renesas.com \\
  \And
  Kaustubh Beedkar \\
  Indian Institue of Technology Delhi \\
  \texttt{kbeedkar@cse.iitd.ac.in} \\
  \AND
  Arjun Jain \\
  Fastcode AI\\
  \texttt{arjun@fastcode.ai}
}

\begin{document}

\maketitle

\begin{abstract}

Vision-Language-Action (VLA) models for autonomous driving produce natural-language reasoning alongside predicted trajectories, but whether this reasoning reflects the causal structure of the scene remains untested. We introduce CausalDriveBench, an evaluation framework grounded in Pearl's Causal Hierarchy (PCH) that tests causal reasoning in driving-specific VLAs through structured visual question answering (QA) and alternative-trajectory prediction. To this end, we construct causal scene graphs over nuScenes that distinguish causally active, dormant, and distractor entities, separating perceptual salience from causal relevance. The benchmark spans all four rungs of PCH (association, intervention, and counterfactual along with causal discovery) for QA generation. For the higher rungs, we additionally provide reference trajectories under specified scene modifications, enabling action-level verification that complements reasoning-level evaluation. In total, the benchmark contains 7,285 verified causal QA pairs and 1,000 counterfactual trajectories derived from nuScenes. We evaluate 10 driving-specific VLAs and 3 general-purpose VLMs, and report three findings. First, the best model reaches only 70.6\% QA accuracy, and 4 of 13 models score below random chance. Second, comparing each driving VLA to the general-purpose VLM that shares its language backbone, the cost of driving fine-tuning ranges from 2 to 34 percentage points on causal QA, with post-training design explaining the spread. Third, causal QA and trajectory accuracy are statistically uncorrelated across models: under counterfactual prompts, predicted trajectories either over-react or collapse onto the observed-scene baseline. Taken together, these results show that neither fluent rationales nor accurate observed-scene trajectories constitute evidence of causal understanding.

\end{abstract}


\section{Introduction}
\label{sec:intro}

\eat{

\kb{this needs some reshaping: we should first motivate why counterfactual reasoning matters in autonomous driving, then explain why grounded CF QA is hard to build at scale, then introduce our principled pipeline as a solution, and only then land the result that current VLA reasoning often appears weakly coupled to action. We don't want to sound like an ablation paper with dataset attached.}

\kb{suggested flow:
    \begin{itemize}
    	\item Why counterfactual reasoning matters in autonomous driving
    	\item why evaluating it is hard/non-trivial/not straighforward
    	\item Why a benchmark is needed
    	\item how our benchmark makes principled ground-truth supervision possinle
    	\item what did we learn when we evaluated current models
    	\item and what lies ahead in terms of this benchmark making inroards when used during training
    \end{itemize}
    }
}

Vision-language-action (VLA) models have emerged as a central paradigm for end-to-end autonomous driving, combining perception, scene understanding, language reasoning, and trajectory generation within a single model~\cite{wang2025alpamayo, zhang2025openread, li2025recogdrive, li2026unidrivevla}. 
In a safety-critical domain, however, accurate trajectories alone are insufficient. Models must also identify which scene elements are causally relevant to ego behavior, rather than relying on brittle statistical correlations or post-hoc rationales. Whether the rationales produced by these models reflect causal reasoning or merely fluent description is an open question.

Consider an ego vehicle approaching an intersection with a green light overhead, a pedestrian by the crosswalk, and a parked vehicle on the right shoulder. These three visible entities play different causal roles. The green light is \emph{active}: ego proceeds because of it. The pedestrian is \emph{dormant}: they do not constrain ego's present motion, but stepping into the crosswalk would make them causally relevant. The parked car is a \emph{distractor}: it is visible but has no plausible path to influencing ego's current behavior. These distinctions matter because what is visible is not the same as causal relevance. It also enables evaluation across Pearl's Causal Hierarchy (PCH)~\cite{pearl2009causal}: rung 1 associational (R1) queries ask what co-occurs with ego proceeding, rung 2 interventional (R2) queries ask what would change if the light turned red, and rung 3 counterfactual (R3) queries ask whether ego would still have proceeded, given that it did proceed, had the light been red. Counterfactual reasoning is particularly important because it conditions on the observed outcome while evaluating an alternative history, the form of reasoning required in rare and safety-critical driving situations.

Evaluating these capabilities is difficult for three main reasons. First, interventions and counterfactuals are not directly observed in driving data. Second, trajectory metrics like Average Displacement Error (ADE) and Collision Rate (CR) are agnostic to the reasoning that produced the trajectory. Third, existing benchmarks do not jointly test causal scene structure, language reasoning, and downstream action. Driving-language benchmarks~\cite{sima2024drivelm, chi2025impromptu} primarily evaluate description and instruction-following while general causal benchmarks~\cite{chen2024causal, chen2024cello} operate over static scenes without an embodied visual setting. As a result, no existing benchmark evaluates whether a driving VLA can connect causal scene structure to both its stated reasoning and its planned behavior.

We introduce \textbf{CausalDriveBench}, a benchmark for causal reasoning in driving VLAs grounded in PCH and built on real-world scenes. Each scene is encoded as a causal scene graph that distinguishes three causal statuses among visible entities: active, dormant and distractors. This partition separates perceptual salience from causal relevance and supports controlled scene interventions, yielding question-answer pairs and reference trajectory targets without free-form annotation. We derive 7{,}285 structured VQA pairs and 1{,}000 reference counterfactual trajectories (or alternative trajectories, used interchangeably) from nuScenes~\cite{caesar2020nuscenes} that cover causal discovery (R0) and the original three PCH rungs (R1--R3); the construction pipeline is designed to extend to additional driving datasets.


We evaluate 13 models: 10 end-to-end driving VLAs and 3 general-purpose VLMs. Three findings emerge. First, causal QA accuracy is brittle even at the top: the best model reaches only 70.6\%, and 4 of 13 models score below random chance. Second, comparing each general-purpose VLM to the driving VLA that shares its backbone, the cost of driving fine-tuning ranges from 2 to 34 percentage points on causal QA - with reward composition, not scale or architecture, explaining the spread. Third, language and action are statistically uncorrelated across models: under explicit counterfactual premises, most models either over-react or stay frozen at their scene baseline trajectory. Neither fluent rationales nor accurate observed-scene trajectories constitute sufficient evidence of causal understanding.

\section{Related Work} 
\label{sec:related_work}

\paragraph{Vision-Language-Action models for autonomous driving.}
Driving VLAs have evolved from passive explainers~\citep{xu2024drivegpt4} through modular pipelines~\citep{zhou2026opendrivevla, yang2025drivemoe, zhang2025safeauto} and unified end-to-end networks~\citep{hwang2024emma, renz2025simlingo, shao2024lmdrive} to reasoning-centric systems that place chain-of-thought at the centre of the control loop~\citep{fu2025orion, chi2025impromptu, zhang2024wisead, li2025recogdrive, zhang2025openread, jiang2025vla4adsurvey}. A subset has begun to engage with causal structure: OmniDrive~\citep{wang2025omnidrive} trains on simulated counterfactual QA, Counterfactual-VLA~\citep{peng2025counterfactual} revises meta-actions before emitting trajectories, Alpamayo-R1~\citep{wang2025alpamayo} introduces decision-grounded Chain-of-Causation reasoning traces, and CoC-VLA~\citep{zhang2025coc} structures output as a causal chain. Each embeds causal supervision into training, but none provides a structured evaluation that diagnoses where on a formal causal hierarchy a given model fails.

\begin{table}[t]
  \caption{Comparison with driving VQA and trajectory benchmarks. \emph{Graph} types: spatial scene graph (SG), QA-dependency graph (QA-Dep), action-rooted tree (Tree), causal scene graph (Causal). \emph{Causal Status} indicates whether active, dormant, and distractor entities are separated. \emph{PCH Layers}: Coverage of QA across rungs. \emph{CF Semantics}: forward simulation (Sim) versus Pearl-form counterfactuals (Pearl). \emph{Alt-traj GT}: ground-truth trajectories under counterfactual antecedents.}
  \label{tab:comparison}
  \centering
  \small
  \setlength{\tabcolsep}{4pt}
  \begin{tabular}{lccccc}
    \toprule
    Benchmark & Graph & Causal Status & PCH Layers & CF Semantics & Alt-traj GT \\
    \midrule
    nuScenes-QA~\citep{qian2024nuscenes}     & SG     & \xmark  & R1     & \xmark & \xmark  \\
    BDD-X~\citep{kim2018textual}             & \xmark & \xmark  & R1     & \xmark & \xmark  \\
    DriveLM~\citep{sima2024drivelm}          & QA-Dep & \xmark  & R1--R2 & \xmark & \xmark  \\
    Reason2Drive~\citep{nie2024reason2drive} & \xmark & \xmark  & R1--R2 & \xmark & \xmark  \\
    DriveLMM-o1~\citep{ishaq2025drivelmm}    & \xmark & \xmark  & R1--R2 & \xmark & \xmark  \\
    DriveAction~\citep{hao2025driveaction}   & Tree   & \xmark  & R1--R2 & \xmark & \xmark  \\
    Impromptu-VLA~\citep{chi2025impromptu}   & \xmark & \xmark  & R1--R2 & \xmark & \xmark  \\
    OmniDrive~\citep{wang2025omnidrive}      & \xmark & Partial & R1--R2 & Sim    & Partial \\
    \midrule
    CausalDriveBench (ours)                  & Causal & \cmark  & R0--R3 & Pearl  & \cmark  \\
    \bottomrule
  \end{tabular}
\end{table}

\paragraph{Comparison with existing driving QA benchmarks.}
Driving VLAs are supervised through sensor-rich corpora~\citep{caesar2020nuscenes, mei2022waymo, wilson2023argoverse, dosovitskiy2017carla, jia2024bench2drive} and a fast-growing layer of language benchmarks spanning action rationales~\citep{kim2018textual}, perceptual VQA~\citep{qian2024nuscenes}, graph-structured QA across perception, prediction, and planning~\citep{sima2024drivelm}, sequential reasoning chains~\citep{nie2024reason2drive, ishaq2025drivelmm}, action-rooted decision trees~\citep{hao2025driveaction}, and corner-case distillations~\citep{chi2025impromptu}. As Table~\ref{tab:comparison} summarises, the axis of variation across this body of work is task pipeline and scenario coverage rather than causal structure: none of these benchmarks distinguishes active, dormant, and distractor entities, covers all four rungs of PCH (R0--R3), or provides counterfactual reference trajectories. The closest comparison point is OmniDrive~\citep{wang2025omnidrive}, whose simulated alternatives are forward rollouts of different ego actions rather than Pearl-form counterfactuals where observed ego behaviour is a load-bearing premise. CausalDriveBench is the first benchmark to unify causal entity status, full PCH coverage, and data-grounded counterfactual trajectories within a single embodied driving setting.

\paragraph{Causal reasoning in machine learning.}
The formalism needed to evaluate causal structure comes from Pearl's structural causal models~\citep{pearl2009causal, bareinboim2022pearl}, which organise causal queries into associational, interventional, and counterfactual layers, with each higher layer encoding information unreachable from those below. Benchmarks operationalising this hierarchy exist for text~\citep{chen2024causal} and for multimodal scene understanding~\citep{komanduri2025causalvlbench}, but causal evaluation in the embodied driving setting remains absent. CausalDriveBench bridges this gap by grounding driving QA in causal scene graphs and structuring evaluation across all rungs of Pearl's hierarchy, distinguishing active, dormant, and distractor entities to expose precisely where on the causal ladder current driving VLAs fail.

\section{CausalDriveBench}
\label{sec:causaldrivebench}

\begin{figure*}[t]
\includegraphics[width=\textwidth, page=1]{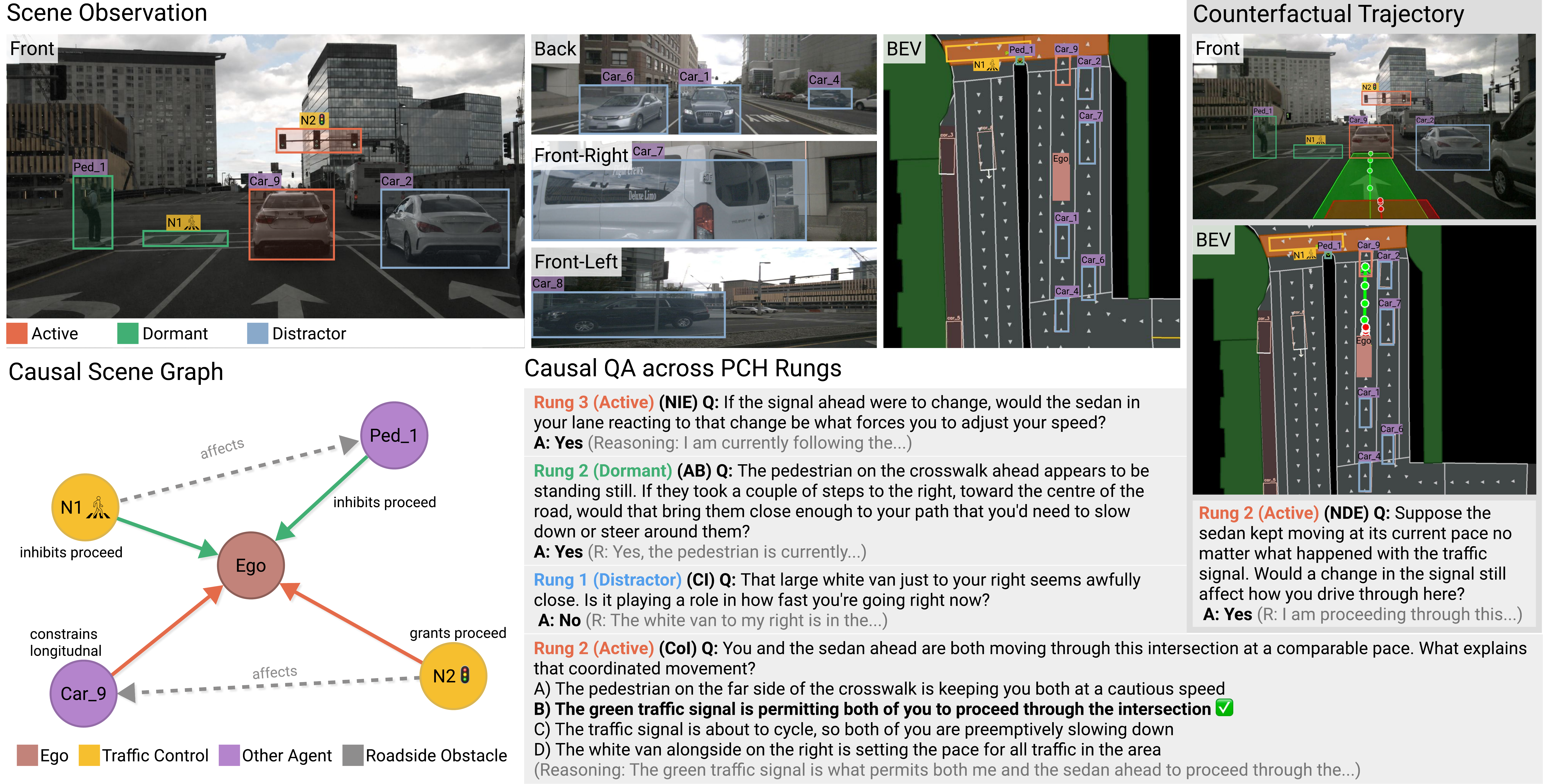}
\caption{\cdb Overview with an example: Each scene is encoded as a causal scene graph extracted from multi-view camera observations and BEV HD map context. The graph distinguishes three causal statuses among visible entities: active (warm), dormant (green), and distractor (blue). From this graph, we derive structured QA pairs spanning all four PCH layers and counterfactual ego trajectory targets under controlled scene modification.}
\label{fig:benchmark}
\end{figure*}

\cdb evaluates causal reasoning in driving VLAs across two complementary channels: structured causal QA (\cref{subsec:causal_qa}) and counterfactual trajectory prediction (\cref{subsec:cf_trajectory}). Both are grounded in a per-scene causal scene graph (\cref{subsec:causal_scene_graph}) that partitions visible entities by their causal relationship to ego (\cref{fig:benchmark}). Each input consists of multi-view camera images over a 2\,s observation window, a navigation command, ego history, and a structured query; the model produces a textual answer with reasoning and, for trajectory questions, a planned ego trajectory. Success on \cdb requires more than perceptual description or trajectory regression: a model must identify which entities cause ego's current behaviour, distinguish them from visually prominent but causally inert entities, reason about how behaviour would change under interventions and counterfactuals, and reflect that reasoning in its planned trajectory.

\subsection{Causal Scene Graph}
\label{subsec:causal_scene_graph}

Every benchmark label, including entity classifications, QA pairs, and counterfactual trajectories, is derived from a per-scene causal scene graph (\cref{fig:benchmark}; top-right panel) that encodes causal relations between visible entities and the ego vehicle.

\paragraph{Schema.}
For a scene, the graph $\cG = (\cV, \cE_{\text{ego}}, \cE_{\text{inter}})$ is an acyclic directed mixed graph (ADMG) over visually grounded scene entities, together with an auxiliary set $\cD$ of distractor nodes retained for evaluation. Nodes $\cV$ fall into four mutually exclusive types (ego, other agents, traffic control and road obstacles) and each records spatial position, per-camera visibility, lane assignment, and motion attributes. Ego edges $\cE_{\text{ego}} \subseteq \cV \times \{\text{ego}\}$ take one of four effect types (grants or inhibits proceed and constrains longitudinal or lateral). Both node and edge types derive from binary decision trees. Inter-node edges $\cE_{\text{inter}}$ encode dependencies between non-ego entities through directed observable causal influence or bi-directional unobserved common causes i.e., confounding (see \cref{app:graph_schema} for more details).

\paragraph{Causal status from interventional semantics.}
The defining feature of \cdb is its three-way partition of visible entities into \emph{active}, \emph{dormant}, and \emph{distractor} (illustrated in \cref{fig:benchmark}), grounded in Pearl's interventional semantics~\citep{pearl2009causal}. Every ego edge encodes the claim that removing the source entity, formally applying the intervention $\text{do}(X = \text{absent})$, would alter ego's behaviour. The partition follows from how this claim is evaluated.

\paragraph{Active.} 
An ego edge is \emph{active} if removing entity $X$ throughout the observation window $(t_0 - \Delta, t_0]$ would change ego's state at $t_0$:
\begin{equation}
    \text{active}(e) \iff
    S(\mathcal{G}_{t_0}) \neq S\!\left(\mathcal{G}_{t_0}^{-X}\right),
    \label{eq:active}
\end{equation}
where $S(\cdot) = (s, a, h)$ is ego's state vector (speed, acceleration, heading change), $\cG_{t_0}^{-X}$ is the counterfactual graph with $X$ absent throughout the window, and $\Delta = 2$\,s.  For example, the red edges from Car\_9 and N2 (green signal) in \cref{fig:benchmark} are active edges. We further sub-classify active edges temporally as \emph{sustaining} (binding at $t_0$) or \emph{triggering} (binding earlier in the window only); see \cref{app:causal_graph}.

\paragraph{Dormant.}
An ego edge is \emph{dormant} if it is currently inactive but could become active via a single physically realisable transition within decision horizon $\delta = 0.5$\,s:
\begin{equation}
    \text{dormant}(e) \iff
    \neg\,\text{active}(e) \,\wedge\,
    \exists\, x' \in R_\delta(X)
    \text{ s.t. } \text{active}(e \mid X \mapsto x'),
    \label{eq:dormant}
\end{equation}
The $\delta$-bound replaces the unbounded existential over ``plausible modifications'' with a single-step reachability test restricted to transitions arising from the entity's own agency, with geometric admissibility conditions in \cref{app:causal_graph}. For example, the green edges from Ped\_1 and N1 (crosswalk) in \cref{fig:benchmark} are dormant as these entities do not currently constrain the ego, but a physically realisable transition (such as pedestrian stepping on the road) could make them active within the decision horizon.

\paragraph{Distractor.}
A visible entity $X$ is a \emph{distractor} if no ego edge originating at $(X, \text{ego})$ is active or dormant:
\begin{equation}
    \text{distractor}(X) \iff
    \forall\, e = (X, \text{ego}) :
    \neg\,\text{active}(e) \,\wedge\,
    \neg\,\text{dormant}(e),
    \label{eq:distractor}
\end{equation}
As an example, the white van (marked with a blue box) in front right view in \cref{fig:benchmark} is a distractor as it has no plausible path of influencing ego’s current behavior. Distractors are excluded from $\cE_{\text{ego}}$ and recorded in $\cD$ with a rejection reason grounded in observable scene geometry.

\subsection{Causal QA}
\label{subsec:causal_qa}


Question types are organised by the causal status of the target entity (bottom-left panel of \cref{fig:benchmark}). \emph{Active-element questions} instantiate eleven scenarios from the CaLM framework~\citep{chen2024causal} over four canonical causal subgraphs (direct, chain, confounding, collider), spanning R0 (discovery) through R3 (counterfactual). \emph{Dormant-element questions} (novel contribution) cover R0--R2 with three types: Discrimination (DQ; distinguishing a dormant entity from an active one), Explanation of Latency (EL; reasoning about why a visible entity is not currently constraining ego), and Activation Boundary (AB; identifying the single transition that would activate a dormant edge). \emph{Distractor-element questions} (also novel) probe causal rejection R0--R3 with four types: Distractor Rejection (DR; identify all distractors), Causal Irrelevance (CI; rejection despite visual co-occurrence), Null Intervention (NI; recognising that $P(Y_\text{ego}\mid \text{do}(\text{remove}\ X)) = P(Y_\text{ego})$), and Wild Counterfactual (WC; reasoning about an alternative world where the distractor's structural function changes). All questions are binary or four-option multiple-choice; full type-by-rung mapping, subtypes, and isolation-safety analysis are in \cref{app:qa_taxonomy}.

\subsection{Counterfactual Trajectory Prediction}
\label{subsec:cf_trajectory}

For R2 and R3 binary QA pairs whose answers imply a behaviour change, we provide reference ego trajectories under the question's modified premise (lower-right panel of \cref{fig:benchmark}). These targets enable action-level verification that a model's predicted trajectory adapts to the modified causal structure: a yes/no answer alone cannot certify that planned behaviour reflects the reasoning produced. Each trajectory is synthesised under the modified scene graph and audited for collision-free passage and drivable-area compliance (\cref{subsec:human_review}); we treat them as physically and causally consistent reference trajectories rather than sensor-grounded ground truth, since any counterfactual evaluation in real-world driving inherits this constraint. Models are scored by Average Displacement Error against the reference, decomposed into CTE-R2 (intervention-conditioned, forward looking) and CTE-R3 (counterfactual-conditioned, past looking) sub-tasks; horizon and coordinate-frame details are in \cref{app:counterfactual_traj}. CTE-R3 is conditioned on a past-looking horizon, which is out-of-distribution relative to standard trajectory prediction; we report it as a stress test of counterfactual reasoning under format shift rather than as a like-for-like comparison with CTE-R2.

\providecolor{cAct1}{HTML}{B85450}  
\providecolor{cAct2}{HTML}{D08770}  
\providecolor{cAct3}{HTML}{E5B582}  
\providecolor{cAct4}{HTML}{8B5A3C}  
\providecolor{cAct5}{HTML}{A23E48}  
\providecolor{cAct6}{HTML}{C77A5E}  

\providecolor{cDor1}{HTML}{5F8B7A}  
\providecolor{cDor2}{HTML}{7FA89B}  
\providecolor{cDor3}{HTML}{A2C4B8}  

\providecolor{cDis1}{HTML}{4A6FA5}  
\providecolor{cDis2}{HTML}{6B8AB8}  
\providecolor{cDis3}{HTML}{8FA8C7}  
\providecolor{cDis4}{HTML}{B5C6DC}  

\providecolor{cR0}{HTML}{E6E0EE}    
\providecolor{cR1}{HTML}{BDA9D1}    
\providecolor{cR2}{HTML}{8B6FB0}    
\providecolor{cR3}{HTML}{5B3A87}    

\providecolor{cSGDirect}{HTML}{E5B582}
\providecolor{cSGCollider}{HTML}{D08770}
\providecolor{cSGConfounding}{HTML}{B85450}
\providecolor{cSGChain}{HTML}{8B5A3C}

\providecommand{\swatch}[1]{\tikz[baseline=-0.5ex]\node[fill=#1, draw=white,
  line width=0.3pt, minimum width=4.5pt, minimum height=4.5pt, inner sep=0pt,
  anchor=base] {};}


\newcommand{\stackbar}[3]{%
  \node[anchor=east, align=right, font=\footnotesize] at (-0.15, #1) {#2};
  \pgfmathsetmacro{\xstart}{0}%
  \xdef\smallcount{0}%
  \foreach \pct/\col/\lab in {#3} {%
    \pgfmathsetmacro{\xend}{\xstart + \pct * 0.11}%
    \fill[\col, draw=white, line width=0.5pt]
      (\xstart, #1 - 0.20) rectangle (\xend, #1 + 0.20);
    \pgfmathsetmacro{\xmid}{(\xstart + \xend)/2}%
    \ifdim \pct pt > 4pt
      \node[font=\scriptsize, gray!20!black] at (\xmid, #1) {\lab};
      \xdef\smallcount{0}%
    \else
      \ifx\\\lab\\\else
        \pgfmathparse{int(mod(\smallcount, 3))}%
        \ifcase\pgfmathresult
          \node[font=\scriptsize, gray!80, anchor=south]
            at (\xmid, #1 + 0.28) {\lab};
          \draw[gray!60, line width=0.3pt]
            (\xmid, #1 + 0.27) -- (\xmid, #1 + 0.20);
        \or
          \node[font=\scriptsize, gray!80, anchor=south]
            at (\xmid, #1 + 0.50) {\lab};
          \draw[gray!60, line width=0.3pt]
            (\xmid, #1 + 0.49) -- (\xmid, #1 + 0.20);
        \or
          \node[font=\scriptsize, gray!80, anchor=south]
            at (\xmid, #1 + 0.72) {\lab};
          \draw[gray!60, line width=0.3pt]
            (\xmid, #1 + 0.71) -- (\xmid, #1 + 0.20);
        \fi
        \pgfmathsetmacro{\tmpcount}{\smallcount + 1}%
        \xdef\smallcount{\tmpcount}%
      \fi
    \fi
    \xdef\xstart{\xend}%
  }%
}

\begin{figure*}[t]
\noindent%
\begin{tikzpicture}

\stackbar{1.9}{Question type}{%
  17.4/cDis1/CI,
  14.7/cDis2/NI,
  13.4/cDis3/WC,
  1.3/cDis4/{DR},
  9.5/cAct1/CR,
  7.8/cAct2/CaI,
  7.0/cAct3/SC,
  4.0/cAct4/Other,
  2.9/cAct5/{NC},
  2.0/cAct6/{CB},
  7.3/cDor1/DQ,
  6.5/cDor2/AB,
  6.2/cDor3/EL
}

\stackbar{1.3}{Active subgraph}{%
  68.1/cSGDirect/{Direct 68.1\%},
  14.8/cSGCollider/{Collider 14.8\%},
  11.8/cSGConfounding/{Confound.},
  5.3/cSGChain/{Chain}
}

\stackbar{0.7}{PCH layer}{%
  33.5/cR3/{R3 33.5\%},
  25.6/cR1/{R1 25.6\%},
  23.9/cR2/{R2 23.9\%},
  17.0/cR0/{R0 17.0\%}
}

\foreach \x/\lab in {0/0\%, 25/25\%, 50/50\%, 75/75\%, 100/100\%} {%
  \pgfmathsetmacro{\xpt}{\x * 0.11}%
  \draw[gray!40, line width=0.3pt, dashed]
    (\xpt, 0.5) -- (\xpt, 2.1);
  \node[font=\scriptsize, gray!70, anchor=north]
    at (\xpt, 0.4) {\lab};
}

\node[font=\footnotesize, anchor=north]
  at (5.5, 0.1) {%
    \swatch{cDis1}\,\swatch{cDis2}\,\swatch{cDis3}\,\swatch{cDis4}\,Distractor types
    \quad
    \swatch{cAct1}\,\swatch{cAct2}\,\swatch{cAct3}\,\swatch{cAct4}\,\swatch{cAct5}\,\swatch{cAct6}\,Active types
    \quad
    \swatch{cDor1}\,\swatch{cDor2}\,\swatch{cDor3}\,Dormant types
    \quad
    \swatch{cR0}\,\swatch{cR1}\,\swatch{cR2}\,\swatch{cR3}\,PCH layers
  };

\end{tikzpicture}

\caption{
Distribution of QA pairs in the diverse-sampled subset of
CausalDriveBench across three aspects: question type grouped by entity
family (top), active subgraph structure within active-element QA (middle), and
PCH layer (bottom). Slice colours encode entity status (red: active,
sage: dormant, blue: distractor). Rungs use purple gradients that are visually orthogonal to the entity-family hues. See \cref{app:dataset} for full per-type counts.
}
\label{fig:qa_distributions_diverse}
\end{figure*}
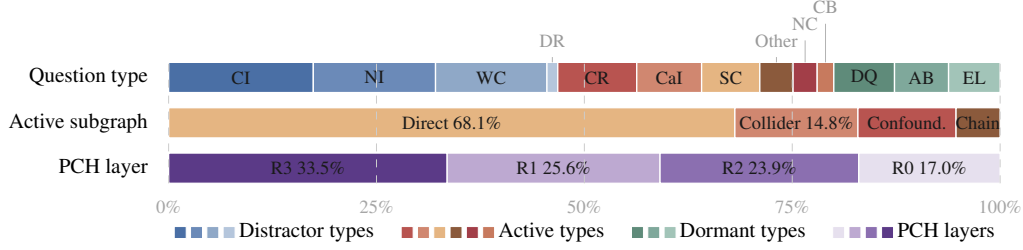

\subsection{Dataset Statistics}
\label{subsec:dataset_statistics}

\cdb comprises 7{,}285 QA pairs and 1{,}000 reference counterfactual trajectories (R2-209 and R3-791) drawn from 850 scenes from nuScenes~\cite{caesar2020nuscenes} trainval split (construction in \cref{sec:construction_and_validation}). Coverage across three aspects is shown in \cref{fig:qa_distributions_diverse}. Distractors dominate the entity-status partition (46.8\%) over active (33.2\%) and dormant (20.0\%) questions, reflecting the natural composition of visible entities. R3 dominates the PCH-layer mix (33.5\%) on the strength of Counterfactual Reasoning (CR), Wild Counterfactual (WC), and Sufficiency Cause (SC); R0--R2 split the remainder (17.0\% / 25.6\% / 23.9\%). 83.3\% of questions are binary and 16.7\% multiple-choice, the skew driven by the four highest-volume types (CI, NI, WC, CR), all binary by construction. Per-type counts, active-subgraph breakdowns, and per-scene coverage statistics are in \cref{app:dataset}.

\section{Benchmark Construction and Verification}
\label{sec:construction_and_validation}

\begin{figure}[t]
    \centering
    \includegraphics[width=\textwidth]{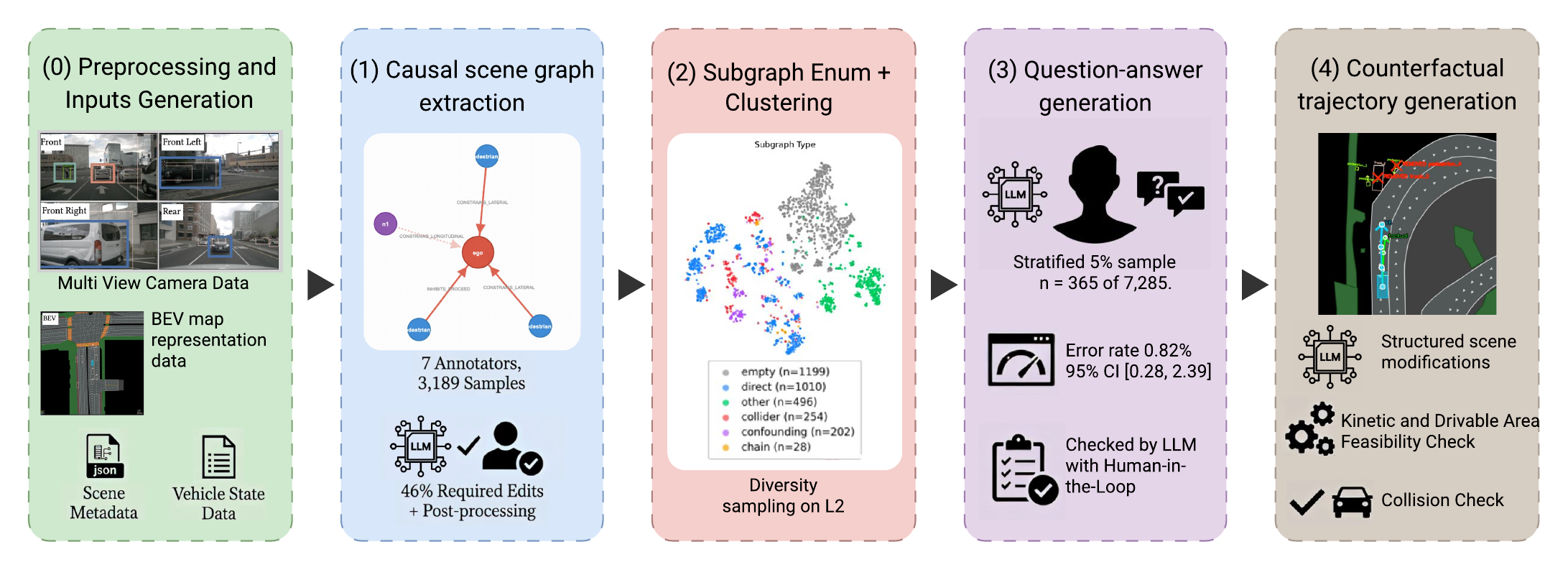}
    \caption{Construction pipeline and validation. Stages interleave deterministic logic, LLM extraction, and human review to produce 7{,}285 QA pairs and 1{,}000 counterfactual trajectories from 3{,}189 candidate samples. Three validation gates certify the released benchmark: graph review at Stage~1, an independent 5\% QA audit in Stage-3, and a trajectory audit at Stage~4. See \cref{sec:construction_and_validation} for more details.
    }
    \label{fig:pipeline}
\end{figure}

Constructing a causal-reasoning benchmark from real-world driving data requires balancing two competing demands: scaling annotation across thousands of scenes, and ensuring that every causal-status label, question, and trajectory target is faithful to the scene it describes. We address both through a hybrid pipeline that interleaves LLM-based extraction with deterministic post-processing and structured human review (\cref{fig:pipeline}).

\subsection{Construction Pipeline}
\label{subsec:pipeline}

The benchmark is constructed in five stages (\cref{fig:pipeline}). Stages~1, 3, and~4 are LLM-assisted; Stages~0 and~2 are fully deterministic; the output of Stage~1 is gated by human review before flowing into the downstream stages. All prompts across stages are released as part of the Supplementary material.

\paragraph{Stage 0: Preprocessing and inputs (deterministic).} 2D bounding boxes for visual entities are projected and rendered onto camera frames with entity labels, the HD map is rasterised and overlaid with ego and agent footprints to produce a top-down BEV, and ego pose plus per-frame annotations are compiled into a structured state record (navigation command, ego kinematics, and per-agent positions). The four-camera stack and BEV are produced for each of four timesteps spanning $[t_0 - 1.5\,\text{s},\, t_0]$ at 2\,Hz.

\paragraph{Stage 1: Causal scene graph extraction (LLM).} For each scene, we extract a causal scene graph in a single pass using Claude Opus~4.6~\citep{claude2025opus46}, conditioned on multi-view images, BEV frames, and agent state (position, speed, heading) over a 2\,s window. The prompt uses structured decision trees to infer entities, ego edges, causal status, and inter-node relations (\cref{app:causal_graph_extraction}). Every extracted graph is subsequently reviewed and edited by human annotators (\cref{subsec:human_review}) before flowing downstream.

\paragraph{Stage 2: Subgraph enumeration and diversity sampling (deterministic).} We deterministically extract canonical causal subgraphs (direct, chain, collider, confounding) using fixed priority rules and generate per-scene subgraph manifests for QA generation. To select a diverse subset, scenes are hierarchically clustered by causal structure (L1) and scene context (L2), retaining one high-richness representative per L2 cluster (\cref{app:subgraph_enumeration,app:clustering}).

\paragraph{Stage 3: Question--answer generation (LLM + rules).} QA pairs are generated using three role-specific prompts targeting active, dormant, and distractor entities, instantiating CaLM scenarios for active elements, discrimination and activation queries for dormant elements, and rejection queries for distractors. The prompt applies four principles during QA generation (see \cref{app:answer_gen,app:qa_generation} for full prompt design).

\paragraph{Stage 4: Counterfactual trajectory generation (LLM + tools + rules).} For each scene, we synthesise reference counterfactual trajectories for at most one Rung~2 and one Rung~3 binary question whose answer implies a behaviour change, prioritising scene-specific over policy-level interventions. Synthesis chains an LLM judge that identifies pinned auxiliary entities, an LLM that produces numeric agent state mutations from the camera and BEV stack, and a tool-using agent that commits ego waypoints against deterministic collision, lane-snap, and kinematic-feasibility checks; a deterministic post-process then integrates committed waypoints into a smoothed trajectory whose physical consistency is guaranteed by construction (\cref{app:counterfactual_traj}).

\subsection{Verification Pipeline}
\label{subsec:human_review}


\paragraph{Causal Graph Verification}
Because Stage~1 is load-bearing, every extracted graph passes through independent human review before downstream stages. Of 3{,}400 samples, 211 agent-empty cases were filtered, and 46\% of the remaining 3{,}189 valid graphs required at least one edit. 90\% of those edits were causal-status flips that most often demote a Stage-1 cause to the distractor pool, evidence that frontier MLLMs do not yet reliably identify causal structure (can also be due to imperfect prompts). Recurring patterns across the review pool were distilled into post-processing rules applied uniformly to all extracted graphs, ensuring the same class of error cannot reappear (annotator pool, interface, full edit-type counts, rule list and inter-annotator agreement scores are in \cref{app:human_review}).

\paragraph{Verification of Causal QAs} To certify label fidelity beyond the per-sample review of graphs, we audited a stratified 5\% random sample of the 7{,}285 released QA pairs ($n=365$). The audit found an observed error rate $\hat{p} = 0.82\%$ (95\% Wilson CI $[0.28,\,2.39]\%$), rejecting an error rate exceeding 5\% at $p = 0.000125$ (one-sided $z$-test, $z = -3.66$). More details in \cref{app:qa_verification}.

\paragraph{Counterfactual Trajectory Audit} All synthesised counterfactual trajectories are audited for collision-free passage and drivable-area compliance: any trajectory intersecting an active agent's category-sized footprint or leaving the lane network defined by the post-intervention scene is eliminated. Of the 1,177 candidate pool, 177 trajectories were eliminated (see \cref{app:trajectory_audit}), leaving 1{,}000 verified counterfactual trajectory targets in the released benchmark.




\section{Evaluation}
\label{sec:eval}

\subsection{Setup}
\label{subsec:eval_framework}


We evaluate 10 driving-specific VLAs and 3 general-purpose VLMs of comparable size ($\approx$8B parameters). Each model receives multi-camera frames across 4 timesteps, ego history, and navigation command alongside a causal-question prompt. All experiments run on a single NVIDIA A100 80\,GB GPU (24 vCPUs, 220\,GB RAM). For binary and multiple-choice questions we parse responses by string matching (Yes/No, A/B/C/D) against the ground-truth label; unparsable outputs are counted as wrong, which lowers effective coverage. Inference VQA and trajectory-generation prompt templates are given in \cref{app:Inference Prompts}.

We report QA accuracy stratified by PCH layer and entity status, and LLM-judge reasoning scores on a 0--2 scale (0: missing, 1: partial, 2: present) across four dimensions: Spatial grounding, Mechanism identification, Conditionality awareness, and Answer consistency (abbreviated in \cref{tab:vqa_results}); Composite is their mean. Each reasoning score is averaged over Claude Sonnet 4.6 \citep{claude2025sonnet46} and OpenAI GPT-4o \citep{openai2025gpt4o}, which agree at Pearson $r = 0.97$ on Composite (\cref{tab:judge_agreement}). For action evaluation, Baseline Trajectory Error (BTE) is the Average Displacement Error (ADE, in metres) between predicted and ground-truth ego trajectories over 3\,s (6 waypoints at 2\,Hz). Counterfactual Trajectory Error (CTE) is the corresponding ADE for the ground-truth trajectory under the structured scene change posed in an R2 or R3 question, split into CTE-R2 (intervention-conditioned, forward-looking, 3\,s / 6 waypoints) and CTE-R3 (counterfactual-conditioned, past-looking, 2\,s / 4 waypoints), both at 2\,Hz.

\begin{table}[t]
\centering
\caption{VQA accuracy (\%) and LLM-judge reasoning scores (0--2 per dimension, higher is better). Best results in each column are in bold. See \cref{subsec:eval_framework} for more details.}
\label{tab:vqa_results}
\resizebox{\textwidth}{!}{%
\begin{tabular}{lrrcccccccccccc}
\toprule
& & & \multicolumn{4}{c}{PCH layer} & \multicolumn{3}{c}{Entity type} & \multicolumn{5}{c}{LLM-judge axis} \\
\cmidrule(lr){4-7} \cmidrule(lr){8-10} \cmidrule(lr){11-15}
Model & $n$ & Overall & R0 & R1 & R2 & R3 & Active & Dormant & Distractor & Comp. & Spat. & Mech. & Cond. & Cons. \\
\midrule
\multicolumn{15}{l}{\textit{Driving-specific VLAs}} \\
ImpromptuVLA-7B \citep{chi2025impromptu}    & 7285 & 69.27\% & 66.43\% & 59.41\% & \textbf{85.10\%} & \textbf{66.93\%} & 69.07\% & 63.92\% & 71.69\% & 1.06 & 1.08 & 1.12 & 0.63 & 1.39 \\
OpenREAD \citep{zhang2025openread}          & 7274 & 63.76\% & 70.10\% & 48.95\% & 78.89\% & 61.03\% & 69.81\% & 60.62\% & 60.80\% & 1.18 & 1.14 & 1.30 & 0.83 & 1.44 \\
Orion \citep{fu2025orion}                   & 7285 & 57.25\% & 25.44\% & \textbf{76.03\%} & 66.40\% & 52.58\% & 30.27\% & 53.26\% & \textbf{78.08\%} & 0.19 & 0.14 & 0.07 & 0.06 & 0.51 \\
UniDrive-VLA \citep{li2026unidrivevla}      & 7285 & 54.87\% & 63.69\% & 29.71\% & 68.24\% & 60.08\% & 65.88\% & 53.75\% & 47.54\% & 1.12 & 1.21 & 1.24 & 0.74 & 1.31 \\
RecogDrive \citep{li2025recogdrive}         & 7158 & 54.25\% & 57.80\% & 38.19\% & 59.20\% & 61.32\% & \textbf{69.91\%} & 48.55\% & 45.92\% & 1.25 & 1.36 & 1.40 & 0.75 & 1.48 \\
SafeAuto \citep{zhang2025safeauto}          & 7285 & 49.42\% & 51.61\% & 43.16\% & 56.62\% & 47.99\% & 56.62\% & 41.79\% & 47.57\% & 0.50 & 0.57 & 0.35 & 0.16 & 0.93 \\
WiseAD \citep{zhang2024wisead}              & 7285 & 39.35\% & 22.87\% & 22.20\% & 44.42\% & 57.30\% & 50.58\% & 32.10\% & 34.50\% & 0.71 & 0.81 & 0.67 & 0.27 & 1.07 \\
Alpamayo-1.5-10B \citep{wang2025alpamayo}   & 7285 & 36.86\% & 36.07\% & 21.66\% & 39.87\% & 46.72\% & 46.36\% & 30.72\% & 32.74\% & 0.71 & 0.87 & 0.78 & 0.38 & 0.81 \\
SimLingo \citep{renz2025simlingo}           & 7285 & 33.95\% & 39.61\% & 25.95\% & 31.42\% & 38.98\% & 44.71\% & 34.23\% & 26.20\% & 0.71 & 0.91 & 0.70 & 0.30 & 0.93 \\
Omnidrive \citep{wang2025omnidrive}         & 7212 & 24.10\% & 6.74\%  & 25.89\% & 16.56\% & 37.04\% & 30.86\% & 20.59\% & 20.88\% & 0.65 & 0.87 & 0.70 & 0.40 & 0.62 \\
\midrule
\multicolumn{15}{l}{\textit{General-purpose VLMs}} \\
Cosmos-Reason-2 \citep{azzolini2025cosmos}  & 7285 & \textbf{70.56\%} & \textbf{78.74\%} & 68.69\% & 74.45\% & 65.04\% & 68.73\% & \textbf{70.17\%} & 72.01\% & 1.19 & 1.15 & 1.25 & 0.70 & \textbf{1.65} \\
Qwen3-VL-8B \citep{bai2025qwen3}            & 7285 & 66.04\% & 53.38\% & 64.99\% & 82.74\% & 61.39\% & 56.74\% & 64.12\% & 73.45\% & \textbf{1.35} & \textbf{1.41} & \textbf{1.45} & \textbf{0.99} & 1.55 \\
InternVL-3.5-8B \citep{chen2024internvl}    & 7285 & 62.91\% & 54.35\% & 51.74\% & 78.31\% & 64.84\% & 64.72\% & 56.70\% & 64.27\% & 1.19 & 1.25 & 1.30 & 0.86 & 1.34 \\
\midrule
Cross-model mean                            & ---  & 52.51\% & 48.22\% & 44.35\% & 60.17\% & 55.48\% & 55.71\% & 48.50\% & 51.97\% & 0.91 & 0.98 & 0.95 & 0.54 & 1.16 \\
\bottomrule
\end{tabular}
}
\end{table}

\subsection{Key Findings}
\label{subsec:eval_results} 

\paragraph{Causal reasoning is brittle even at the top of the leaderboard.}
The best model (Cosmos-Reason2) reaches 70.6\% QA accuracy, only 24.8\,pp above the 45.8\% random-chance baseline given the benchmark's binary--MCQ mix; its 4.5--7.7\,pp margin over the other two VLMs is consistent with its world-model post-training (Physical-AI SFT/RL atop a Qwen2.5-VL backbone). 7 of 13 models score below 55\%; 4 score below 40\%. The LLM-judge breakdown locates the deficit precisely: conditionality awareness is the weakest reasoning dimension for every model on the leaderboard, with cross-model mean 0.54 against 1.16 for answer consistency (\cref{tab:vqa_results}). Counter to the expected monotonic trend across PCH layers, R1 accuracy (cross-model mean 44.4\%) sits \emph{below} both R2 (60.2\%) and R3 (55.5\%); the dip is driven by the Collider Bias (CB) question type, where strong models suffer answer-format bias while weak models exploit it (\cref{app:cb_gap}). The active counterfactual battery and the distractor Wild Counterfactual stay at moderate accuracy throughout (cross-type means 53--58\%) with no model dominant across the suite (\cref{app:per_qa_type_accuracy}). The pattern is not confined to the bottom of the table: Orion answers 57.3\% of questions correctly while posting the lowest reasoning composite (0.19) and conditionality (0.06) on the leaderboard, and Cosmos-Reason-2 leads on QA but ties for third on judge composite (1.19, with InternVL-3.5-8B) behind Qwen3-VL-8B (1.35) and RecogDrive (1.25). Fluent answers can coexist with content-thin reasoning, and the QA rank does not predict the reasoning rank.

\paragraph{Driving fine-tuning hurts causal QA unevenly, and reward design explains why.}
General-purpose VLMs lead driving VLAs by 18\,pp on average causal QA accuracy. To isolate the cost of driving-specific post-training from confounding effects of backbone and scale, we compare each of the 4 driving VLAs whose base language model matches one of our 3 VLMs against that VLM directly. The cost ranges from $-2.3$\,pp (OpenREAD on Qwen3-VL-8B) to $-33.7$\,pp (Alpamayo-1.5-10B on Cosmos-Reason-2): a 31\,pp spread within architecturally matched pairs. Two driving VLAs sharing the \emph{same} Qwen3-VL-8B backbone (OpenREAD and UniDrive-VLA) diverge by 9\,pp, attributable entirely to differences in their RFT reward design: rewards that couple model reasoning to in-distribution trajectories produce the largest losses, while rewards that grade QA correctness independently of action produce the smallest (\cref{app:backbone_analysis,app:rft_analysis}).

\paragraph{Driving post-training also inverts the direction of causal-discrimination bias.}
The Active--Distractor accuracy gap averages $+6.8$\,pp across the 10 driving VLAs (over-attributing causal status to perceptually salient distractors) and $-6.5$\,pp across the 3 VLMs (over-rejecting), with 8 of 10 VLAs and 2 of 3 VLMs on opposite sides of zero (\cref{app:cross_cutting}). 
An attention-masking probe on a single representative scene provides an existence proof: replacing each model's chain-of-thought with rationales prescribing mutually inconsistent behaviours leaves AlpaMayo-1.5's predicted trajectory unchanged in every input-mask configuration, and OpenREAD's unchanged whenever vision is available. For these four models on this scene, the reasoning channel is not load-bearing for trajectory generation ((see \cref{app:attention_masking,app:representational_analysis}).

\paragraph{Language and motor planning are statistically decoupled.}
Trajectory regression and causal QA accuracy are essentially uncorrelated across the 13 models: Pearson(BTE, VQA) $r = -0.1$, and a model's BTE rank and QA rank differ by 4.5 of 13 positions on average. UniDrive-VLA holds the best BTE (0.68\,m) but only 7th-ranked QA; Cosmos-Reason-2 holds the best QA but ranks 5th on BTE (1.79\,m) much better than the other two VLMS: InternVL-3.5-8B ranks 12th on BTE but 5th on QA. Under counterfactual prompts, predicted trajectories either drift from the synthesised counterfactual targets (4.46\,m interventional, 3.09\,m counterfactual ADE) or stay close to the observed-scene trajectory; \cref{app:cte_regimes} decomposes this dichotomy into three structurally distinct failure modes (frozen, wild-swerving, default-to-stop). WiseAD is the only model whose CTE-R2 (1.97\,m) and CTE-R3 (1.57\,m) ADE fall below its BTE (2.91\,m) for a \emph{positive} reason: proportionate counterfactual differentiation rather than collapse to a short default trajectory; four other models (Omnidrive, Orion, SafeAuto, InternVL-3.5-8B) achieve sub-BTE CTEs only by defaulting to short generic trajectories on already-poor baselines.

\paragraph{Implications.}
Together, these findings position \cdb as a diagnostic instrument rather than a leaderboard: neither fluent rationales nor accurate observed-scene trajectories constitute evidence of causal understanding. The training-paradigm dependence of causal-QA degradation (\cref{app:rft_analysis}) and the language--action decoupling together point to concrete targets for next-generation driving VLAs that operate in a safety-critical domain: post-training reward designs that explicitly grade causal-reasoning quality, and tighter coupling between the language and trajectory heads.

\begin{table}[t]
\centering
\caption{Trajectory error comparison of models (ADE in metres). Best mean ADE results are in bold. See \cref{subsec:eval_framework} for more details.}
\label{tab:trajectory_ade}
\resizebox{\textwidth}{!}{%
\begin{tabular}{l rrrrr rrrrr rrrrr}
\toprule
& \multicolumn{5}{c}{BTE (m)} & \multicolumn{5}{c}{CTE-R2 (m)} & \multicolumn{5}{c}{CTE-R3 (m)} \\
\cmidrule(lr){2-6} \cmidrule(lr){7-11} \cmidrule(lr){12-16}
Model & $n$ & Mean & Std & Min & Max & $n$ & Mean & Std & Min & Max & $n$ & Mean & Std & Min & Max \\
\midrule
\multicolumn{16}{l}{\textit{Driving-specific VLAs}} \\
UniDrive-VLA \citep{li2026unidrivevla}       & 815 & \textbf{0.68} & 0.60 & 0.02 & 4.51  & 209 & 2.90 & 1.74 & 0.14 & 8.06  & 791 & 1.96 & 1.60 & 0.04 & 10.49 \\
RecogDrive \citep{li2025recogdrive}         & 815 & 1.19          & 0.65 & 0.14 & 6.63  & 209 & 4.30 & 2.21 & 0.44 & 11.66 & 791 & 1.97 & 1.41 & 0.22 & 9.77  \\
ImpromptuVLA-7B \citep{chi2025impromptu}    & 814 & 1.27          & 1.58 & 0.00 & 18.45 & 203 & 5.69 & 4.58 & 0.00 & 19.63 & 781 & 2.97 & 2.49 & 0.00 & 15.40 \\
Alpamayo-1.5-10B \citep{wang2025alpamayo}   & 815 & 1.31          & 1.25 & 0.00 & 6.79  & 209 & 2.70 & 2.16 & 0.01 & 11.90 & 791 & 1.70 & 1.50 & 0.00 & 10.25 \\
WiseAD \citep{zhang2024wisead}              & 814 & 2.91          & 3.00 & 0.00 & 16.93 & 205 & \textbf{1.97} & 1.80 & 0.05 & 9.96  & 767 & \textbf{1.57} & 1.37 & 0.00 & 9.56  \\
OpenREAD \citep{zhang2025openread}           & 810 & 3.97          & 6.73 & 0.00 & 63.57 & 189 & 4.34 & 4.65 & 0.00 & 41.11 & 672 & 4.61 & 4.77 & 0.00 & 29.46 \\
SimLingo \citep{renz2025simlingo}          & 815 & 4.65          & 2.88 & 0.52 & 16.32 & 209 & 5.07 & 2.78 & 0.63 & 13.65 & 791 & 2.12 & 1.77 & 0.28 & 11.38 \\
Omnidrive \citep{wang2025omnidrive}          & 815 & 5.39          & 2.67 & 0.22 & 15.26 & 209 & 4.51 & 1.92 & 1.29 & 8.55  & 790 & 2.48 & 1.34 & 0.17 & 9.79  \\
SafeAuto \citep{zhang2025safeauto}          & 815 & 5.43          & 4.24 & 0.00 & 18.06 & 209 & 4.53 & 3.07 & 0.01 & 13.00 & 791 & 3.35 & 1.77 & 0.57 & 12.63 \\
Orion \citep{fu2025orion}             & 815 & 8.10          & 4.93 & 0.28 & 23.27 & 209 & 7.88 & 3.56 & 0.86 & 16.65 & 791 & 4.73 & 2.88 & 0.02 & 14.36 \\
\midrule
\multicolumn{16}{l}{\textit{General-purpose VLMs}} \\
Qwen3-VL-8B \citep{bai2025qwen3}       & 806 & 3.68          & 4.10 & 0.00 & 29.54 & 181 & 5.76 & 4.67 & 0.00 & 21.78 & 750 & 3.58 & 3.19 & 0.00 & 19.04 \\
Cosmos-Reason-2 \citep{azzolini2025cosmos}    & 815 & 1.79          & 1.63 & 0.00 & 7.76  & 209 & 3.46 & 3.73 & 0.00 & 21.53 & 791 & 4.91 & 4.84 & 0.00 & 25.14 \\
InternVL-3.5-8B \citep{chen2024internvl}   & 815 & 7.71          & 6.76 & 0.00 & 30.27 & 201 & 4.81 & 4.03 & 0.00 & 22.58 & 787 & 4.18 & 3.93 & 0.00 & 24.34 \\
\midrule
Cross-model mean   & --- & 3.70          & ---  & ---  & ---   & --- & 4.46 & ---  & ---  & ---   & --- & 3.09 & ---  & ---  & ---   \\
\bottomrule
\end{tabular}
}
\end{table}

\section{Limitations and Conclusion}
\label{sec:conclusion}


\paragraph{Limitations}
The current release is built on nuScenes, which restricts the benchmark to urban driving in Boston and Singapore at 2\,Hz keyframe rate; the construction pipeline is dataset-agnostic and we plan to extend it to Argoverse2 and OpenScenes for highway and suburban coverage. Graph labels are produced by an LLM-assisted pipeline with mandatory human review at every scene: 46\% of extracted graphs received at least one correction during review, after which an independent 5\% audit on the resulting QA pairs reported a 0.82\% error rate (\cref{subsec:human_review}). Counterfactual trajectory targets are synthesised by a planner or LLM conditioned on the modified scene graph and audited for collision-free passage and drivable-area compliance, since the human-driver counterfactual is by construction unobservable in recorded data.

\paragraph{Conclusion}
We presented \cdb, the first benchmark to evaluate causal reasoning in driving VLAs across all four rungs of Pearl's Causal Hierarchy through 7{,}285 structured VQA pairs and 1{,}000 counterfactual trajectories, all grounded in densely informative causal scene graphs. Across 10 driving VLAs and 3 general-purpose VLMs, language and action remain decoupled: VQA accuracy and trajectory error are statistically uncorrelated, and predicted trajectories rarely respond appropriately to counterfactual prompts. Driving-specific post-training degrades causal QA by up to 33.7\,pp on the same backbone and induces a learned bias toward perceptually salient distractors that is absent from the base VLMs. Current driving VLAs do not yet meet the causal-reasoning bar that safety-critical autonomy demands.

Closing this gap requires sustained research on three fronts: post-training rewards that explicitly grade causal-reasoning quality, supervision signals that penalise reliance on perceptual salience, and architectures that wire the language channel into motor planning rather than alongside it. We release \cdb as a diagnostic benchmark that makes these failure modes measurable, in autonomous driving and in embodied agents more broadly.

\bibliographystyle{plainnat}
\bibliography{references}

\appendix

\section{Formal Causal Scene Graph Definitions}
\subsection{Graph Schema}
\label{app:graph_schema}

\paragraph{Nodes.} The set of nodes $\mathcal{V}$ contains discrete visually grounded entities drawn from the four categories listed in Table~\ref{tab:node_categories}. We derive these categories from a decision tree over two observable binary properties: \emph{agency} (can the entity's future state change by its own volition?), and \emph{rule imposition} (does the entity encode a codified traffic regulation?). Combined with the unique ego node, this yields four mutually exclusive, collectively exhaustive types (Figure~\ref{fig:taxonomy_trees}a). The taxonomy is closed: any novel entity must possess or lack agency, must or must not impose a traffic rule placing it in exactly one leaf.

Each node carries attributes: type, semantic label, position relative to ego (in ego-centric coordinates where $+x$ is forward and $+y$ is left), visibility across camera views, lane assignment, and motion state (stationary, moving, predicted trajectory). Only entities visible in at least one camera view are included.

\begin{table}[h]
\centering
\caption{Node categories in the causal scene graph, derived from the decision tree in Figure~\ref{fig:taxonomy_trees}a.}
\label{tab:node_categories}
\begin{tabular}{lp{6.2cm}}
\toprule
\textbf{Category} & \textbf{Examples} \\
\midrule
\texttt{EGO} & The ego vehicle itself (unique, exactly one per scene) \\
\texttt{OTHER\_AGENT} & Cars, trucks, buses, pedestrians, cyclists, active emergency vehicles - any entity whose future state can change by its own volition \\
\texttt{TRAFFIC\_CONTROL} & Signals, signs, stop lines, crosswalk markings - entities that impose a codified traffic regulation \\
\texttt{ROAD\_OBSTACLE} & Construction cones, temporary barriers, debris - inert, localised entities requiring only spatial avoidance \\
\bottomrule
\end{tabular}
\end{table}

\paragraph{Ego edges.} The edge set $\mathcal{E}_{\text{ego}} \subseteq \mathcal{V} \times \{\text{ego}\}$ contains directed edges from scene entities to the ego vehicle. We derive the edge taxonomy from a parallel decision tree (Figure~\ref{fig:taxonomy_trees}b) over three binary properties of the causal influence on ego: \emph{discrete vs.\ continuous} (is the control constraint a go/no-go gate or a modulation of the current trajectory?), and \emph{axis or polarity} (for continuous constraints, does it act on speed or heading? for discrete constraints, does it permit or prohibit?). This yields four mutually exclusive, collectively exhaustive effect types (Table~\ref{tab:edge_effects}). As with nodes, the taxonomy is closed by construction: any control constraint is either discrete or continuous; and each further splits on a binary property of either permission or polarity. 

Each edge $e = (X, \text{ego})$ carries four attributes: (i) an \emph{effect type} from Table~\ref{tab:edge_effects}; (ii) a \emph{causal status} $\mathrm{cs}(e) \in \{\textsc{active-sustaining}, \textsc{active-triggering}, \textsc{dormant}\}$ recording whether $X$ is constraining ego at $T{=}0$ (sustaining), was constraining ego during the 2s observation window but is no longer (triggering), or is not constraining ego but lies one realistic transition away from doing so (dormant); (iii) an \emph{affected-state} field $\mathcal{D}(e) \subseteq \{s, a, h\}$ specifying which ego state dimensions are influenced; and (iv) a \emph{justification} grounding the relationship in observable scene properties with explicit spatial coordinates. The causal-status field is the formal hook that separates active and dormant entities in Pearl's sense: active edges carry empirical behavioral evidence at during the observation window, while dormant edges satisfy a
single-step reachability test (\cref{app:causal_status}).

\begin{table}[h]
\centering
\caption{Ego edge effect types, derived from the decision tree in Figure~\ref{fig:taxonomy_trees}b.}
\label{tab:edge_effects}
\begin{tabular}{lp{5.8cm}l}
\toprule
\textbf{Effect type} & \textbf{Description} & \textbf{$\mathcal{D}(e)$} \\
\midrule
\texttt{GRANTS\_PROCEED} & Signal or sign permitting ego to enter a region (discrete, permits) & $s, a$ \\
\texttt{INHIBITS\_PROCEED} & Entity blocking or prohibiting ego from entering a region: yielding, obstruction, gap waiting (discrete, prohibits) & $s, a$ \\
\texttt{CONSTRAINS\_LONGITUDINAL} & Entity modulating ego's speed or acceleration within its current trajectory: lead-vehicle following, speed limiting (continuous, speed axis) & $s, a$ \\
\texttt{CONSTRAINS\_LATERAL} & Entity modulating ego's heading or lane position: lateral bounding, evasive steering (continuous, heading axis) & $h$ \\
\bottomrule
\end{tabular}
\end{table}

\begin{figure}[h!]
\centering

\begin{minipage}{\textwidth}
\centering
\resizebox{\textwidth}{!}{%
\begin{tikzpicture}[
    >=Stealth,
    decision/.style={
        rectangle, rounded corners=3pt,
        draw=qcolor, thick, fill=qcolor!8,
        text width=3.2cm, align=center,
        font=\small\sffamily, minimum height=0.85cm,
    },
    leaf/.style={
        rectangle, rounded corners=3pt,
        draw=leafgreen!80!black, thick, fill=leafgreen!12,
        text width=2.8cm, align=center,
        font=\small\ttfamily\bfseries, minimum height=0.8cm,
    },
    yeslbl/.style={font=\footnotesize\sffamily\bfseries, text=yescolor, fill=white, inner sep=2pt},
    nolbl/.style={font=\footnotesize\sffamily\bfseries, text=nocolor, fill=white, inner sep=2pt},
    arr/.style={draw=qcolor!70, thick, -Stealth, rounded corners=4pt},
]

\node[decision]  at ( 1.25,  0.0)  (q1)  {Is it ego?};
\node[leaf]      at (-4.0,  -2.0)  (ego) {EGO};
\node[decision]  at ( 6.5,  -2.0)  (q2)  {Has agency?};
\node[leaf]      at ( 2.0,  -4.0)  (agt) {OTHER\_AGENT};
\node[decision]  at (11.0,  -4.0)  (q3)  {Imposes codified\\traffic rule?};
\node[leaf]      at ( 8.0,  -6.0)  (tc)  {TRAFFIC\_CONTROL};
\node[leaf]      at (14.0,  -6.0)  (ro)  {ROAD\_OBSTACLE};

\draw[arr] (q1.south) -- ++(0,-0.65) -| (ego.north)
    node[yeslbl, pos=0.25, left] {Yes};
\draw[arr] (q1.south) -- ++(0,-0.65) -| (q2.north)
    node[nolbl, pos=0.25, right] {No};
\draw[arr] (q2.south) -- ++(0,-0.65) -| (agt.north)
    node[yeslbl, pos=0.25, left] {Yes};
\draw[arr] (q2.south) -- ++(0,-0.65) -| (q3.north)
    node[nolbl, pos=0.25, right] {No};
\draw[arr] (q3.south) -- ++(0,-0.65) -| (tc.north)
    node[yeslbl, pos=0.25, left] {Yes};
\draw[arr] (q3.south) -- ++(0,-0.65) -| (ro.north)
    node[nolbl, pos=0.25, right] {No};

\end{tikzpicture}%
}
\smallskip

{\small\textbf{(a)} Node-type decision tree. Two binary properties, agency and rule imposition, together with the unique ego node yield four mutually exclusive, collectively exhaustive types.}
\end{minipage}

\vspace{0.8cm}

\begin{minipage}{\textwidth}
\centering
\resizebox{\textwidth}{!}{%
\begin{tikzpicture}[
    >=Stealth,
    decision/.style={
        rectangle, rounded corners=3pt,
        draw=qcolor, thick, fill=qcolor!8,
        text width=3.2cm, align=center,
        font=\small\sffamily, minimum height=0.85cm,
    },
    leaf/.style={
        rectangle, rounded corners=3pt,
        draw=leafblue!80!black, thick, fill=leafblue!12,
        text width=3.0cm, align=center,
        font=\small\ttfamily\bfseries, minimum height=0.8cm,
    },
    blbl/.style={font=\footnotesize\sffamily\bfseries, text=qcolor, fill=white, inner sep=2pt},
    arr/.style={draw=qcolor!70, thick, -Stealth, rounded corners=4pt},
]

\node[decision]  at ( 5.0,  0.0)   (q1)  {Discrete (go/no-go)\\or continuous?};
\node[decision]  at (-1.0, -2.4)   (q2b) {Permits or\\prohibits entry?};
\node[decision]  at (11.0, -2.4)   (q2a) {Which control\\axis?};
\node[leaf]      at (-4.0, -4.8)   (gp)  {GRANTS\_PROCEED};
\node[leaf]      at ( 2.0, -4.8)   (ip)  {INHIBITS\_PROCEED};
\node[leaf]      at ( 8.0, -4.8)   (cl)  {CONSTRAINS\_\\LONGITUDINAL};
\node[leaf]      at (14.0, -4.8)   (cla) {CONSTRAINS\_\\LATERAL};

\draw[arr] (q1.south) -- ++(0,-0.65) -| (q2b.north)
    node[blbl, pos=0.25, left] {Discrete};
\draw[arr] (q1.south) -- ++(0,-0.65) -| (q2a.north)
    node[blbl, pos=0.25, right] {Continuous};
\draw[arr] (q2b.south) -- ++(0,-0.65) -| (gp.north)
    node[blbl, pos=0.25, left] {Permits};
\draw[arr] (q2b.south) -- ++(0,-0.65) -| (ip.north)
    node[blbl, pos=0.25, right] {Prohibits};
\draw[arr] (q2a.south) -- ++(0,-0.65) -| (cl.north)
    node[blbl, pos=0.25, left] {Speed};
\draw[arr] (q2a.south) -- ++(0,-0.65) -| (cla.north)
    node[blbl, pos=0.25, right] {Heading};

\end{tikzpicture}%
}
\smallskip

{\small\textbf{(b)} Edge-type decision tree. Two binary properties, discrete vs.\ continuous control and axis or polarity, yield four mutually exclusive, collectively exhaustive types.}
\end{minipage}

\caption{Both taxonomies are derived from binary decision trees over objectively testable properties. Node types split on agency and rule imposition (after the unique ego split). Edge types split on discrete vs.\ continuous control and axis or polarity. Each tree is exhaustive and produces mutually exclusive leaves, ensuring both taxonomies are \emph{closed by construction}, not by enumeration of driving scenarios.}
\label{fig:taxonomy_trees}
\end{figure}

\paragraph{Inter-node edges.} The edge set $\mathcal{E}_{\text{inter}} \subseteq (\mathcal{V} \setminus \{\text{ego}\}) \times (\mathcal{V} \setminus \{\text{ego}\})$ encodes causal relationships between non-ego entities and has two relation types. \texttt{AFFECTS} is directed and denotes an observable causal dependence: a traffic signal governing a lead vehicle's behaviour, or a parked vehicle occluding a pedestrian. \texttt{CONFOUNDED\_BY} is bi-directed and denotes an unobserved common cause: two vehicles simultaneously braking for a hazard outside the camera frame, for instance. The confounding edge makes $\cG$ semi-Markovian: unobserved confounding relation is explicitly represented rather than assumed away, which is essential for counterfactual queries in scenes where multiple agents respond to off-camera events. For \texttt{CONFOUNDED\_BY} edges, we additionally record an \texttt{inferred\_cause} attribute naming the unobserved variable (e.g., ``unseen hazard ahead''). Each inter-node edge justification must cite the $x$-positions of both endpoints from the scene metadata. These edges are what enable chain, fork, and collider graph structures referenced in \cref{subsec:dataset_statistics}, with \texttt{CONFOUNDED\_BY} specifically encoding the semi-Markovian fork structure that text-only causal reasoning cannot assess.

\paragraph{Distractors.} The distractor set $\mathcal{D}$ contains entities that are visible in the scene but have no ego edges and no plausible path to one within a single decision horizon. Each distractor $d \in \mathcal{D}$ carries a type, position relative to ego, and a \emph{rejection reason} grounded in observable scene geometry: behind a physical barrier, d-separated from ego by a blocked path, on the far sidewalk with no crossing intent, and so on. Distractors are retained rather than discarded because the QA pipeline uses them to generate the causal-irrelevance and discrimination question families (\cref{tab:question_taxonomy}), which test whether a model can distinguish visible-but-irrelevant entities from causally active ones.

\paragraph{Ego state.} The graph records the ego vehicle's current state $\mathbf{S} = (s, a, h)$: speed $s$, inferred acceleration $a \in \{\text{accelerating}, \text{decelerating}, \text{constant}\}$, and inferred heading change $h \in \{\text{straight}, \text{left}, \text{right}\}$. Ego state is the target of every ego edge's \emph{affected-state} field and the conditioning variable for all counterfactual trajectory queries.

\paragraph{Scene context.} Each scene record carries three free-text metadata fields alongside the graph: \emph{road geometry} (lane count and markings, intersection layout, road-edge features), the \emph{navigation command} (ego's intended manoeuvre, e.g.\ ``proceed straight through the intersection'' or ``change lanes left''), and \emph{environment conditions} (e.g.\ ``clear daytime'', ``rainy night''). Although listed first in the graph record for readability, they conceptually wrap the graph: the navigation command fixes the policy under which causal status is evaluated, so a vehicle to the left may be dormant under ``proceed straight'' but causally active under ``change lanes left''. Road geometry and environment conditions are retained as auxiliary context for the QA generator (\cref{subsec:pipeline}) and for stratified evaluation across scene types.

\subsection{Causal Status: Active, Dormant, and Distractor}
\label{app:causal_status}

CausalDriveBench partitions visible scene entities into three causal statuses: those that shape ego's current behavior, those that could shape it under a single physically realizable scene change, and those that cannot and are merely visible. We ground this partition in Pearl's interventional semantics \citep{bareinboim2022pearl}, \citep{pearl2009causal}: the \emph{active} criterion encodes the claim that removing the entity, formally applying the intervention $\mathrm{do}(X = \text{absent})$, would alter ego's state. The \emph{dormant} and \emph{distractor} criteria are obtained by relaxing this test along time and negating it, respectively.

\paragraph{Sustaining vs.\ triggering subtypes.} For question generation we further label each active edge by \emph{when} its influence is exerted. Let $\mathcal{G}^{-X, t_0}_{t_0}$ denote the graph in which $X$ is removed only at the instant $t_0$. Then
\begin{align}
    \mathrm{sustaining}(e) \;&\iff\; \mathrm{active}(e) \;\wedge\; S(\mathcal{G}_{t_0}) \neq S\!\left(\mathcal{G}^{-X, t_0}_{t_0}\right), \\
    \mathrm{triggering}(e) \;&\iff\; \mathrm{active}(e) \;\wedge\; S(\mathcal{G}_{t_0}) = S\!\left(\mathcal{G}^{-X, t_0}_{t_0}\right).
\end{align}
A sustaining entity is a live constraint at $t_0$: removing it now would change what ego does next, e.g.\ a lead vehicle ego is currently following, or a parked truck whose presence is forcing a sustained lateral offset. A triggering entity has already done its causal work: removing it at $t_0$ alone leaves ego on its current trajectory, but its earlier presence during the window initiated ego's committed maneuver, e.g.\ a green signal ego has just passed through, or an earlier turn-arrow ego has already begun executing. The distinction does not affect graph inclusion (both subtypes are active) but determines the temporal framing of downstream questions; questions at any rung of PCH can target either subtype.

\paragraph{Behavioral evidence.} The active criterion is evaluated through concrete behavioral evidence rather than against the theoretical space of motions ego could perform. We check three response channels at $t_0$ and across the window: \emph{longitudinal} (is ego's speed or acceleration profile shaped by $X$?), \emph{lateral} (is ego holding a position offset or steering pattern because of $X$?), and \emph{path} (would ego enter a qualitatively different region of road if $X$ were absent throughout the window?). Lateral response includes sustained position offset, not only active steering: an ego vehicle holding a displacement from lane center to clear an adjacent parked truck is responding to that truck even with the wheel centered. If all three channels return ``no change,'' the active criterion fails.

\paragraph{Operational restriction on $\mathcal{R}_\delta$.} Without further constraint, $\mathcal{R}_\delta(X)$ admits any kinematically reachable state, which would expand the dormant set combinatorially across every visible vehicle and pedestrian. We therefore restrict $\mathcal{R}_\delta(X)$ to transitions arising from the entity's own agency, applied only to entities of type \texttt{OTHER\_AGENT} that exhibit observable intent toward ego's path. Three exclusions follow. First, speculative mechanical events on otherwise static entities, such as a door opening on a parked car, cargo falling from a truck bed, or a tire blowout, are not transitions of the entity's agency and are excluded. Second, static agents with no observed momentum or intent signals are excluded: a stationary pedestrian near the crosswalk facing away from ego does not become dormant simply because they could in principle move. Third, agents whose trajectories have already passed ego's decision point are excluded: a cyclist that has just crossed in front of ego or a pedestrian who has finished crossing is not dormant despite recent proximity. Concretely, a pedestrian at the curb with body orientation toward the road and forward motion qualifies; a pedestrian walking parallel to the road on the sidewalk does not, even when adjacent to a crosswalk; a parked car on the shoulder does not, regardless of door state. This restriction trades a small amount of theoretical coverage for a tractable, behaviorally grounded dormant set.

In addition to agency-based exclusion, three geometric conditions must hold jointly: (i)~\emph{single-step reachability}, $X$ can reach a causally relevant state through one realizable action within $\delta$; (ii)~\emph{no physical barrier}, an unobstructed path exists between $X$ and ego's road at $t_0$; and (iii)~\emph{spatial relevance}, $X$'s current position is in or near ego's lane, or $X$'s recent trajectory shows convergence toward ego's path. A pedestrian on a fenced sidewalk 15\,m from ego's lane fails (ii) and (iii) regardless of intent signals.

\label{app:causal_graph}

\section{Causal QA Taxonomy}
\label{app:qa_taxonomy}
\begin{table}[h!]
  \caption{Question types in CausalDriveBench. Types marked $\dagger$ are novel contributions introduced in this work; the remaining 12 instantiate established CaLM causal scenarios in the embodied driving setting. Format codes: B (binary), M (multiple-choice), M+B (both formats supported).}
  \label{tab:question_taxonomy}
  \centering
  \small
  \begin{tabular}{l c c p{8.4cm}}
    \toprule
    Type & Rung & Format & Description \\
    \midrule
    \multicolumn{4}{l}{\textit{Active-element types: CaLM causal scenarios over active subgraphs}} \\
    CaI & 0 & M    & \textbf{Causal Identification.} Identify which visible element is currently constraining ego. \\
    CA  & 0 & M+B  & \textbf{Causal Attribution.} Trace a chain of causal influence from a distal element through a mediator to ego. \\
    CB  & 1 & B    & \textbf{Collider Bias.} Apply explaining-away: when two independent causes share ego as a collider, learning one reduces the necessity of the other. \\
    CoI & 2 & M    & \textbf{Confounder Identification.} Identify a shared cause that links multiple active influences on ego. \\
    BAS & 2 & M    & \textbf{Backdoor Adjustment Set.} Identify the complete set of variables to control for when isolating one element's effect on ego. \\
    CDE & 2 & M+B  & \textbf{Controlled Direct Effect.} Reason about an element's direct effect on ego when its mediator is held fixed. \\
    CR  & 3 & B    & \textbf{Counterfactual Reasoning.} Given ego's observed behavior, predict whether that behavior would persist had an active element been in a different state. \\
    SC  & 3 & B    & \textbf{Sufficient Cause.} Determine whether a single element, in isolation from all other active influences, is sufficient to produce ego's observed behavior. \\
    NC  & 3 & B    & \textbf{Necessary Cause .} Determine whether removing a single element would change ego's observed behavior, applying redundancy checking against other concurrent causes. \\
    NIE & 3 & M+B  & \textbf{Natural Indirect Effect.} Decompose how an element's effect on ego propagates through a mediator under that mediator's natural response. \\
    NDE & 3 & M+B  & \textbf{Natural Direct Effect.} Decompose an element's residual direct effect on ego when the mediator is held at its natural value. \\
    \midrule
    \multicolumn{4}{l}{\textit{Dormant-element types}$^\dagger$: \textit{visible entities one realizable transition from constraining ego}} \\
    DQ  & 0 & M+B  & \textbf{Discrimination Questions.} Distinguish a dormant entity that looks active from one that is genuinely active, including the empty-graph fallback when no active entity exists. Subtypes: Binary, Contrast, Empty. \\
    EL  & 1 & M+B  & \textbf{Explanation of Latency.} Explain why a visible entity is not currently constraining ego: shielding by an active element, structural decoupling, or shared cause without mutual influence. Subtypes: Shield, Decouple, Conditional. \\
    AB  & 2 & M+B  & \textbf{Activation Boundary.} Identify a single physically realizable change to the dormant entity, to ego's intent, or to a third element that would activate the dormant edge. Subtypes: Element, Ego, Context. \\
    \midrule
    \multicolumn{4}{l}{\textit{Distractor-element types}$^\dagger$: \textit{visible entities outside the causal graph}} \\
    DR & 0 & M  & \textbf{Distractor Rejection.} Multiple-choice fallback used when both the active and dormant sets are empty: the correct answer is that no visible element is currently affecting ego. \\
    CI  & 1 & B    & \textbf{Causal Irrelevance.} Reject a false attribution of ego's current behavior to a perceptually salient but causally irrelevant entity. \\
    NI  & 2 & B    & \textbf{Null Intervention.} Recognize that removing a distractor produces zero change in ego's behavior, $P(Y_{\mathrm{ego}} \mid \mathrm{do}(\text{remove}\,X)) = P(Y_{\mathrm{ego}})$. \\
    WC  & 3 & B    & \textbf{Wild Counterfactual.} Reason about a hypothetical world where the distractor's structural function changes; the answer depends on scene-specific geometry rather than generic policy. \\
    \bottomrule
  \end{tabular}
\end{table}

\subsection{Active-Element Question Types}
\label{app:active_qa}

Eleven question types are generated from active-edge subgraphs, each a CaLM causal scenario instantiated over four PCH layers (causal discovery and the three rungs of association, intervention, and counterfactual) and four graph structures (direct, chain, confounding, collider). Table~\ref{tab:question_taxonomy} gives the full mapping; the same type may appear in multiple cells when the underlying scenario applies to several graph structures (e.g., CDE on chain and confounding), so the count of twelve refers to types rather than table cells.

The mapping from PCH layer to CaLM scenario is as follows. \textbf{Causal discovery (Rung 0):} \emph{CaI} (causal identification) instantiates pairwise causal discovery; \emph{CA} (causal attribution) instantiates causal chain identification. \textbf{Association (Rung 1):} \emph{CB} (collider bias) instantiates explaining away. \textbf{Intervention (Rung 2):} \emph{CoI} (confounder identification) instantiates confounding identification; \emph{BAS} instantiates the backdoor adjustment set; \emph{CDE} instantiates the controlled direct effect. \textbf{Counterfactual (Rung 3):} \emph{CR} instantiates counterfactual reasoning; \emph{SC} and \emph{NC} instantiate the sufficiency condition (SC) and necessity condition (NC), respectively; \emph{NIE} and \emph{NDE} instantiate the natural indirect and natural direct effects.

Four types (CA, CDE, NIE, NDE) support both multiple-choice and binary answer formats. The binary variant produces a trajectory-testable yes/no answer: the downstream pipeline generates a counterfactual trajectory under the specified intervention and compares it against the observed one, providing an action-level evaluation channel that complements the language-level one.

\begin{table}[h]
  \caption{Active question type taxonomy in CausalDriveBench mapped to causal graph structures and Pearl's rungs of inference. Empty cells indicate the graph structure does not produce a non-trivial question at that rung. Example, a direct graph has no mediator for Rung-2 mediation}
  \label{tab:active_graph_map}
  \footnotesize
  \setlength{\tabcolsep}{4pt}
  \renewcommand{\arraystretch}{1.2}
  \centering
  \begin{tabular}{lllll}
    \toprule
    \textbf{Structure} & \textbf{Discovery (R0)} & \textbf{Association (R1)} & \textbf{Intervention (R2)} & \textbf{Counterfactual (R3)} \\
    \midrule
    Direct ($A \rightarrow$ ego) & CaI & -- & -- & CR, SC \\
    Chain ($A \rightarrow B \rightarrow$ ego) & CA & -- & CDE & NIE, NDE \\
    Confounding ($A \rightarrow$ \{$B \rightarrow$ ego\}) & CaI & -- & CoI, BAS, CDE & CR, NIE, NDE, SC \\
    Collider ($A \rightarrow$ ego $\leftarrow B$) & -- & CB & -- & NC \\
    \bottomrule
  \end{tabular}
\end{table}

Empty cells are principled: they reflect cases where a given graph structure does not produce a non-trivial question at that rung. For instance, direct graphs have no mediator for Rung~2 mediation questions, and collider structures are observationally invisible at Rung~0.

\paragraph{Isolation safety.} Not every question type can be answered from a single causal substructure considered alone. We distinguish \emph{isolation-safe} types, whose answers are correct regardless of other concurrent active influences (e.g., CaI, CA), from \emph{full-graph-conditioned} types, whose answers depend on the full set of active causes at $t_0$ (e.g., CR, SC, NC, BAS, CoI). For the latter, both the question preamble and the ground-truth answer must reference all active influences. A CR question of the form ``would you still be driving this way if the light turned green?'' must account for whether a lead vehicle redundantly constrains ego's speed, since in that case the answer is ``yes'' despite the light change. The downstream generation pipeline propagates interventions along directed inter-node edges (descendant propagation) and freezes non-descendants and \texttt{CONFOUNDED\_BY}-linked entities at observed values, ensuring that ground-truth answers respect the semi-Markovian structure of the graph.

\paragraph{Sustaining vs.\ triggering counterfactual frames.} Active edges are sub-classified as sustaining or triggering (\cref{app:causal_status}). This distinction determines the counterfactual frame for CR, SC, and NC questions. For sustaining elements, the counterfactual is interventional at $t_0$: ``you are currently \emph{[observed action]}; had this element been \emph{[different state]}, would you still be doing this?'' For triggering elements, the counterfactual is retrospective over the observation window: ``had this element never appeared during your approach, would you be in your current committed state?'' Applying the wrong frame produces incorrect ground truth: a triggering element typically shows no behavioral change at $t_0$ because its live constraint has expired, yet remains causally necessary for ego's current committed maneuver. For example, a green signal ego has just passed is triggering; the correct CR question is whether ego would have entered the intersection had the signal been red 2s ago, not whether ego's instantaneous state at $t_0$ depends on the (already-passed) signal.

\subsection{Dormant-Element Question Types}
\label{app:dormant_qa}

A dormant element has no active ego edge by definition and therefore cannot occupy any role in a direct, chain, confounding, or collider graph. The active-element question types are accordingly inapplicable. To evaluate reasoning about dormant entities, we introduce three categories spanning Rungs 0--2: \emph{discrimination} (DQ), \emph{explanation} (EL), and \emph{activation} (AB). These extend CaLM's causal scenarios to the visual-embodied setting: DQ tests sub-associational classification (can the model distinguish causal from non-causal status from a camera image alone?), EL tests conditional reasoning over the observed graph (can the model name what blocks or shields a potential causal path?), and AB tests interventional activation (a $\mathrm{do}$-query over a single-step modification of the scene). Table~\ref{tab:question_taxonomy} gives the full taxonomy with subtypes and formats.Each rung requires strictly more reasoning capability than the one below. 

\paragraph{DQ (Rung 0)} DQ requires scene observation: classifying whether a visible element is currently constraining ego or merely visible, with binary questions reserved for elements that look active from the camera images (perspective, size, or apparent motion that suggests interaction). 

\paragraph{EL (Rung 1)} EL requires conditional inference: explaining why a visible element is not currently constraining ego, attributing the latency to one of three mechanisms. \emph{Shielding} occurs when an active element blocks the dormant element's influence (e.g., a red light makes a crossing vehicle irrelevant); \emph{decoupling} reflects a structural absence of causal path (the entity is behind ego, on a different road, or behind a physical barrier); and \emph{correlation without causation} arises when ego and the dormant element share an observable property due to a common cause rather than mutual influence (both are stopped because of the same red light). For confounded pairs, the explanation cites the unobserved common cause recorded in the \texttt{CONFOUNDED\_BY} edge's \texttt{inferred\_cause} field. 

\paragraph{AB (Rung 2)} AB requires interventional reasoning: identifying a single physically realizable change to the element, to ego, or to a third entity that would activate the dormant edge within one decision horizon ($\delta = 0.5\,\mathrm{s}$). Activation premises must be physically plausible within this bound (e.g., a pedestrian stepping off a curb, a vehicle beginning to drift), excluding implausible transitions such as a stationary vehicle accelerating to highway speed.


\paragraph{Why no Rung 3 for dormant elements.} The exclusion follows from a logical constraint in the causal scenario definitions, not a design choice. CaLM's Rung 3 scenarios (probability of sufficiency PS, probability of necessity PN, and counterfactual reasoning CR) all require a factual outcome caused by the element. PS asks: given $X{=}x$ and $Y{=}y$, would $Y$ still be $y$ under $X{=}x'$? PN asks: given $X{=}x$ and $Y{=}y$, would $Y$ have been $y'$ under $X{=}x'$? Both condition on the factual pair $(X{=}x, Y{=}y)$. For a dormant element the pair does not exist: the element produced no outcome. PS and PN are undefined, and CR has no factual state to abduct from. Questions of the form ``had this dormant element activated, would your trajectory have changed?'' are forward-looking interventions, $\mathrm{do}(\text{activate})$ evaluated against the current scene, not retrospective counterfactuals. They sit at Rung 2 regardless of whether the answer depends on scene-specific geometry.

\subsection{Distractor-Element Question Types}
\label{app:distractor_qa}

Distractor elements lie outside the causal graph entirely: they have no ego edge, no inter-node edge, and no plausible single-step transition that would make either appear within one decision horizon. Yet they remain visually present (or were recently present) in the scene, and a robust VLA must reject them rather than infer spurious causal influence. We introduce four binary question types and one fallback multiple-choice type spanning Rungs 1--3 (\cref{tab:question_taxonomy}). The four binary types cover two orthogonal axes of causal rejection: \emph{factual rejection} (CI, NI) and \emph{counterfactual rejection} (WC). The DR fallback handles scenes in which both the active and dormant sets are empty.

The distractor types extend CaLM's causal scenarios in two directions that text-based benchmarks cannot address. First, CI and NI test the \emph{negative} case of causal reasoning: not ``what causes $Y$?'' but ``does $X$ \emph{not} cause $Y$ despite appearing in the scene?'' CI tests rejection at the associational level (visual co-occurrence does not entail causal influence); NI tests recognition that a removal intervention has zero downstream effect, $P(Y_{\text{ego}} \mid \mathrm{do}(\text{remove}\,X)) = P(Y_{\text{ego}})$. Second, WC test counterfactual reasoning under \emph{modified structural equations}: the element's behaviour is hypothetically replaced with one it never exhibited, requiring the model to construct an alternative SCM rather than perturb the observed one.

\paragraph{DR (Rung 0).} A large fraction of scenes contain neither active nor dormant elements, typically open-road segments with no immediate constraints on ego behaviour. To avoid generating no questions for these scenes, we add a single multiple-choice question whose correct answer is a natural-language ``none of these'' option (``my driving isn't being shaped by any of these'') and whose distractor options are the most visually salient elements from the \texttt{distractors} array. This is the only MCQ generated by the distractor pipeline and exists primarily as a coverage rather than a difficulty mechanism.

\paragraph{CI (Rung 1).} CI is restricted to high visual-salience distractors -- elements that look deceptively relevant due to size, motion, or perspective-induced apparent proximity despite physical separation. A truck on a parallel roadway visible in ego's left mirror is a CI candidate; a small sign on the far sidewalk is not. The reasoning must identify what \emph{is} causing the behaviour (a specific active element) and explain why the high-salience distractor is not, naming the perceptual cue that makes it look relevant.

\paragraph{NI (Rung 2).} NI tests the definitional property of distractors: removal produces no change in ego behaviour. The reasoning must trace through every active influence and confirm that none of them depend on the distractor. NI questions are optionally generated in \emph{contrast pairs} with an active-element removal question that uses the same phrasing template; the active version requires a Yes answer and the distractor version a No, so a model cannot exploit phrasing cues alone. We cap contrast pairs at two per scene to avoid stylistic repetition.

\paragraph{WC (Rung 3).} WC tests reasoning about an alternative world where the distractor's structural function changes. Unlike dormant AB, which is bounded to transitions within the $\delta = 0.5\,\mathrm{s}$ reachability horizon, WC premises may be implausible but must remain physically coherent: a parked vehicle can swerve, a pedestrian can run, a building cannot move. The deeper distinction from AB is that WC modifies the element's structural equation (it behaves in a fundamentally different way), whereas AB modifies the element's state within the existing SCM (it transitions to a reachable configuration). This is why WC is Rung 3 despite targeting a causally inert element: the counterfactual world has different causal mechanisms, not just different variable values. To force scene-specific reasoning, premises must be \emph{partial} (drift, encroachment) rather than total (full lane entry). A premise like ``had the parked car drifted half a lane toward you'' yields an answer that depends on ego's clearances and speed; ``had it pulled directly into your lane'' admits the generic answer that any obstacle requires braking, and is rejected by the generation pipeline.

\subsection{Answer Generation Principles}
\label{app:answer_gen}

Four principles govern ground-truth answer generation across all question types. They apply at distinct stages of evaluation: \emph{descendant propagation} determines which entities change under a counterfactual; \emph{redundancy checking} establishes whether removing a constraint actually unbinds ego given other concurrent influences; the \emph{behavioral impact test} translates the resulting graph into a concrete driving action; and \emph{rung verification} confirms that questions labeled as counterfactual genuinely require counterfactual reasoning rather than interventional reasoning in retrospective phrasing.

\paragraph{Descendant propagation.} When a counterfactual question modifies entity $X$, all causal descendants of $X$ reachable via directed \texttt{AFFECTS} edges are modified accordingly, while non-descendants are frozen at their observed values. Entities linked by \texttt{CONFOUNDED\_BY} edges are not descendants of either endpoint; the bidirected edge represents a shared unobserved cause, and intervening on one endpoint does not propagate to the other. For example, if a red light turns green, a lead vehicle governed by that light is released (descendant via directed \texttt{AFFECTS}), but a vehicle in the adjacent lane responding to an unseen signal remains frozen (linked only by \texttt{CONFOUNDED\_BY}, not a descendant). The frame in which the intervention is evaluated depends on the element's causal status (\cref{app:causal_status}): for active-sustaining elements the intervention is at $t_0$, while for active-triggering elements it is anchored to the observation window $[t_0 - \Delta, t_0]$.

\paragraph{Redundancy check.} In confounding structures, a common cause $C$ and a downstream element $B$ may both constrain ego concurrently. Removing $B$ in isolation may leave $C$ still binding, so the observed behavior would persist. But removing $C$ releases $B$ as well via descendant propagation, which can change behavior even when $B$'s direct effect is redundant. We therefore evaluate redundancy after applying descendant propagation: trace which entities remain active in the counterfactual graph, and check whether any of them independently produces the observed behavior. A CR question that ignores redundancy will assign the wrong answer for confounding structures roughly half the time, since either $C$-removal or $B$-removal cases will be misclassified depending on which path is overlooked.

\paragraph{Behavioral impact test.} Counterfactual questions ask whether driving \emph{behavior} would change, not whether the theoretical space of possible movements expands. Removing a non-binding constraint does not change behavior. We evaluate three response channels against ego's observed state: \emph{longitudinal} (is ego's speed or acceleration profile shaped by $X$?), \emph{lateral} (is ego holding a position offset or steering pattern because of $X$?), and \emph{path} (would ego enter a qualitatively different region of road in $X$'s absence?). A maintained lateral offset, e.g., holding two metres off lane center because of an adjacent parked truck, is a lateral response despite no active steering input; the test must recognize sustained position offsets as behavioral evidence. The output of the test is one of the concrete driving actions \{\textsc{stop, yield, lane-change, swerve, accelerate, decelerate, follow, avoid, no-change}\}; reasoning expressed in abstract terms (``the trajectory changes'') is rejected during validation. If all three channels return ``no change,'' the answer is \textsc{no-change} regardless of how the possibility space has expanded.

\paragraph{Rung verification.} Counterfactual-sounding questions are labeled Rung 3 only when the observed ego behavior is a load-bearing premise of the answer. We apply two tests in sequence. The \emph{load-bearing test} removes the observed ego behavior from the question and asks whether the question can still be answered; if yes, the question is interventional regardless of retrospective phrasing and is relabeled to Rung 2. The \emph{scene-specificity test} asks whether the answer would be identical in any other scene with the same element types and the same counterfactual premise; if yes, the answer follows from generic driving rules and is relabeled to Rung 2. Genuine Rung 3 answers require both a load-bearing observed outcome and abduction from scene-specific geometry, distances, or speeds. This verification applies to active-element counterfactuals (CR, SC, NC, and Rung 3 instantiations of NIE and NDE) and distractor wild counterfactuals (WC). Dormant questions are capped at Rung 2 by the CaLM-defined undefinedness of PS and PN for entities that produced no factual outcome (\cref{app:dormant_qa}).

\section{Extended Dataset Statistics}  %
\label{app:dataset}
\subsection{Dataset Composition}

\begin{figure*}[t]
\centering
\includegraphics[width=\textwidth]{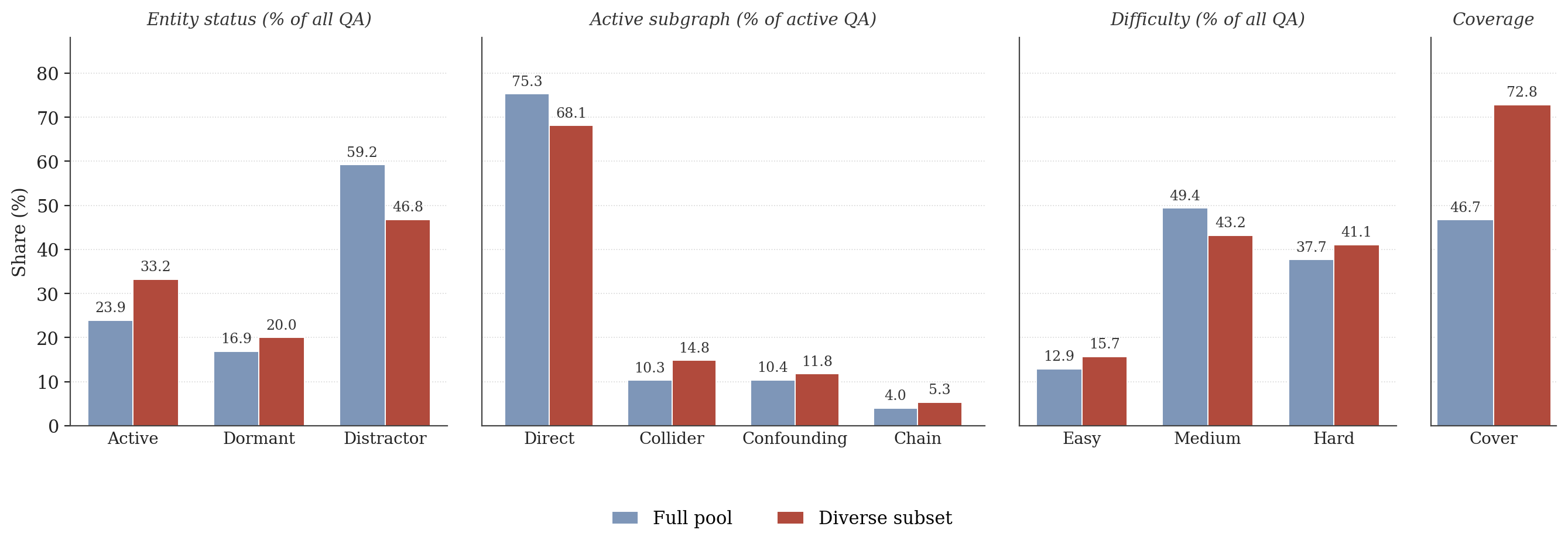}
\caption{Effect of Stage 2 diversity sampling on benchmark composition.}
\label{fig:full_vs_diverse}
\end{figure*}

CausalDriveBench is constructed from nuScenes~\cite{caesar2020nuscenes}, which provides 850 trainval scenes of 20\,s duration each, recorded in dense urban environments in Boston and Singapore with multi-sensor annotations. From each scene we extract independent reasoning samples by anchoring at evenly spaced timestamps and capturing four keyframes per sample over a 2\,s window (at $T = -1.5, -1.0, -0.5, 0\,\mathrm{s}$). Anchors are spaced 6\,s apart along each scene timeline, yielding roughly four samples per scene and 3,400 samples in total. We discard 211 samples that contain no agents within the 30\,m vehicle and 20\,m VRU radii, or whose ego behaviour cannot be attributed to any observable scene entity (a parked ego with no occupant in view, for instance). The remaining 3,189 samples are processed by the extraction pipeline (\ref{subsec:pipeline}) and verified by human reviewers (\cref{subsec:human_review}). After two-level hierarchical clustering (Stage 2) we obtain 815 second-level groups; the highest-richness sample from each group, scored by a weighted combination of causal-node count, active-edge count, inter-node-edge count, and per-category rarity, is retained as the cluster's representative. This leader-selection sampling produces a diverse subset of 815 samples that preserve coverage of every question category present in the source pool while concentrating on causally richer scenes (\cref{fig:full_vs_diverse}). The full benchmark contains 7,285 question-answer pairs across these 815 samples, with mean per-scene yields of 4.08 active, 3.34 dormant, and 4.23 distractor questions on the scenes where each pipeline produced output. We also generate the counterfactual trajectories for 1,177 questions across rung-2 and rung-3 which when filtered for collision-free and road layout-following paths yield 1,000 final counterfactual trajectories. 

\providecolor{cAct1}{HTML}{B85450}  
\providecolor{cAct2}{HTML}{D08770}  
\providecolor{cAct3}{HTML}{E5B582}  
\providecolor{cAct4}{HTML}{8B5A3C}  

\providecolor{cDor1}{HTML}{5F8B7A}  
\providecolor{cDor2}{HTML}{7FA89B}  
\providecolor{cDor3}{HTML}{A2C4B8}  

\providecolor{cDis1}{HTML}{4A6FA5}  
\providecolor{cDis2}{HTML}{6B8AB8}  
\providecolor{cDis3}{HTML}{8FA8C7}  
\providecolor{cDis4}{HTML}{6F5B8A}  

\providecolor{cR0}{HTML}{5A6B7B}    
\providecolor{cR1}{HTML}{7090A8}    
\providecolor{cR2}{HTML}{C89668}    
\providecolor{cR3}{HTML}{B85450}    

\providecolor{cEasy}{HTML}{5F8B7A}    
\providecolor{cMedium}{HTML}{C89668}  
\providecolor{cHard}{HTML}{B85450}    

\tikzset{sliceedge/.style={draw=white, line width=0.6pt}}

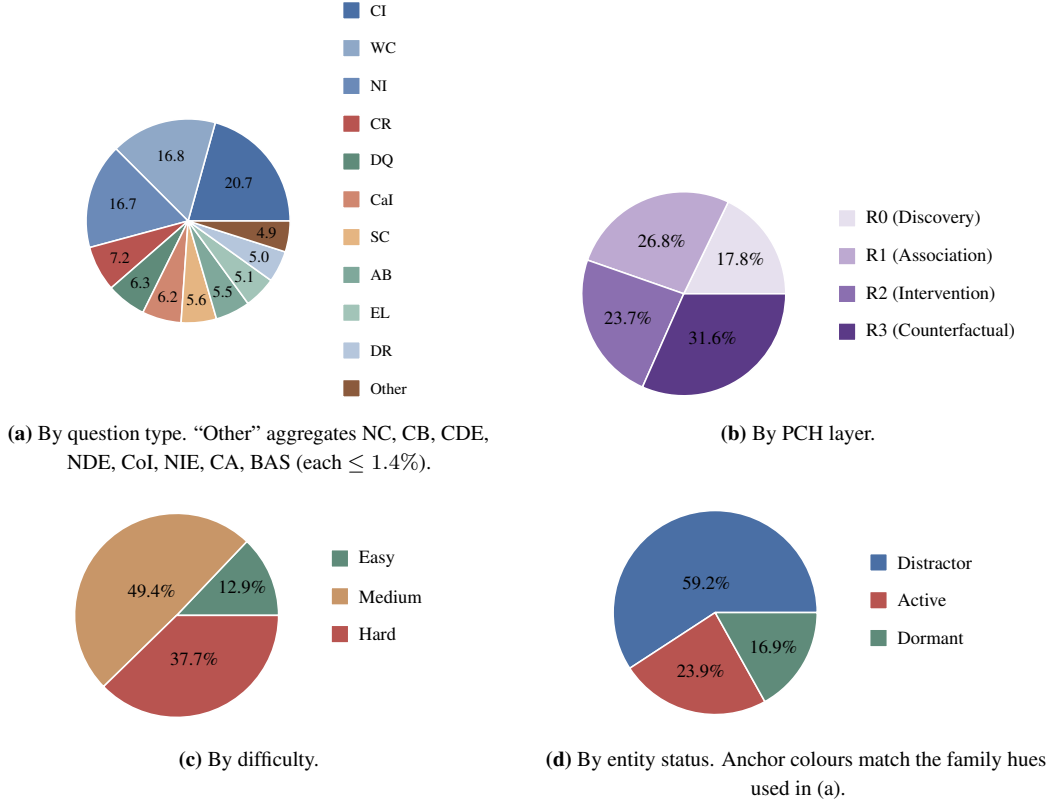
\begin{figure*}[t]
\centering

\begin{minipage}[t]{0.48\linewidth}
\centering
\begin{tikzpicture}[font=\tiny]
\pie[
    radius=1.35,
    text=legend,
    color={cDis1, cDis3, cDis2, cAct1, cDor1, cAct2, cAct3, cDor2, cDor3, cDis4, cAct4},
    sum=auto,
    after number=,
    style=sliceedge,
]{
    20.7/CI,
    16.8/WC,
    16.7/NI,
    7.2/CR,
    6.3/DQ,
    6.2/CaI,
    5.6/SC,
    5.5/AB,
    5.1/EL,
    5.0/DR,
    4.9/Other
}
\end{tikzpicture}
\\[3pt]
{\footnotesize\textbf{(a)} By question type. ``Other'' aggregates NC, CB, CDE, NDE, CoI, NIE, CA, BAS (each $\leq 1.4\%$).}
\end{minipage}%
\hfill
\begin{minipage}[t]{0.48\linewidth}
\centering
\begin{tikzpicture}[font=\scriptsize]
\pie[
    radius=1.35,
    text=legend,
    color={cR0, cR1, cR2, cR3},
    sum=auto,
    after number=\%,
    style=sliceedge,
]{
    17.8/R0 (Discovery),
    26.8/R1 (Association),
    23.7/R2 (Intervention),
    31.6/R3 (Counterfactual)
}
\end{tikzpicture}
\\[3pt]
{\footnotesize\textbf{(b)} By PCH layer.}
\end{minipage}

\vspace{10pt}

\begin{minipage}[t]{0.48\linewidth}
\centering
\begin{tikzpicture}[font=\scriptsize]
\pie[
    radius=1.35,
    text=legend,
    color={cEasy, cMedium, cHard},
    sum=auto,
    after number=\%,
    style=sliceedge,
]{
    12.9/Easy,
    49.4/Medium,
    37.7/Hard
}
\end{tikzpicture}
\\[3pt]
{\footnotesize\textbf{(c)} By difficulty.}
\end{minipage}%
\hfill
\begin{minipage}[t]{0.48\linewidth}
\centering
\begin{tikzpicture}[font=\scriptsize]
\pie[
    radius=1.35,
    text=legend,
    color={cDis1, cAct1, cDor1},
    sum=auto,
    after number=\%,
    style=sliceedge,
]{
    59.2/Distractor,
    23.9/Active,
    16.9/Dormant
}
\end{tikzpicture}
\\[3pt]
{\footnotesize\textbf{(d)} By entity status. Anchor colours match the family hues used in (a).}
\end{minipage}

\caption{Distribution of QA pairs in CausalDriveBench across the full unfiltered set of 3{,}189 verified scenes (22{,}952 QA pairs), shown by (a) question type, (b) PCH layer, (c) difficulty, and (d) entity status. In (a), warm hues mark active-element types, sages mark dormant types, and blues/purples mark distractor types; (d) gives the per-family aggregate using each family's anchor hue. Compared to the diversity-sampled subset (\cref{fig:qa_distributions_diverse}), the per-category and per-rung shares are nearly identical, indicating that the proportional sampling at rate 0.5 within Stage 2 clusters preserves the question-type composition of the source pool while reducing it to 1{,}887 representative scenes.}
\label{fig:qa_distributions_full}
\end{figure*}

\subsection{Need for clustering and sampling}
\label{sec:qt_distribution}

Figure~\ref{fig:qa_distributions_full}(a) shows the QA-pair share per question type and \cref{fig:full_vs_diverse} reports the underlying counts for all 3189 samples resulting in 22,952 QA pairs. In \cref{fig:full_vs_diverse} each bar pair shows the
share of the relevant denominator: Entity status and Difficulty are normalised against all QA pairs;
Active subgraph is normalised against active-element QA only (so each pool sums to 100
that facet); Coverage reports the percentage of scenes in which at least one active-element question
was generated. Clustering followed by leader sampling retains 25.6\% of source
scenes while shifting the benchmark in three reviewer-relevant directions: (i) toward causally engaged
scenes (Active share +9.3 pp, Distractor share -12.4 pp), (ii) toward rarer causal subgraphs (Collider, Confounding, and Chain each grow by 1.2 to 1.5× at the expense of the abundant Direct structure),
and (iii) toward harder questions (Hard share +3.4 pp). Active-scene coverage rises from 46.7\% to
72.8\%, indicating that the scenes dropped during clustering were predominantly those without rich
active-element content.

The diverse QA distribution across rungs (\cref{fig:qa_distributions_diverse} reflects the structure of the underlying causal graphs and the per-scene generation budget. Rung 3 dominates (33.5\%) because three of the most prolific question types are R3 by construction: counterfactual reasoning (CR) and wild counterfactual (WC) account for over 1,600 questions between them, and sufficient cause (SC) adds another 510. Rung 1 is the next largest (25.6\%) because causal irrelevance (CI), the highest-frequency distractor type at 1,267 questions, sits at R1. Rung 2 (23.9\%) is anchored by null intervention (NI) and dormant activation (AB). Rung 0 is the smallest (17.0\%) because only five categories contribute to it (CaI, CA, AR, DQ, DR-Empty), and CaI and CA are gated by isolation safety and the active-element budget.


\subsection{Causal Graph Structure Distribution}
\label{sec:graph_distribution}

Active-element question types are licensed by the causal graph structures present in the scene's active subgraph. Table~\ref{tab:qa_subgraph_distribution} reports structure prevalence across the 2,418 active-element questions; dormant and distractor questions are not associated with any of the four canonical structures and are listed as ``no active structure'' in aggregate.

\begin{table}[h!]
  \caption{Distribution of QA pairs by underlying active-element graph structure. Dormant and distractor questions are aggregated as ``no active structure'' since they are generated outside the active-edge enumeration.}
  \label{tab:qa_subgraph_distribution}
  \centering
  \begin{tabular}{l r r}
    \toprule
    Graph Structure & QA pairs & \% of total \\
    \midrule
    Direct ($A \rightarrow$ ego)                     & 1{,}646 & 22.6 \\
    Collider ($A \rightarrow$ ego $\leftarrow B$)    & 359     & 4.9  \\
    Confounding ($C \rightarrow$ ego, $C \rightarrow B \rightarrow$ ego) & 286 & 3.9 \\
    Chain ($A \rightarrow B \rightarrow$ ego)        & 127     & 1.7  \\
    \midrule
    No active structure (dormant + distractor)       & 4{,}867 & 66.8 \\
    \midrule
    Total                                            & 7{,}285 & 100.0 \\
    \bottomrule
  \end{tabular}
\end{table}

Direct structures dominate the active set, since most scenes contain at least one entity with a direct active causal relationship to ego. Confounding structures appear in 286 questions despite the inter-node edge requirement, supported by the deterministic subgraph extractor's enforcement of the chain-versus-confounding subsumption rule (\cref{app:subgraph_enumeration}). Chains are the rarest of the four structures (1.7\% of all QA, 5.3\% within the active set) because pure chains require a distal node $A$ that is not itself an ego parent; many candidates that initially appear chain-like are reclassified as confounding by the extractor. Colliders, which require two genuinely independent active causes, appear in 359 questions; the four-test independence check (\cref{app:subgraph_enumeration}) is conservative by design, since most co-occurring active influences in driving scenes share a common cause. Compared to the unfiltered source pool, the diverse-sampled subset visibly amplifies the rare structures: Collider, Confounding, and Chain shares within the active subset all grow at the expense of Direct, as shown in Figure~\ref{fig:full_vs_diverse}.

\subsection{Active, Dormant, and Distractor Coverage}
\label{sec:active_dormant_coverage}

CausalDriveBench distinguishes three causal statuses for visible entities: \emph{active} entities currently constraining ego, \emph{dormant} entities one realisable transition from constraining ego, and \emph{distractor} entities outside the causal graph (\cref{fig:qa_distributions_diverse}). Across the 815 sampled scenes, 72.8\% have at least one active-element question, 53.4\% have at least one dormant-element question, and 98.9\% have at least one distractor-element question; the last figure reflects that almost every visible driving scene contains at least one perceptually salient but causally irrelevant entity. Distractors contribute 46.8\% of all QA pairs (3,412), active elements 33.2\% (2,418), and dormants 20.0\% (1,455). The dominance of distractor questions is a substantive feature of the benchmark rather than an extraction artefact: in real driving scenes, most visible entities are distractors at any given moment, and a benchmark whose evaluation surface reflects this proportion measures a model's ability to reject false causal attributions as well as to identify true ones. Diversity sampling visibly shifted this distribution toward causally engaged scenes; the unfiltered source pool of 3,189 scenes had only 23.9\% active QA share and 46.7\% active-scene coverage (Figure~\ref{fig:full_vs_diverse}).

\subsection{Format and Difficulty Distribution}
\label{sec:format_difficulty}

83.3\% of questions are binary (Yes/No) and 16.7\% are multiple-choice (Table~\ref{tab:format}). The binary skew follows from two design choices: the four high-volume types (CI, NI, WC, CR) are binary by construction, and dual-format types prefer the binary variant when the scene supports trajectory generation. By difficulty, 15.7\% of questions are easy (Rung 0), 43.2\% medium (Rung 1, isolation-safe Rung 2), and 41.1\% hard (full-graph-conditioned Rung 2 and Rung 3). The hard fraction is concentrated in active-element questions, of which 65.4\% are hard, compared with 30.2\% for dormant and 28.5\% for distractor questions.

\begin{table}[h!]
  \caption{Format and difficulty distribution.}
  \label{tab:format}
  \centering
  \begin{tabular}{l r r}
    \toprule
    Property & Count & \% \\
    \midrule
    \multicolumn{3}{l}{\textit{Format}} \\
    Binary                & 6{,}072 & 83.3 \\
    Multiple choice       & 1{,}213 & 16.7 \\
    \addlinespace
    \multicolumn{3}{l}{\textit{Difficulty}} \\
    Easy                  & 1{,}145 & 15.7 \\
    Medium                & 3{,}146 & 43.2 \\
    Hard                  & 2{,}994 & 41.1 \\
    \bottomrule
  \end{tabular}
\end{table}

\section{Benchmark Construction Details}  %
\label{app:pipeline}
\subsection{Causal scene graph extraction}
\label{app:causal_graph_extraction}


The extraction prompt operationalises Pearl's Structural Causal Model at a single timestep $T{=}0$. Each ego-edge encodes an interventional claim, that $\mathrm{do}(\text{remove entity})$ at $T{=}0$ would change ego's next action, and unobserved common causes are represented as bidirected \texttt{CONFOUNDED\_BY} edges, yielding a semi-Markovian graph. Entities are partitioned into four causal categories: \emph{active-sustaining} (live constraint at $T{=}0$), \emph{active-triggering} (initiated ego's current committed manoeuvre during $T{=}{-}1.5$ to $T{=}0$ but does not hold at $T{=}0$), \emph{dormant} (passes a single-step reachability test within a $0.5$\,s decision horizon), and \emph{distractor} (fails both active and dormant tests). Active-sustaining and active-triggering are temporal subtypes of active edges, orthogonal to the PCH layer at which a question targeting the edge can be posed.

The model is supplied with multi-camera footage across four observation timesteps, four rendered BEV semantic maps, a natural-language scene-state summary that aggregates kinematic descriptors per agent, and the ego's navigation command. Modality precedence is fixed: scene-state speed and motion trends are authoritative over camera estimates, cameras are authoritative for signal state and physical confirmation of intrusion, and BEV is authoritative for lane membership via explicit boundary tracing rather than coordinate proxies. Road geometry, environment conditions, and navigation intent are recorded as free-text fields in \texttt{scene\_context} rather than as nodes. A bidirectional BEV--camera cross-check is required before any edge is asserted, and several known failure modes (signal state inferred from ego behaviour, lane membership inferred from $y$-offset) are explicitly proscribed.

Extraction proceeds through ten decision trees that specify the full procedure. Tree~1 gates entity inclusion via camera observability and physical-barrier separation, with a walkway-adjacency exception for pedestrians near crosswalks. Tree~2 separates infrastructure from obstacles. Tree~3 assigns each agent to ego's lane, the opposing lane, or an adjacent same-direction lane through boundary tracing, with a dedicated bypass for crosswalk pedestrians that handles intent assessment and crosswalk-portion membership. Trees~4a, 4b, 5, 6, and~7 assign one of the four ego-edge types (\texttt{CONSTRAINS\_LONGITUDINAL}, \texttt{CONSTRAINS\_LATERAL}, \texttt{INHIBITS\_PROCEED}, \texttt{GRANTS\_PROCEED}) according to source type and lane relationship. Tree~8 then resolves causal status through a path-membership gate (\textsc{q\_path}), a behavioural-response check across the entire observation window, a redundancy rule that preserves both edges when two entities independently constrain ego, a retrospective active-triggering check, and the single-step reachability test for dormant candidates. Tree~9 enumerates inter-node edges across two \texttt{AFFECTS} layers (infrastructure to agent, and agent to agent or obstacle) plus \texttt{CONFOUNDED\_BY} for genuinely unobserved common causes, with explicit straddling and observed-cause gates that suppress spurious latent confounders. Tree~10 validates each edge against the schema.

The output is a JSON object containing scene context, ego state with a behaviour explanation grounded in observable entities, the node and edge sets, and a \texttt{distractors} array in which every excluded but camera-visible entity is logged with a rejection reason citing the specific tree question that failed and the spatial or kinematic evidence supporting that failure. This distractor record is essential downstream: it forms the candidate pool for distractor-type question generation and supplies the audit trail used by the automated reviewer. Several procedural safeguards run throughout, including an output-discipline rule that forbids emitting self-flagged invalid edges, a stopped-at-green reconciliation that forces resolution of unexplained ego behaviour, and the redundancy rule that prevents two genuinely independent active constraints from collapsing into one.

\subsection{Stage 2a: Subgraph enumeration}
\label{app:subgraph_enumeration}

The QA prompts generate questions over four canonical causal substructures (direct, chain, collider, observed and unobserved confounding). Identifying these structures inside the prompt is unreliable: the language model frequently miscounts overlapping pair-level configurations or fails to enforce subsumption rules between confounding and chain structures. We therefore extract subgraphs offline through a deterministic procedure that takes the scene graph as input and emits an authoritative subgraph manifest consumed by the prompts. The procedure resolves overlaps via a fixed priority rule: confounding takes precedence over collider at the pair level, while chain extraction is restricted to distal nodes that are not themselves ego parents (otherwise the structure is necessarily confounding rather than a pure chain). For each pair of active ego parents we test for an observed directed inter-node edge between them, which yields a confounding-observed structure with the upstream node as confounder and the downstream node as mediator; absent such an edge, we test for a \texttt{CONFOUNDED\_BY} link between the pair, which yields a confounding-unobserved structure encoding a shared latent cause. Pairs that survive both tests and whose endpoints play no confounder, mediator, or chain-mediator role anywhere in the graph are admitted as colliders, an intentionally conservative criterion that reflects the rarity of genuinely independent active influences in driving scenes. Direct structures are emitted per active ego edge and annotated with the source node's role under any subsuming confounding structure (independent, mediator, or confounder); the role determines which question types the QA generator may issue against the direct structure. A confounder role restricts the direct structure to Rung 0 identification (CaI) since CR and SC questions on a confounder require propagating the intervention through the mediator and therefore escalate to the confounding structure, while independent and mediator roles permit the full direct slate (CaI, CR, SC). The output of this stage is a per-scene subgraph manifest with stable structure identifiers of the subgraph that the QA prompts reference verbatim in question metadata, allowing every generated question to be traced back to the structure it was derived from.

\subsection{Stage 2b: Clustering and Diversity-driven scene selection.} 
\label{app:clustering}

Raw extraction yields a large pool of scene graphs from each source dataset, but a uniform random sample under-represents the rare causal configurations that drive evaluation difficulty (multi-hop chains, semi-Markovian forks, scenes with both dormant and distractor entities). To select a diverse benchmark subset we cluster scenes hierarchically over two complementary kernel families, then sample within clusters at a fixed rate. The first level groups scenes by causal structure: a convex combination $K_1 = w_a K_{\text{active}} + w_d K_{\text{dormant}} + w_{\text{dist}} K_{\text{distractor}} + w_{\text{topo}} K_{\text{topology}}$ with equal weights $w = 0.25$, where each component kernel is a weighted Jaccard similarity over a histogram of (effect type, source category) pairs for active edges, (effect type, source category, inter-link flag) tuples for dormant edges, source-category bins for distractors, and (subgraph pattern, source category) tuples enumerated by an offline subgraph extractor for topological structure. The second level refines each first-level cluster by scene context: $K_2$ is the mean of four Jina-embedding cosine similarities computed over road geometry, environment conditions, navigation intent, and ego-behaviour explanation. Both levels apply complete-link agglomerative clustering with a Kneedle-selected distance threshold, producing first-level \emph{families} (causally similar scenes) and second-level \emph{groups} (causally similar scenes that also share scene context). Within each second-level group we sample proportionally at a rate of $0.5$, retaining at least one scene per group and preferring scenes that diversify across nuScenes scene IDs over those that share the same scene with the current pick. Each scene is scored by a richness measure (counts of causal nodes, active edges, inter-node edges, dormant edges, weighted by structural rarity) and the highest-scoring eligible scene in each iteration is added until the per-group budget is met. The output of this stage is a manifest of selected scene graphs that flow into QA generation in Stage 3.

\begin{figure}[t]
\centering
\includegraphics[width=\linewidth]{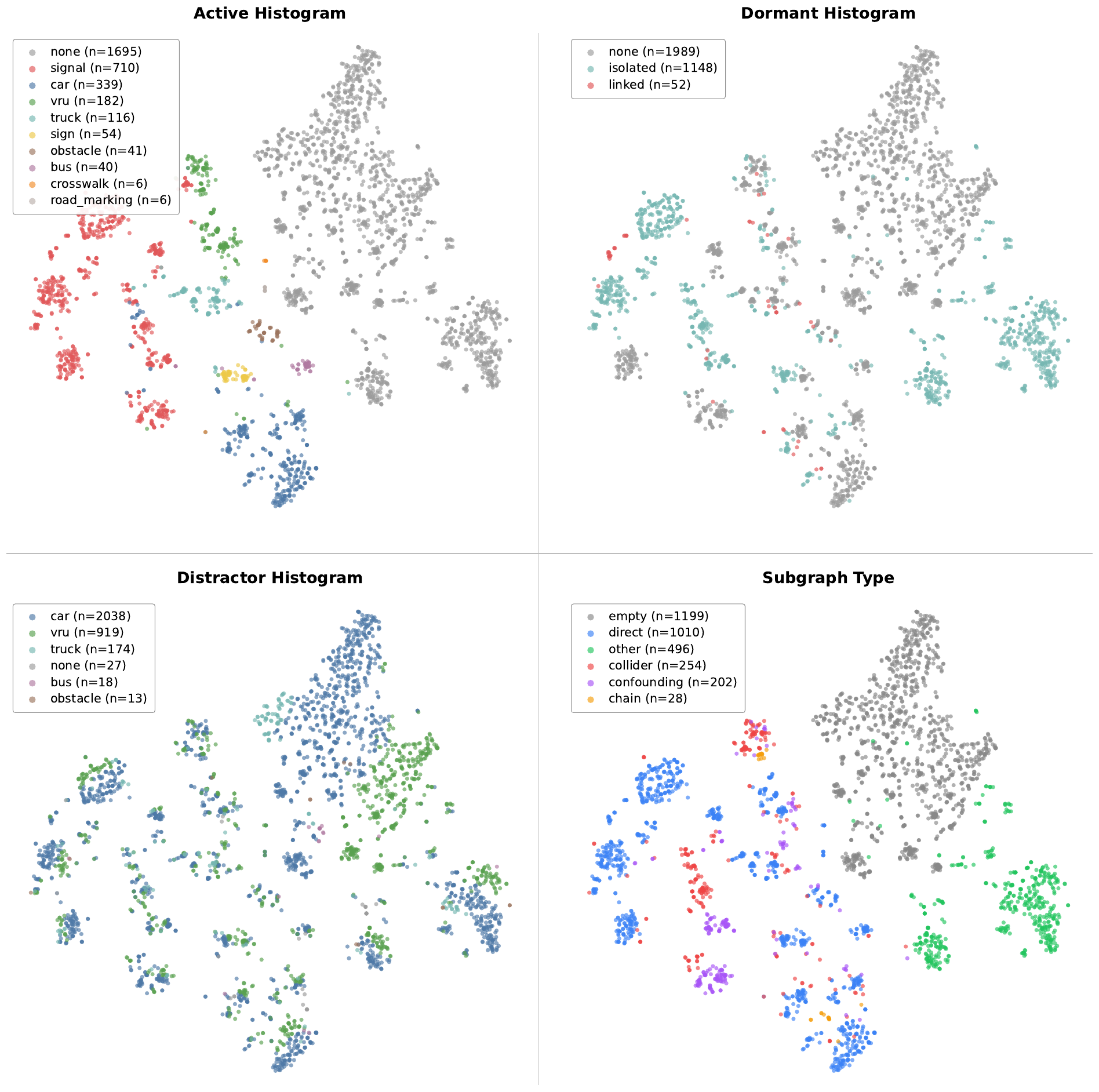}
\caption{Level-1 clustering kernels visualised on the joint embedding. Two-dimensional t-SNE of the combined hierarchical-clustering kernel $K_{\text{combined}} = 0.5,K_1 + 0.5,K_2$ over $n{=}3{,}189$ verified scene samples, recoloured by each of the four Level-1 categorical axes that the $K_1$ kernels hash on. (top-left) dominant active causal-source category (signal / vehicle / vulnerable road user / etc.); (top-right) dormant-edge link status -- whether the sample has any dormant ego-edge and, if so, whether it connects to an active cause via an inter-node edge (\emph{linked}) or not (\emph{isolated}); (bottom-left) dominant distractor category from the graph's rejected-cause list; (bottom-right) the dominant active-subgraph topology (empty / direct / chain / collider / confounding). The visible structure within and across panels indicates that the Level-1 axes capture complementary aspects of causal-graph diversity rather than redundant ones.}
\label{fig:subgraph_distribution}
\end{figure}

\begin{figure}[t]
\centering
\includegraphics[width=\linewidth]{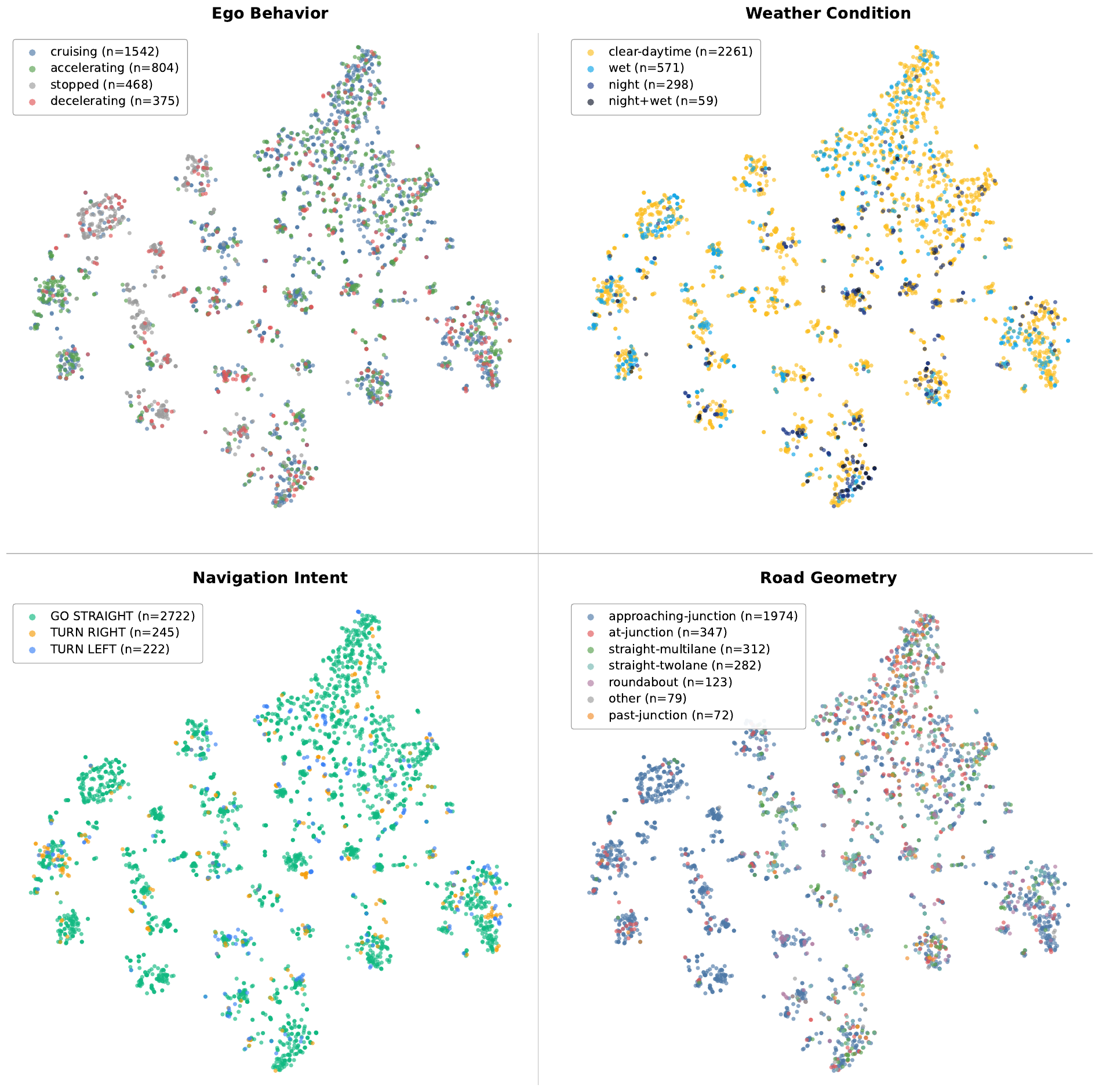}
\caption{\textbf{Level-2 clustering kernels visualised on the joint embedding.} Two-dimensional t-SNE of the combined hierarchical-clustering kernel $K_{\text{combined}} = 0.5\,K_1 + 0.5\,K_2$ over $n{=}3{,}189$ verified scene samples, recoloured by each of the four Level-2 categorical axes that the $K_2$ kernels embed. \textbf{(top-left)} ego speed-trajectory class derived from the per-sample \texttt{state.json} future trajectory (stopped / decelerating / cruising / accelerating); \textbf{(top-right)} weather condition parsed from the nuScenes \texttt{scene.description} field; \textbf{(bottom-left)} the ego's navigation intent at $T{=}0$ (\textsc{go straight}, \textsc{turn left}, \textsc{turn right}); \textbf{(bottom-right)} road-geometry context with junction temporal phase as the primary axis (approaching / at / past) and lane structure as the secondary axis. Together with Figure~\ref{fig:subgraph_distribution}, the panels show that scene-context diversity is largely orthogonal to the causal-graph structural diversity captured by the Level-1 kernels.}
\label{fig:scene_context_distribution}
\end{figure}

\subsection{Stage 3: Question-answer generation.} 
\label{app:qa_generation}

QA pairs are generated by three role-specific prompts that operate over the same scene graph but target disjoint entity sets. All prompts are added in the supplement material.

\begin{itemize}
  \item The \emph{active-element prompt} enumerates all graph structures (direct, chain, confounding, collider) over active edges, applies the four-test collider independence check to filter spurious colliders, and generates twelve question types instantiating CaLM causal scenarios across causal discovery and the three rungs of PCH (Table~\ref{tab:question_taxonomy}).
  \item The \emph{dormant-element prompt} classifies each dormant entity by activation plausibility (high, medium, low), latency mechanism (shielded, decoupled, correlated), and trajectory sensitivity (geometry-grounded vs.\ policy-level), then generates DQ, EL, and AB questions according to the eligibility gates in Table~\ref{tab:question_taxonomy}.
  \item The \emph{distractor-element prompt} classifies each distractor by visual salience and counterfactual coherence, then generates CI, NI, and WC questions, plus a DR multiple-choice fallback for scenes whose active and dormant sets are both empty.
\end{itemize}

\noindent All three prompts share four discipline rules. (1)~A mandatory scratchpad forces explicit graph parsing and structure enumeration before any question is written, which empirically reduces hallucinated edges and misclassified entities. (2)~The scene-level preamble is restricted to navigation command and current ego action; spatial detail required for geometry-grounded questions is supplied in a per-question \texttt{spatial\_context} field, ensuring scene context enters the model's input only for questions whose answers depend on it. (3)~Questions and preambles are written in second person and answers and reasoning in first person, preventing voice leakage between question and rationale. (4)~A banned-word list excludes graph terminology, internal pipeline labels, and sensor-format names from all question, option, and reasoning text, so that an evaluation-time VLA cannot infer question type from surface phrasing. The full set of construction principles for ground-truth answers (descendant propagation, redundancy checking, behavioral impact, and rung verification) is given in \cref{app:answer_gen}.

\subsection{Stage 4: Counterfactual trajectory generation.}
\label{app:counterfactual_traj}

For every scene we generate ground-truth alternate ego trajectories under at most one Rung~2 binary intervention (drawn from CDE-Binary, NIE-Binary, NDE-Binary, or AB-Binary) and at most one Rung~3 binary counterfactual (drawn from CR, NC, or WC). This per-scene budget exists because trajectory synthesis is more expensive than QA generation by roughly an order of magnitude; the budget is allocated to the binary question that maximises three preferences in order: trajectory-changing answers over no-change answers, scene-specific over policy-level reasoning, and unique behavioural effects over redundant ones. Each selected question carries a \texttt{counterfactual\_intervention} record specifying the target entity, the operation (\texttt{remove}, \texttt{state\_change}, \texttt{freeze}, \texttt{reposition}, \texttt{speed\_change}, or \texttt{activate}), the temporal anchor ($t_0$ for present-time interventions, $t_0 - 1.5\,$s for counterfactual-past interventions), the frozen entities (non-descendants and \texttt{CONFOUNDED\_BY}-linked entities), and the propagated descendants (entities reachable via directed \texttt{AFFECTS} edges). All trajectory geometry is expressed in the \emph{anchor frame}: ego-local at the intervention timestep, with $+x$ forward and $+y$ left.

Trajectory synthesis proceeds in three sub-stages, each addressing a distinct failure mode of single-shot generation.

\paragraph{(4a) Trajectory-change judgement and auxiliary intervention inference.}
The QA generator records a \texttt{trajectory\_changes} flag from its own reasoning, but the flag is unreliable: roughly 7\% of flagged-true cases do not actually entail a behavioural change (a redundant constraint still gates ego), and roughly 3\% of flagged-false cases do (the question implies a manoeuvre the flag misses). We therefore re-evaluate every flag with Claude Haiku~4.5, conditioned on the question, the correct answer, the reasoning trace, the rung, the operation, and the anchor-frame agent list. The same call additionally identifies any \emph{auxiliary} agents whose behaviour is pinned by the question wording but not recorded in the original intervention spec, for example a question that says ``the SUV ahead kept rolling at its current pace'' pins that SUV to constant-velocity rollout regardless of how it would otherwise respond. Cases judged to retain factual ego behaviour are dropped from the trajectory pool, and auxiliary entities are appended to the intervention specification with an action drawn from a fixed vocabulary (\texttt{remove}, \texttt{freeze\_position}, \texttt{freeze\_velocity}, \texttt{state\_change}, \texttt{reposition}, \texttt{speed\_change}, \texttt{activate}).

\paragraph{(4b) Numeric state mutation.}
The intervention specification names entities and actions but does not yet contain the concrete physical states the alternate scene must reach. Three actions are deterministic and bypass the model: \texttt{remove} emits a null state, \texttt{freeze\_position} copies the observed pose with speed zeroed, and \texttt{freeze\_velocity} copies the observed pose and speed verbatim (including zero, when the agent was already stationary). The remaining four actions require visual reasoning to place or re-pose an agent on a drivable surface, and are routed to Claude Opus~4.6 conditioned on the four-camera stack at the anchor timestep, the four-frame BEV sequence covering $[t_0 - 1.5, t_0]$, the QA block, the proposed intervention specification, and the anchor-frame \texttt{state.json}. The model emits a \texttt{change\_state} record $(t, x, y, \mathrm{yaw}, \mathrm{speed})$ for every mutated entity, validated against a small JSON schema and merged back into the intervention specification of the source question.

\paragraph{(4c) Agentic trajectory synthesis.}
The third sub-stage synthesises the alternate ego trajectory itself. A tool-using LLM agent (Claude Opus~4.6) commits waypoints one at a time, conditioned on the visual context at the anchor and a tool surface that exposes the post-intervention scene: a heading-aware lane centerline drawn from the nuScenes map (including the \texttt{lane\_connector} layer that covers intersection bridges, without which mid-turn cases consistently select an incorrect straight lane); the post-intervention agent list with category-sized footprints; a path-target tool returning the lane-tracking pose at a given time and target speed; a curvature query that suggests a comfort-bounded peak speed for the upcoming arc; and a road-feature query that returns stop lines, pedestrian crossings, and traffic-light polygons along the upcoming centerline together with their arc distances. The agent emits eight waypoints over four seconds for Rung~2 cases (anchor at $t_0$) or three waypoints over 1.5 seconds for Rung~3 cases (anchor at $t_0 - 1.5\,$s), each spaced at $0.5\,$s and each preceded by a $t = 0$ anchor at the origin for connectivity. Every \texttt{commit\_waypoint} call triggers three deterministic in-loop checks before acceptance: \emph{lane-snap}, projecting $(x, y)$ onto the nearest lane centerline; \emph{collision}, computed as Separating-Axis-Theorem polygon overlap between the ego footprint ($4.5 \times 1.85\,$m at the committed pose) and every active agent's category-sized footprint (trucks at $7.0 \times 2.5\,$m, pedestrians at $0.7 \times 0.7\,$m, and so on); and \emph{kinematic feasibility}, requiring $|\Delta v| \leq 1.5\,$m/s and $|\Delta y| \leq 0.75\,$m between consecutive 0.5-s waypoints. Failing checks return the offending agent identifier and minimum-clearance value to the agent, which is instructed to revise the waypoint in place.

A deterministic post-process then re-integrates the committed waypoints into a final smoothed trajectory whose physical consistency is guaranteed by construction rather than relying on the agent's adherence to the kinematic checks. Speed is trapezoidally integrated into a monotonically non-decreasing forward coordinate, so the trajectory cannot reverse. The lateral coordinate is smoothed by an exponential moving average (with a tighter coefficient on the first step from the anchor, where models tend to overshoot from rest) and capped against a yaw-feasibility bound: a vehicle moving forward $\Delta x$ cannot translate laterally by more than ${\sim}0.7\,|\Delta x|$, corresponding to a maximum yaw rate that matches the comfort threshold of NAVSIM's Predictive Driver Model scores \citep{dauner2024navsim}. A soft repulsion term pushes the trajectory away from any active agent's oriented footprint within $1.5\,$m. The output is a list of $(t, x, y, \mathrm{heading}, \mathrm{speed})$ records expressed in the anchor frame, written back to the source QA file as \texttt{questions[i].meta.alternate\_trajectory}; this record serves as the candidate ground-truth counterfactual trajectory for the CF-ADE metric of \cref{sec:eval}, subject to the post-hoc collision-and-drivable-area audit of \cref{subsec:human_review}.

\section{Verification Details}
\label{app:verification_pipeline}

\subsection{Graph Verification}
\label{app:human_review}

\paragraph{Review interface.} The review tool is a custom web application that displays the multi-view camera stack, BEV rasterisation, structured agent-state record, and extracted causal scene graph in a single-pane layout. Annotators interact with the graph via per-node and per-edge editing controls: a node panel exposes type, category, label, position, visibility, and causal status; an edge panel exposes effect type (\texttt{GRANTS\_PROCEED}, \texttt{INHIBITS\_PROCEED}, \texttt{CONSTRAINS\_LONGITUDINAL}, \texttt{CONSTRAINS\_LATERAL}), endpoints, justification text, and the active/dormant flag. Annotators may accept a sample as-is, edit individual nodes or edges, demote an entity to the distractor pool, or remove it entirely. All edits are committed through an undo/redo stack and stored as a versioned graph artefact alongside the pre-edit graph; this preserves a full audit trail for later diagnostic analysis. Screenshots of the inspection and editing interfaces are in \cref{app:examples}.

\paragraph{Per-edit-type counts.} Across the 3{,}189 reviewed samples, the raw count of edit operations broke down as in \cref{tab:edit_counts}. Node-to-distractor demotion is the dominant operation by a wide margin: when annotators reclassify an entity as causally irrelevant, its outgoing ego-edges are dropped automatically, which accounts for the bulk of the 2{,}054 ego-edge removals. Additions of any kind (527 ego-edges, 107 inter-node, 89 nodes) are comparatively rare, indicating that Stage~1 over-generates causal candidates more often than it omits them.

\begin{table}[h]
\centering
\small
\begin{tabular}{lr}
\toprule
Edit operation & Count \\
\midrule
Node $\to$ distractor demotion        & 1{,}739 \\
Ego-edge removed                      & 2{,}054 \\
Inter-node edge removed               & 1{,}446 \\
Ego-edge added                        & 527 \\
Ego-edge causal-status string changed & 338 \\
Ego-edge active flag toggled          & 327 \\
Distractor $\to$ node promotion       & 314 \\
Ego-edge justification reworded       & 276 \\
Inter-node edge added                 & 107 \\
New node added                        & 89 \\
New distractor added                  & 9 \\
\bottomrule
\end{tabular}
\caption{Per-operation edit counts across the 3{,}189 reviewed samples. Operations are not mutually exclusive: a single review session may apply multiple edits to one graph.}
\label{tab:edit_counts}
\end{table}

\paragraph{Deterministic post-processing rules.} Patterns surfacing repeatedly during human review were formalised as deterministic rules and applied to all extracted graphs immediately after Stage~1, before any human review begins. Three representative rules formalise the most common review patterns:

\begin{enumerate}
\item \textbf{Rear-agent demotion.} \texttt{OTHER\_AGENT} nodes whose centroid lies behind ego (negative $x$ in ego-local coordinates) are demoted to the distractor pool, since trailing vehicles rarely cause ego's forward behaviour. Affected 1{,}483 scenes.
\item \textbf{Redundant crosswalk dropping.} When a pedestrian already exerts an active \texttt{INHIBITS\_PROCEED} edge on ego, any \texttt{TRAFFIC\_CONTROL} crosswalk-to-ego edge in the same lane segment is dropped, since the pedestrian-induced constraint subsumes the crosswalk-induced constraint and retaining both double-counts the same physical hazard.
\item \textbf{Distractor-type filtering.} The distractor pool is restricted to \texttt{OTHER\_AGENT} entries; \texttt{TRAFFIC\_CONTROL} and \texttt{ROAD\_OBSTACLE} entities never enter the distractor pool, since by construction distractors are rejected agent candidates rather than rejected infrastructure. Affected 256 scenes.
\end{enumerate}

In addition, two universal cleanup passes are applied to every graph: a disconnected-node pruning pass removes any node with no directed causal path to ego (such nodes contribute neither active, dormant, nor distractor questions), and a justification-alignment pass rewrites edge justification text whose factual claims contradict the post-edit graph structure using a deterministic template keyed on the edge's effect type and endpoints. Across the 3{,}189 reviewed samples, this rule set fired on 2{,}209 graphs (69\%); cases not amenable to deterministic resolution were left to human review.

\paragraph{Inter-annotator Agreement on Causal Labels}

To assess the reliability of the causal-graph schema, we measured inter-annotator agreement on a held-out 50 random samples. Two annotators independently labelled causal status, ego-edge effect type, and inter-node edges from the same multi-view camera and BEV inputs, without access to each other's labels or to the LLM-extracted graph. Agreement is reported per decision type rather than as a single graph-level score, since the underlying decisions are structurally distinct and combine differently into the final graph. Cohen's $\kappa$ values appear in \cref{tab:annotator_agreement}.

\begin{table}[h]
\centering
\caption{Inter-annotator agreement on causal labels over 50 random samples. Cohen's $\kappa$ reported per decision type. Landis-Koch interpretation: 0.41--0.60 moderate, 0.61--0.80 substantial, 0.81--1.00 almost perfect.}
\label{tab:annotator_agreement}
\begin{tabular}{lcc}
\toprule
Decision & Metric & Value \\
\midrule
Causal status (3-class, overall)        & $\kappa$ & 0.81 \\
\quad Active vs rest                    & $\kappa$ & 0.87 \\
\quad Dormant vs rest                   & $\kappa$ & 0.63 \\
\quad Distractor vs rest                & $\kappa$ & 0.91 \\
Ego edge effect type (4-class)          & $\kappa$ & 0.85 \\
Inter-node edge type (4-class)          & $\kappa$ & 0.75 \\
\bottomrule
\end{tabular}
\end{table}

Causal-status agreement is almost-perfect overall ($\kappa = 0.81$), with distractor (0.91) and active (0.87) classifications reaching the highest agreement and dormant (0.63) substantially lower. The dormant figure reflects genuine boundary cases in single-step physical reachability: while the agency-restriction and geometric-admissibility conditions of \cref{app:causal_status} constrain the dormant set, judgments about pedestrian intent and short-horizon agent trajectories remain a source of legitimate disagreement. We surface these boundary cases explicitly through the activation-plausibility classifier in the dormant-element prompt (\cref{app:qa_generation}), which records each dormant assignment as high, medium, or low plausibility, allowing downstream consumers of the benchmark to filter by confidence. Ego-edge effect-type agreement ($\kappa = 0.85$) reflects the binary structure of the edge-type decision tree (\cref{app:graph_schema}). Inter-node edge agreement is the lowest in the schema ($\kappa = 0.75$); AFFECTS edges between observable entities reach higher agreement, but CONFOUNDED\_BY edges, which by definition cite unobserved common causes, attract more disagreement on what constitutes a shared latent cause versus two independent responses to the same event.

\subsection{QA Verification}
\label{app:qa_verification}

\paragraph{Audit protocol.} Beyond the per-sample graph review of \cref{app:human_review}, we conduct an independent verification of the released QA pairs themselves to certify benchmark-quality label fidelity. A stratified random sample of $n = 365$ pairs (approximately 5\% of the 7{,}285 released) is drawn from the post-clustering benchmark, with strata defined by the type-by-rung cells of \cref{app:qa_taxonomy} to ensure proportional coverage of all entity-status, PCH-layer, and question-type combinations represented in the release. 
Each sampled QA is re-evaluated by an annotator who was not involved in the original review of the underlying scene graph; the annotator is presented with the multi-view camera stack, BEV, agent-state record, and the question text, but is blinded to the released ground-truth label and to all pipeline metadata (entity status, PCH layer, generating prompt, causal scene graph).
A QA is marked an error if and only if the independent annotator's answer differs from the released label.

\paragraph{Statistical procedure.} The audit identified $k=3$ errors in $n=365$ pairs, yielding $\hat{p} = k/n = 0.82\%$. We report a 95\% Wilson score interval~\citep{wilson1927probable} rather than the normal-approximation Wald interval, since the Wald interval is unreliable for small $\hat{p}$ near zero; the Wilson interval for these counts is $[0.28\%,\,2.39\%]$. To formally reject the hypothesis that the true error rate exceeds 5\%, we apply a one-sided $z$-test against $H_0: p \geq 0.05$:
\begin{equation}
z \;=\; \frac{\hat{p} - p_0}{\sqrt{p_0(1 - p_0)/n}} \;=\; \frac{0.0082 - 0.05}{\sqrt{0.05 \cdot 0.95 / 365}} \;=\; -3.66,
\label{eq:audit_ztest}
\end{equation}
yielding $p\text{-value} = 0.000125$. We therefore reject the hypothesis that the benchmark's true error rate exceeds 5\% with high confidence, certifying the released benchmark to standard label-fidelity expectations for evaluation suites of this scale.

\subsection{Trajectory Audit}
\label{app:trajectory_audit}

The trajectory audit is the final validation gate of the construction pipeline, applied to every reference counterfactual trajectory produced by Stage~4 (\cref{subsec:pipeline}). Although Stage~4 commits ego waypoints against per-step collision, lane-snap, and kinematic-feasibility checks, the audit re-validates the smoothed integrated trajectory at higher temporal resolution and against the full post-intervention scene to catch failure modes that survive per-step gating, such as interpolation artefacts between committed waypoints or near-miss collisions that fall just outside the per-step safety margin.

\paragraph{Collision check.} Each trajectory is rasterised at 2\,Hz 
and the swept ego footprint at each timestep is intersected against every other agent's category-sized footprint at the corresponding time in the post-intervention scene. Footprint dimensions follow nuScenes category conventions (e.g., 4.7\,m $\times$ 1.9\,m for cars, 0.5\,m radius for pedestrians).
A trajectory is rejected if any non-zero intersection is detected within a fixed safety buffer of the agent footprint.

\paragraph{Drivable-area check.} The post-intervention lane network is recomputed from the nuScenes HD map after applying the counterfactual scene modifications: removed \texttt{TRAFFIC\_CONTROL} entities may unlock previously inaccessible lanes, repositioned agents may render previously open lanes blocked, and so on. A trajectory is rejected if any rasterised waypoint falls outside the unionised drivable polygon defined by lanes consistent with the scene's navigation command.

\paragraph{Audit results.} Of the candidate pool of 1,177 produced by Stage~4, 177 trajectories failed at least one check.
The remaining 1{,}000 reference trajectories form the released set used for CTE-R2 and CTE-R3 evaluation (\cref{subsec:cf_trajectory}). We treat these as physically and causally consistent reference trajectories rather than sensor-grounded ground truth: the human-driver counterfactual is by construction unobservable in recorded data, since we cannot replay the scene with the counterfactual premise enacted, and any counterfactual evaluation in real-world driving inherits this constraint.

\section{Additional Experimental Results}  %
\label{app:evaluation}

\subsection{Per-question-type Accuracy}
\label{app:per_qa_type_accuracy}

Table~\ref{tab:per_type} reports accuracy across the 18 canonical \cdb question types. The cross-type variance pattern is informative for benchmark design: the high-variance MCQ types (BAS $\sigma=38.5$, CoI $\sigma=35.6$, DR $\sigma=32.7$, CaI $\sigma=31.0$) discriminate models cleanly because they reward both reasoning and reliable instruction-following, while the low-variance types (CDE $\sigma=10.4$, WC $\sigma=12.6$, CR $\sigma=12.6$) cluster in a 45--65\% band that no model breaks through. The active-element counterfactual battery (CR, NDE, NIE, NC) and the distractor wild counterfactual (WC) form a uniform difficulty ceiling: difficulty here is genuine, not concentrated in a few weak models.

\begin{table}[h!]
\centering
\caption{Per-question-type accuracy (\%) across the 18 canonical \cdb types. Bold = best per column. Driving VLAs upper block, general VLMs lower block.}
\label{tab:per_type}
\resizebox{\textwidth}{!}{%
\begin{tabular}{lcccccccccccccccccc}
\toprule
Model & CaI & CA & CB & CoI & BAS & CDE & CR & SC & NC & NDE & NIE & DQ & EL & AB & DR & CI & NI & WC \\
\midrule
\multicolumn{19}{l}{\textit{Driving-specific VLAs}} \\
ImpromptuVLA-7B & 87.5 & 71.4 & 26.9 & 91.5 & 90.9 & 69.4 & \textbf{67.6} & 59.0 & 70.9 & 78.3 & 52.0 & 51.4 & 72.4 & 69.8 & 21.7 & 58.6 & \textbf{93.0} & 70.1 \\
OpenREAD & \textbf{87.9} & 87.8 & 45.6 & 93.2 & 90.9 & 67.7 & 61.7 & 64.7 & 71.4 & 52.2 & 58.0 & 47.9 & 72.9 & 63.2 & 76.3 & 41.0 & 86.6 & 56.7 \\
Orion & 0.0 & 0.0 & 12.1 & 0.0 & 0.0 & 43.6 & 65.1 & 14.7 & 55.9 & 54.4 & 36.0 & 59.9 & 46.3 & 52.4 & 0.0 & \textbf{94.1} & 79.9 & 63.0 \\
UniDrive-VLA & 82.4 & 89.8 & 58.4 & \textbf{98.3} & 95.5 & 61.3 & 49.4 & 66.5 & 65.7 & 58.7 & 60.0 & 37.1 & 60.6 & 65.8 & 83.5 & 15.4 & 68.0 & 63.2 \\
RecogDrive & 87.6 & \textbf{100.0} & 44.0 & 88.7 & 90.5 & 62.7 & 55.2 & 78.1 & 58.6 & 65.9 & \textbf{66.7} & 24.1 & 61.7 & 63.2 & 48.4 & 29.1 & 54.7 & 57.8 \\
SafeAuto & 66.8 & 63.3 & 24.8 & 81.4 & 77.3 & 56.5 & 61.9 & 38.6 & 59.2 & \textbf{82.6} & 62.0 & 27.0 & 56.1 & 44.7 & 87.6 & 40.7 & 59.1 & 39.7 \\
WiseAD & 26.7 & 36.7 & \textbf{70.5} & 37.3 & 45.5 & 69.4 & 57.2 & 60.2 & 55.4 & 47.8 & 66.0 & 19.9 & 23.2 & 54.1 & 11.3 & 16.2 & 38.4 & 56.4 \\
Alpamayo-1.5-10B & 47.2 & 42.9 & 66.4 & 30.5 & 9.1 & 46.8 & 35.3 & 61.6 & 41.3 & 39.1 & 42.0 & 25.1 & 20.5 & 46.5 & 27.8 & 16.7 & 37.9 & 48.4 \\
SimLingo & 51.7 & 65.3 & 72.5 & 64.4 & 31.8 & 45.2 & 38.4 & 37.5 & 31.9 & 41.3 & 62.0 & 21.0 & 43.4 & 40.3 & 54.6 & 14.3 & 24.0 & 41.3 \\
Omnidrive & 6.3 & 16.3 & 50.7 & 5.5 & 0.0 & 49.2 & 28.5 & 59.4 & 33.2 & 31.1 & 18.8 & 7.0 & 26.1 & 30.5 & 0.0 & 23.0 & 8.9 & 33.4 \\
\midrule
\multicolumn{19}{l}{\textit{General-purpose VLMs}} \\
Cosmos-Reason-2 & 86.8 & 93.9 & 28.9 & \textbf{98.3} & \textbf{100.0} & 64.5 & 58.5 & 69.8 & 68.1 & 63.0 & 52.0 & \textbf{65.6} & \textbf{79.1} & 66.9 & \textbf{92.8} & 69.8 & 77.6 & 66.7 \\
Qwen3-VL-8B & 58.5 & 67.4 & 33.6 & 86.4 & 77.3 & 71.0 & 64.8 & 37.1 & 68.5 & 67.4 & 58.0 & 45.0 & 76.4 & \textbf{73.8} & 57.7 & 64.7 & 88.0 & 70.3 \\
InternVL-3.5-8B & 73.8 & 85.7 & 49.7 & 86.4 & 86.4 & \textbf{71.0} & 62.5 & 54.7 & 68.1 & 67.4 & 56.0 & 29.9 & 71.7 & 72.3 & 54.6 & 44.9 & 81.3 & \textbf{71.7} \\
\midrule
Mean & 58.7 & 63.1 & 44.9 & 66.3 & 61.2 & 59.8 & 54.3 & 54.0 & 57.5 & 57.6 & 53.0 & 35.5 & 54.6 & 57.2 & 47.4 & 40.7 & 61.3 & 56.8 \\
$\sigma$ & 31.0 & 31.0 & 19.0 & 35.6 & 38.5 & 10.4 & 12.6 & 17.3 & 13.9 & 15.2 & 13.6 & 17.3 & 20.9 & 13.6 & 32.7 & 25.1 & 26.9 & 12.6 \\
\bottomrule
\end{tabular}}
\end{table}

\subsection{Active-element Accuracy by Graph Topology}
\label{app:active_eval}

Table~\ref{tab:active_topology} breaks down active-element accuracy by the four canonical causal subgraphs of \ref{subsec:pipeline}. Direct structures dominate volume (68.1\% of active QA) and are the easiest on average (cross-model mean 56.5\%); Confounding (11.8\%) and Chain (5.3\%) require tracing inter-node edges and are correspondingly harder; Collider (14.8\%) is the most discriminating because it specifically probes the explaining-away pattern that no current model handles well. The pattern is consistent with the conditionality bottleneck identified in the reasoning-judge breakdown.

\begin{table}[h!]
\centering
\caption{Active-element accuracy (\%) by graph topology. Dormant and distractor questions have empty graph structure by construction. Bold = best per column.}
\label{tab:active_topology}
\begin{tabular}{lccccc}
\toprule
Model & Active overall & Direct & Chain & Confounding & Collider \\
\midrule
\multicolumn{6}{l}{\textit{Driving-specialist VLAs}} \\
RecogDrive & 69.9 & \textbf{73.2} & \textbf{74.0} & 70.9 & 52.5 \\
OpenREAD & 69.8 & 71.0 & 68.5 & 74.8 & 60.7 \\
ImpromptuVLA-7B & 69.1 & 71.1 & 72.4 & 76.2 & 52.7 \\
UniDrive-VLA & 65.9 & 65.0 & 68.5 & 74.5 & \textbf{62.1} \\
SafeAuto & 56.6 & 55.9 & 68.5 & 69.9 & 45.4 \\
WiseAD & 50.6 & 47.7 & 49.6 & 53.2 & \textbf{62.4} \\
Alpamayo-1.5-10B & 46.4 & 47.6 & 36.2 & 36.0 & 52.7 \\
SimLingo & 44.7 & 42.0 & 54.3 & 51.1 & 48.5 \\
Omnidrive & 30.9 & 30.6 & 24.4 & 23.4 & 40.1 \\
Orion & 30.3 & 27.9 & 27.6 & 36.0 & 37.6 \\
\midrule
\multicolumn{6}{l}{\textit{General-purpose VLMs}} \\
Cosmos-Reason-2 & 68.7 & 71.2 & 71.7 & \textbf{75.9} & 50.7 \\
InternVL-3.5-8B & 64.7 & 64.1 & 73.2 & 70.6 & 59.9 \\
Qwen3-VL-8B & 56.7 & 54.1 & 66.9 & 71.7 & 53.5 \\
\bottomrule
\end{tabular}
\end{table}

\subsection{Reasoning Quality}
\label{app:reasoning_quality}
The LLM judge scores all 13 models across four reasoning dimensions on a 0--2 scale, with each score the mean of two independent judges (Claude Sonnet 4.6 and ChatGPT 4o). Composite scores and per-dimension breakdowns appear in Table~\ref{tab:reasoning_scores}; inter-judge agreement is summarised in Table~\ref{tab:judge_agreement}. Two patterns are universal. First, \textbf{conditionality awareness is the weakest dimension across the board}: it is the lowest-scoring dimension for every one of the 13 models, with no exceptions. The cross-model mean of 0.54 (against 0.98 for spatial grounding, 0.95 for mechanism identification, and 1.16 for answer consistency) marks conditionality as the principal reasoning bottleneck across all evaluated architectures. Second, \textbf{answer consistency is the strongest dimension} for 11 of the 13 models (the two exceptions, Alpamayo-1.5-10B and Omnidrive, score higher on spatial grounding by 0.06 and 0.25 respectively), indicating that most models reliably maintain coherence between their reasoning text and their final answer; the failure is in the reasoning's content, not its alignment to the chosen answer. The conditionality deficit corresponds directly to the R1 collider-bias result analysed in \cref{app:cb_gap}: correctly suppressing spuriously opened collider paths requires precisely the conditional causal reasoning that this dimension measures.

A notable third pattern is the rank inversion at the top of the leaderboard. Cosmos-Reason-2 holds the highest overall VQA accuracy (70.56\%) yet falls behind both Qwen3-VL-8B (composite 1.35) and RecogDrive (1.25) on reasoning quality, tied with InternVL-3.5-8B at 1.19. Orion sits at the opposite corner: it answers 57.25\% of questions correctly (above the cross-model mean of 52.51\%) while posting the leaderboard's lowest reasoning composite (0.19) and conditionality (0.06), the starkest case of correct answers paired with content-thin reasoning. RecogDrive completes the picture from a third direction: mid-pack on QA at 54.25\% but second-best on reasoning composite at 1.25, consistent with its planner-only RFT design that leaves the underlying InternVL-3.5-8B backbone untouched. Together these cases sharpen the case that QA accuracy on its own does not certify causal reasoning depth.

\paragraph{Inter-judge agreement.} To check that these conclusions are not artefacts of any single judge's stylistic biases, we score every model with both Claude Sonnet 4.6 and ChatGPT 4o and report Pearson and Spearman correlations between the two judges' per-model dimension scores in Table~\ref{tab:judge_agreement}. Agreement is uniformly strong: composite Pearson $r = 0.97$ and Spearman $\rho = 0.92$, with mechanism identification reaching $r = 0.99$. Even the weakest-correlated dimension, answer consistency, registers $r = 0.83$ and $\rho = 0.81$, well above conventional thresholds for substantial agreement. Both the conditionality bottleneck and the QA--reasoning rank inversion reproduce under either judge in isolation, so the patterns reported in Table~\ref{tab:reasoning_scores} are robust to judge choice.

\begin{table}[h!]
\centering
\caption{LLM-judge reasoning scores (0--2 per dimension; higher is better), averaged over Claude Sonnet 4.6 and ChatGPT 4o. Composite is the mean of the four dimensions. Bold = best per column. Driving VLAs upper block, general VLMs lower block.}
\label{tab:reasoning_scores}
\begin{tabular}{lrrrrr}
\toprule
Model & Composite & Spatial & Mechanism & Conditionality & Consistency \\
\midrule
\multicolumn{6}{l}{\textit{Driving-specialist VLAs}} \\
RecogDrive         & 1.25 & 1.36 & 1.40 & 0.75 & 1.48 \\
OpenREAD           & 1.18 & 1.14 & 1.30 & 0.83 & 1.44 \\
UniDrive-VLA       & 1.12 & 1.21 & 1.24 & 0.74 & 1.31 \\
ImpromptuVLA-7B    & 1.06 & 1.08 & 1.12 & 0.63 & 1.39 \\
SimLingo           & 0.71 & 0.91 & 0.70 & 0.30 & 0.93 \\
WiseAD             & 0.71 & 0.81 & 0.67 & 0.27 & 1.07 \\
Alpamayo-1.5-10B   & 0.71 & 0.87 & 0.78 & 0.38 & 0.81 \\
Omnidrive          & 0.65 & 0.87 & 0.70 & 0.40 & 0.62 \\
SafeAuto           & 0.50 & 0.57 & 0.35 & 0.16 & 0.93 \\
Orion              & 0.19 & 0.14 & 0.07 & 0.06 & 0.51 \\
\midrule
\multicolumn{6}{l}{\textit{General-purpose VLMs}} \\
Qwen3-VL-8B        & \textbf{1.35} & \textbf{1.41} & \textbf{1.45} & \textbf{0.99} & 1.55 \\
InternVL-3.5-8B    & 1.19 & 1.25 & 1.30 & 0.86 & 1.34 \\
Cosmos-Reason-2    & 1.19 & 1.15 & 1.25 & 0.70 & \textbf{1.65} \\
\midrule
Cross-model mean   & 0.91 & 0.98 & 0.95 & 0.54 & 1.16 \\
\bottomrule
\end{tabular}
\end{table}

\begin{table}[h!]
\centering
\caption{Inter-judge agreement on per-model reasoning scores between Claude Sonnet 4.6 and ChatGPT 4o across the 13 evaluated models, computed per dimension over the 13 model-level scores.}
\label{tab:judge_agreement}
\begin{tabular}{lrr}
\toprule
Dimension & Pearson $r$ & Spearman $\rho$ \\
\midrule
Composite                  & 0.97 & 0.92 \\
Spatial grounding          & 0.96 & 0.93 \\
Mechanism identification   & 0.99 & 0.97 \\
Conditionality awareness   & 0.94 & 0.94 \\
Answer consistency         & 0.83 & 0.81 \\
\bottomrule
\end{tabular}
\end{table}

\subsection{Rung 1 Analysis: Collider Bias and the Conditionality Gap}
\label{app:cb_gap}

R1 in \cdb contains exactly one question type per source: Collider Bias (CB) for active questions, Explanation of Latency (EL) for dormant questions, and Causal Irrelevance (CI) for distractor questions. CB is the type that most directly stresses Pearl's collider semantics: conditioning on a collider opens a spurious path between its parents, creating apparent statistical dependence where no causal link exists. In the driving setting, a road junction is a collider between the routes of two converging vehicles; observing both present simultaneously creates the illusion of speed correlation that a model must learn to suppress. Per-model R1 accuracy across the three sources appears in Table~\ref{tab:rung1_breakdown}.

CB accuracy reveals a striking \textbf{inverse relationship} with overall accuracy that strengthens at the top of the leaderboard, with Pearson correlation of $-0.62$ across the 13 models. The three highest-overall models all show large negative deltas from overall to CB: ImpromptuVLA-7B (overall 69.27\%, CB 26.85\%, $\Delta = -42.4$\,pp), Cosmos-Reason-2 (70.56\%, CB 28.86\%, $\Delta = -41.7$\,pp), and Qwen3-VL-8B (66.04\%, CB 33.56\%, $\Delta = -32.5$\,pp). OpenREAD ($-18.1$\,pp) and InternVL-3.5-8B ($-13.3$\,pp) sit in a milder version of the same pattern. The relationship inverts at the bottom: SimLingo (overall 33.95\%, CB 72.48\%, $\Delta = +38.5$\,pp), WiseAD (39.35\%, CB 70.47\%, $\Delta = +31.1$\,pp), and Alpamayo-1.5-10B (36.86\%, CB 66.44\%, $\Delta = +29.6$\,pp) score \emph{above} their overall on CB.

This pattern reflects \emph{answer-format bias}: CB questions in the benchmark frequently have ``No'' as the correct answer (the spurious path is not active), and models with a strong modal-No bias score well on CB while failing broadly across the rest of the benchmark. The LLM judge corroborates this interpretation: conditionality awareness is the weakest reasoning dimension for every judged model (Table~\ref{tab:reasoning_scores}), directly mirroring the CB failure mode in the strongest-overall models. Genuine collider-bias mastery requires recognising that a causal path is conditionally active only when its collider is observed, not merely outputting the modal binary response.

The dual cases on dormant and distractor questions sharpen the picture further. EL, the dormant R1 type, is well handled by general VLMs and the strongest VLAs (60--79\%) but poorly handled by every other driving VLA (16--26\%), suggesting that explaining \emph{why} a visible element is not currently constraining ego is a capability concentrated in models with strong base reasoning. CI, the distractor R1 type, is the lowest-scoring R1 source overall, with 9 of 13 models scoring below 50\%; this mirrors the causal-hallucination failure mode visible in the QA-source breakdown (\cref{app:active_eval}).

\begin{table}[h!]
\centering
\caption{Rung 1 accuracy decomposed by question source (\%). CB targets active-graph collider structures; EL targets dormant entities and probes structural decoupling, shielding, or shared cause; CI targets perceptually salient distractors. Driving VLAs upper block, general VLMs lower block; each block sorted by overall accuracy.}
\label{tab:rung1_breakdown}
\begin{tabular}{lrrrr}
\toprule
Model & CB (Active) & EL (Dormant) & CI (Distractor) & Overall \\
\midrule
\multicolumn{5}{l}{\textit{Driving-specialist VLAs}} \\
ImpromptuVLA-7B    & 26.85\% & 72.38\% & 58.64\% & 69.27\% \\
OpenREAD           & 45.64\% & 72.93\% & 40.95\% & 63.76\% \\
Orion              & 12.08\% & 46.33\% & 94.08\% & 57.25\% \\
UniDrive-VLA       & 58.39\% & 60.58\% & 15.39\% & 54.87\% \\
RecogDrive         & 43.97\% & 61.69\% & 29.14\% & 54.25\% \\
SafeAuto           & 24.83\% & 56.12\% & 40.73\% & 49.42\% \\
WiseAD             & 70.47\% & 23.16\% & 16.18\% & 39.35\% \\
Alpamayo-1.5-10B   & 66.44\% & 20.49\% & 16.73\% & 36.86\% \\
SimLingo           & 72.48\% & 43.43\% & 14.29\% & 33.95\% \\
Omnidrive          & 50.69\% & 26.06\% & 23.00\% & 24.10\% \\
\midrule
\multicolumn{5}{l}{\textit{General-purpose VLMs}} \\
Cosmos-Reason-2    & 28.86\% & 79.06\% & 69.77\% & 70.56\% \\
Qwen3-VL-8B        & 33.56\% & 76.39\% & 64.72\% & 66.04\% \\
InternVL-3.5-8B    & 49.66\% & 71.71\% & 44.91\% & 62.91\% \\
\bottomrule
\end{tabular}
\end{table}

\subsection{Trajectory Evaluation}
\label{app:trajectory_eval}

\paragraph{BTE.}
UniDrive-VLA achieves the best baseline trajectory performance (mean ADE = 0.68\,m), followed by RecogDrive (1.19\,m), ImpromptuVLA-7B (1.27\,m), and Alpamayo-1.5-10B (1.31\,m). The 12$\times$ spread (0.68 to 8.10\,m) reflects architectural differences: models with dedicated waypoint-regression heads vastly outperform those relying on autoregressive JSON generation of coordinate tokens. BTE accuracy does not track VQA accuracy; the Pearson correlation between BTE ADE and overall VQA accuracy across the 13 models is approximately zero. Cosmos-Reason-2 (top VQA at 70.56\%) sits 5th on BTE (1.79\,m), and InternVL-3.5-8B (5th VQA at 62.91\%) sits 12th on BTE (7.71\,m), while UniDrive-VLA (best BTE) reaches only 54.87\% on VQA. Causal language understanding and trajectory regression are statistically independent capabilities in current VLA architectures.

\paragraph{CTE.}
CTE performance is uniformly poor across nearly all models. The dominant failure mode is \texttt{TRAJECTORY\_SAME\_AS\_BTE}: most models output near-identical waypoints regardless of the counterfactual condition, with Orion at 98.7\%, RecogDrive at 89.0\%, and SimLingo at 80.4\%. WiseAD is the lone exception, producing a meaningful 48.7\% \texttt{GOOD\_COUNTERFACTUAL} rate (2.4$\times$ the next-best); its CTE-R2 and CTE-R3 ADE values (1.97\,m and 1.57\,m) sit \emph{below} its BTE ADE (2.91\,m), indicating that its counterfactual trajectories track the modified ground truth more closely than its baseline tracks the original. No other model approaches this profile.

The gap between a model's ability to \emph{describe} what should happen under a counterfactual (measured by VQA) and its ability to \emph{act accordingly} (measured by CTE) is the central finding of \cdb. The top VQA performers translate causal QA accuracy into counterfactual trajectory adaptation only weakly: ImpromptuVLA-7B reaches 15.04\% \texttt{GOOD\_CF}, Qwen3-VL-8B 14.93\%, and Cosmos-Reason-2's rerun pipeline emits a different grade taxonomy (\texttt{GOOD}~=~6.60\%, \texttt{EXCELLENT}~=~8.10\%, together 14.70\%). Alpamayo-1.5-10B reaches 16.80\% \texttt{GOOD\_CF} despite an \texttt{ANSWER\_MISMATCH} flag firing on 57.40\% of CTE samples, indicating that its trajectories shift in the right direction even when its language answer disagrees. Current VLA architectures do not reliably close the loop between causal language reasoning and motor planning.

\begin{table}[h!]
\centering
\caption{CTE grade and flag distribution. \texttt{GOOD\_CF} fires when the counterfactual trajectory differs meaningfully from BTE in the prescribed direction. \texttt{TRAJ\_SAME\_BTE} fires when the counterfactual is essentially unchanged from BTE. Percentages are relative to total CTE count. Flags can co-occur, so columns need not sum to 100\%. Driving VLAs upper block, general VLMs lower block; each block sorted by \texttt{GOOD\_CF}. Bold = best per column (\texttt{GOOD\_CF} higher is better, others lower). SafeAuto's grade and flag fields were not retained in its trajectory report. Alpamayo-1.5-10B's high \texttt{FLAGGED} share is dominated by an \texttt{ANSWER\_MISMATCH} flag (57.40\%). Cosmos-Reason-2's rerun pipeline emits a \texttt{GOOD}/\texttt{EXCELLENT} grade taxonomy rather than \texttt{INCONCLUSIVE}/\texttt{FLAGGED}, so the four columns shown read 0\%; the full grade breakdown for Cosmos is \texttt{GOOD}=6.60\%, \texttt{EXCELLENT}=8.10\%.}
\label{tab:cte_flags}
\resizebox{\textwidth}{!}{%
\begin{tabular}{lrrrrr}
\toprule
Model & Total CTE & GOOD\_CF (\%) & INCONCLUSIVE (\%) & FLAGGED (\%) & TRAJ\_SAME\_BTE (\%) \\
\midrule
\multicolumn{6}{l}{\textit{Driving-specialist VLAs}} \\
WiseAD             & 972  & \textbf{48.66} & 33.85 & \textbf{17.49} & \textbf{17.49} \\
UniDrive-VLA       & 1000 & 20.10 & 37.90 & 42.00 & 42.00 \\
Alpamayo-1.5-10B   & 1000 & 16.80 & 2.50  & 63.50 & 34.00 \\
ImpromptuVLA-7B    & 984  & 15.04 & 45.53 & 42.28 & 22.97 \\
Omnidrive          & 999  & 10.71 & 42.64 & 46.65 & 46.65 \\
OpenREAD           & 861  & 10.22 & 46.92 & 42.86 & 42.86 \\
RecogDrive         & 1000 & 3.90  & 7.10  & 89.00 & 89.00 \\
SimLingo           & 1000 & 3.40  & 16.20 & 80.40 & 80.40 \\
Orion              & 1000 & 0.00  & 1.30  & 98.70 & 98.70 \\
SafeAuto           & 1000 & N/A   & N/A   & N/A   & N/A \\
\midrule
\multicolumn{6}{l}{\textit{General-purpose VLMs}} \\
InternVL-3.5-8B    & 988  & 15.99 & \textbf{61.84} & 22.17 & 22.17 \\
Qwen3-VL-8B        & 931  & 14.93 & 60.58 & 24.49 & 24.49 \\
Cosmos-Reason-2    & 1000 & 0.00  & 0.00  & 0.00  & 0.00  \\
\bottomrule
\end{tabular}}
\end{table}

\subsection{CTE Failure-mode Regimes}
\label{app:cte_regimes}

The dominant \texttt{TRAJECTORY\_SAME\_AS\_BTE} statistic compresses three structurally distinct counterfactual-trajectory failure modes into a single bucket. The CTE-R2/BTE ratio together with the \texttt{TRAJ\_SAME\_BTE} flag separates them; per-model assignments appear in Table~\ref{tab:cte_regimes}.

\textbf{Regime A: Frozen planners} (CTE-R2/BTE within 10\% of unity, high \texttt{TRAJ\_SAME\_BTE}). Orion (ratio 0.97, 98.7\% same), SimLingo (1.09, 80.4\%), OpenREAD (1.09, 42.9\%). The trajectory is essentially invariant to the counterfactual prompt; this is the canonical ornamental-reasoning failure.

\textbf{Regime B: Wild-swerving planners} (CTE-R2/BTE $\gg 1$). ImpromptuVLA-7B (4.48), UniDrive-VLA (4.26), RecogDrive (3.61), Alpamayo-1.5-10B (2.06), Cosmos-Reason-2 (1.93), Qwen3-VL-8B (1.57). When prompted with a counterfactual, these models swing 1.5 to 4.5$\times$ further from BTE than BTE itself deviates from the observed ground truth: the model attempts to respond to the intervention but lacks the structured representation to do so proportionately.

\textbf{Regime C: Default-to-stop planners} (CTE-R3/BTE $<1$). 6 of 13 models, including SimLingo (0.46), Omnidrive (0.46), WiseAD (0.52), InternVL-3.5-8B (0.54), Orion (0.58), SafeAuto (0.62). The counterfactual trajectory is shorter than the baseline, mechanically suspicious: the most likely interpretation is that under counterfactual prompts these models default to short generic trajectories (deceleration to stop, lane-keeping at constant speed) which happen to be closer to many CTE-R3 ground-truth trajectories than the model's confident BTE prediction was.

WiseAD is the single positive outlier across all three regimes. Its CTE-R2/BTE ratio of 0.68 sits in the deceleration band but is the only model where this reflects proportionate counterfactual differentiation rather than collapse to a default trajectory.

\begin{table}[h!]
\centering
\caption{CTE failure-mode regime classification. A = frozen, B = wild-swerving, C = default-to-stop. Some models exhibit characteristics of more than one regime (B+C: wild on R2 and stop-collapsing on R3; A+C: frozen on R2 and stop-collapsing on R3). WiseAD is the single positive outlier with proportionate counterfactual differentiation in the prescribed direction.}
\label{tab:cte_regimes}
\resizebox{\textwidth}{!}{%
\begin{tabular}{lcccccc}
\toprule
Model & BTE (m) & CTE-R2 (m) & CTE-R3 (m) & CTE-R2/BTE & CTE-R3/BTE & Regime \\
\midrule
\multicolumn{7}{l}{\textit{Driving-specialist VLAs}} \\
WiseAD             & 2.91 & 1.97 & 1.57 & 0.68 & 0.54 & C (positive) \\
UniDrive-VLA       & 0.68 & 2.90 & 1.96 & 4.26 & 2.88 & B \\
Alpamayo-1.5-10B   & 1.31 & 2.70 & 1.70 & 2.06 & 1.30 & B+C \\
ImpromptuVLA-7B    & 1.27 & 5.69 & 2.97 & 4.48 & 2.34 & B \\
Omnidrive          & 5.39 & 4.51 & 2.48 & 0.84 & 0.46 & C \\
OpenREAD           & 3.97 & 4.34 & 4.61 & 1.09 & 1.16 & A \\
RecogDrive         & 1.19 & 4.30 & 1.97 & 3.61 & 1.66 & A/B \\
SimLingo           & 4.65 & 5.07 & 2.12 & 1.09 & 0.46 & A+C \\
Orion              & 8.10 & 7.88 & 4.73 & 0.97 & 0.58 & A \\
SafeAuto           & 5.43 & 4.53 & 3.35 & 0.83 & 0.62 & C \\
\midrule
\multicolumn{7}{l}{\textit{General-purpose VLMs}} \\
InternVL-3.5-8B    & 7.71 & 4.81 & 4.18 & 0.62 & 0.54 & C \\
Qwen3-VL-8B        & 3.68 & 5.76 & 3.58 & 1.57 & 0.97 & B \\
Cosmos-Reason-2    & 1.79 & 3.46 & 4.91 & 1.93 & 2.74 & B \\
\bottomrule
\end{tabular}}
\end{table}

\subsection{Backbone Analysis}
\label{app:backbone_analysis}

Of the 10 driving VLAs evaluated, 5 share their base language model with one of the 3 VLMs in our comparison group. This lets us measure the cost of driving-specific post-training directly, by comparing each driving VLA against the vanilla version of its own backbone rather than against an unrelated architecture. Backbone assignments are summarised in Table~\ref{tab:backbone_map}, and the resulting same-backbone gaps are reported in Table~\ref{tab:backbone_gaps}.

\begin{table}[h!]
\centering
\caption{Language backbone of each evaluated model. Driving VLAs in the upper block, general VLMs in the lower block. Models marked $^{*}$ appear as the backbone of one or more driving VLAs in the upper block.}
\label{tab:backbone_map}
\begin{tabular}{lll}
\toprule
Model & Language backbone & Source \\
\midrule
\multicolumn{3}{l}{\textit{Driving-specific VLAs}} \\
ImpromptuVLA-7B    & Qwen2.5-VL-7B          & \citep{chi2025impromptu} §2 \\
OpenREAD           & Qwen3-VL-8B$^{*}$      & \citep{zhang2025openread} §3.1 \\
UniDrive-VLA       & Qwen3-VL-8B$^{*}$      & \citep{li2026unidrivevla} Tab.~3 \\
RecogDrive         & InternVL3-8B$^{*}$     & \citep{li2025recogdrive} App.~E \\
Alpamayo-1.5-10B   & Cosmos-Reason$^{*}$    & \citep{wang2025alpamayo} §3.1 \\
WiseAD             & MobileVLM (CLIP + MobileLLaMA) & \citep{zhang2024wisead} §3.1 \\
Orion              & Vicuna v1.5            & \citep{fu2025orion} §4.2 \\
Omnidrive          & LLaVA-7B               & \citep{wang2025omnidrive} \\
SafeAuto           & -- (not surfaced)     & \citep{zhang2025safeauto} \\
SimLingo           & -- (not surfaced)     & \citep{renz2025simlingo} \\
\midrule
\multicolumn{3}{l}{\textit{General-purpose VLMs}} \\
Cosmos-Reason-2$^{*}$  & Qwen2.5-VL + Physical AI SFT/RL & \citep{azzolini2025cosmos} \\
Qwen3-VL-8B$^{*}$      & self                            & \citep{bai2025qwen3} \\
InternVL-3.5-8B$^{*}$  & self                            & \citep{chen2024internvl} \\
\bottomrule
\end{tabular}
\end{table}

\paragraph{Same-backbone gaps.}
For each model pair where backbone is shared, Table~\ref{tab:backbone_gaps} reports the change in overall causal QA accuracy from the vanilla VLM to the driving-specialised version. The cost of driving fine-tuning ranges from negligible ($-2.3$\,pp for OpenREAD on Qwen3-VL-8B) to severe ($-33.7$\,pp for Alpamayo-1.5-10B on Cosmos-Reason). The 31-percentage-point spread within this controlled comparison shows that driving fine-tuning is not a single intervention with a uniform effect on causal QA. Two driving VLAs trained on the same Qwen3-VL-8B backbone (OpenREAD and UniDrive-VLA) lose 2.3 and 11.2 percentage points respectively - a 4.9$\times$ difference attributable entirely to differences in their post-training procedures. The next subsection (\cref{app:rft_analysis}) examines these procedures and identifies a specific design choice that correlates with the size of the gap.

\begin{table}[h!]
\centering
\caption{Overall causal QA accuracy when the same language backbone is post-trained for general-purpose use (left) versus driving (right). $\Delta$ measures the cost of driving-specific post-training in percentage points.}
\label{tab:backbone_gaps}
\begin{tabular}{lrlrr}
\toprule
Backbone & Vanilla VLM (\%) & Driving VLA & Driving VLA (\%) & $\Delta$ (pp) \\
\midrule
Qwen3-VL-8B    & 66.04 & OpenREAD          & 63.76 & $-2.3$ \\
Qwen3-VL-8B    & 66.04 & UniDrive-VLA      & 54.87 & $-11.2$ \\
InternVL3-8B   & 62.91 & RecogDrive        & 54.25 & $-8.7$ \\
Cosmos-Reason  & 70.56 & Alpamayo-1.5-10B  & 36.86 & $-33.7$ \\
\bottomrule
\end{tabular}
\end{table}

\paragraph{Caveats.}
ImpromptuVLA-7B is built on Qwen2.5-VL but the vanilla Qwen2.5-VL is not in our evaluation list, so we cannot directly compute its same-backbone $\Delta$. Its absolute QA accuracy of 69.27\% is competitive with Cosmos-Reason-2 (70.56\%, also Qwen2.5-VL based) and meaningfully above the other driving VLAs - suggesting limited or no causal-QA loss - but this is suggestive evidence rather than a controlled comparison. The backbones of SafeAuto and SimLingo were not surfaced in our reading of the source papers; both sit in the moderately-degraded band of the leaderboard but their specific backbones cannot be confirmed. WiseAD, Orion, and Omnidrive use older or lightweight backbones (MobileVLM, Vicuna v1.5, LLaVA-7B) that are weaker than the Qwen3 / InternVL3 / Cosmos family, so part of their causal-QA performance is inherited from their backbone choice rather than caused by their driving fine-tuning.

\subsection{Post-training Reward Design and Causal-QA Degradation}
\label{app:rft_analysis}

Three of the driving VLAs in our list (Alpamayo-1.5-10B, OpenREAD, RecogDrive) are post-trained with reinforcement fine-tuning (RFT) on top of supervised fine-tuning, and they design their RFT rewards in three structurally distinct ways. The differences in reward design correspond closely to the differences in causal-QA degradation observed in \cref{app:backbone_analysis}.

\begin{table}[h!]
\centering
\caption{RFT reward design across the three RFT-trained driving VLAs in our roster, alongside the resulting same-backbone causal-QA degradation. The reward composition column lists the components of the RFT reward signal; the policy column indicates which model parameters are updated during RFT.}
\label{tab:rft_design}
\resizebox{\textwidth}{!}{%
\begin{tabular}{lllr}
\toprule
Model & RFT reward composition & Policy updated & $\Delta$ QA (pp) \\
\midrule
Alpamayo-1.5-10B & Reasoning quality + \emph{reasoning--action consistency} + trajectory L2/collision/jerk & Whole VLA (GRPO) & $-33.7$ \\
RecogDrive       & NAVSIM PDMS (collisions, drivable area, comfort) only       & Diffusion planner only (DiffGRPO; VLM frozen) & $-8.7$ \\
OpenREAD         & \emph{LLM-as-critic on QA semantic match} + embedding similarity + trajectory ADE & Whole VLA (GRPO) & $-2.3$ \\
\bottomrule
\end{tabular}}
\end{table}

\paragraph{The reasoning--action consistency reward as a probable cause of catastrophic degradation.}
Alpamayo's reward function explicitly grades how well its predicted trajectory matches its own reasoning trace - a ``consistency'' reward - on top of separate reasoning-quality and trajectory-quality components~\citep[§5.3.2]{wang2025alpamayo}. Their own ablation (Table~9 of the cited work) shows that optimising for reasoning quality \emph{alone} actually degrades trajectory ADE; the consistency reward is introduced to prevent this and works as designed within their evaluation distribution. The trade-off, however, is that the consistency reward necessarily anchors reasoning to in-distribution trajectory patterns: any reasoning that diverges from the family of explanations supporting typical motion is penalised, including reasoning that is correct on out-of-distribution causal queries. We hypothesise that this is the mechanism behind Alpamayo's 33.7-percentage-point causal-QA loss - the largest in the roster - from a base (Cosmos-Reason) that itself reaches 70.56\%.

\paragraph{The QA-as-critic reward as a candidate antidote.}
OpenREAD uses Qwen3-LLM as an automatic judge that scores whether the predicted answer to an open-ended driving knowledge query semantically matches a reference answer~\citep[§3.3.2]{zhang2025openread}. Critically, this QA-quality reward is \emph{not} coupled to the trajectory output: the model is rewarded for getting the answer right regardless of the trajectory it would produce. The trajectory-quality reward (exponential ADE) is a separate component. The result is a 2.3-pp loss on causal QA from the same Qwen3-VL-8B backbone - an order of magnitude smaller than Alpamayo's degradation, despite both models receiving heavy RFT.

\paragraph{The frozen-VLM control case.}
RecogDrive applies DiffGRPO only to its diffusion-based trajectory planner, leaving the VLM frozen during RFT~\citep[§4.3]{li2025recogdrive}. The RFT reward is the NAVSIM Predictive Driver Model Score, a simulator metric covering collisions, drivable-area compliance, and comfort. The language head therefore receives no direct RFT signal; its causal-QA degradation ($-8.7$\,pp from InternVL3-8B base) is attributable entirely to the upstream Stage-1 supervised fine-tuning on driving QA data. This middle outcome shows that selective RFT on the action channel does not catastrophically degrade the language channel, but also does not help it.

\paragraph{Caveats and scope.}
The above is an interpretation grounded in (i) the reward designs reported in each paper, (ii) Alpamayo's own ablation diagnosing the SFT-only and reasoning-only failure modes that motivated its consistency reward, and (iii) the empirical correlation between reward composition and causal-QA outcome across just three samples. It is not a controlled experiment within our paper - we do not retrain any model with a swapped reward function. The hypothesis is that reward composition - particularly whether the reward couples reasoning to in-distribution action or rewards open-ended QA correctness - is a controllable design lever that practitioners can use to tune the language--action trade-off. \cdb is positioned to be the supervisory signal for the latter: a causal-reasoning-aware reward that explicitly grades collider-bias suppression, distractor rejection, and counterfactual consistency - none of which is targeted by current driving VLAs.

\subsection{Cross-cutting Analyses}
\label{app:cross_cutting}

Three quantitative observations support the main-paper claims and may be useful to reviewers seeking to interpret the leaderboard.

\paragraph{Binary accuracy determines the overall ranking.} Pearson correlation between binary-only accuracy and overall accuracy is 0.95 across the 13 models, while Pearson(MCQ-only, overall) is 0.67. Although MCQ accounts for 16.7\% of questions, the binary slice essentially fixes the ranking. Orion's overall rank of 6 of 13 (57.25\%) is artificially deflated by a 0.0\% MCQ score; on binary alone it ranks 4th at 68.7\%, between Qwen3-VL-8B and InternVL-3.5-8B. We attribute Orion's MCQ score to an output-parsing failure on the A/B/C/D format rather than a reasoning failure.

\paragraph{Active--Distractor direction is family-stratified.} Driving VLAs and VLMs separate cleanly on the sign of the Active--Distractor gap. Across the 10 driving VLAs the mean gap is $+6.8$\,pp (Active above Distractor), with 8 of 10 between $+9$ and $+24$\,pp. Across the 3 VLMs the mean gap is $-6.5$\,pp, with two of three favouring Distractor. Driving-specific training inverts the correct direction of causal discrimination: VLAs over-attribute causal status to perceptually salient irrelevant entities, while VLMs either remain neutral or over-reject. This sign inversion is corroborating evidence for the heterogeneous-degradation pattern documented in \cref{app:backbone_analysis}: even when a driving VLA shares its backbone with a VLM, the post-training procedure can reverse the direction of causal-discrimination bias at the language head.

\paragraph{Trajectory regression and causal QA are statistically independent.} Pearson(BTE, overall VQA accuracy) is approximately 0.00 across the 13 models. UniDrive-VLA achieves the best BTE (0.68\,m) but ranks 7th on QA; Cosmos-Reason-2 ranks 5th on BTE (1.79\,m) but 1st on QA; InternVL-3.5-8B ranks 12th on BTE (7.71\,m) but 5th on QA. The mean rank shift between the two orderings is 4.5 positions out of 13. This independence supports the architectural reading in \cref{app:attention_masking}: the trajectory head and the language head are trained as separate capabilities and the language head is not integrated into trajectory generation in current driving VLAs.

\subsection{Attention Masking Experiments}
\label{app:attention_masking}

\begin{figure}[h!]
\centering
\includegraphics[width=0.85\textwidth]{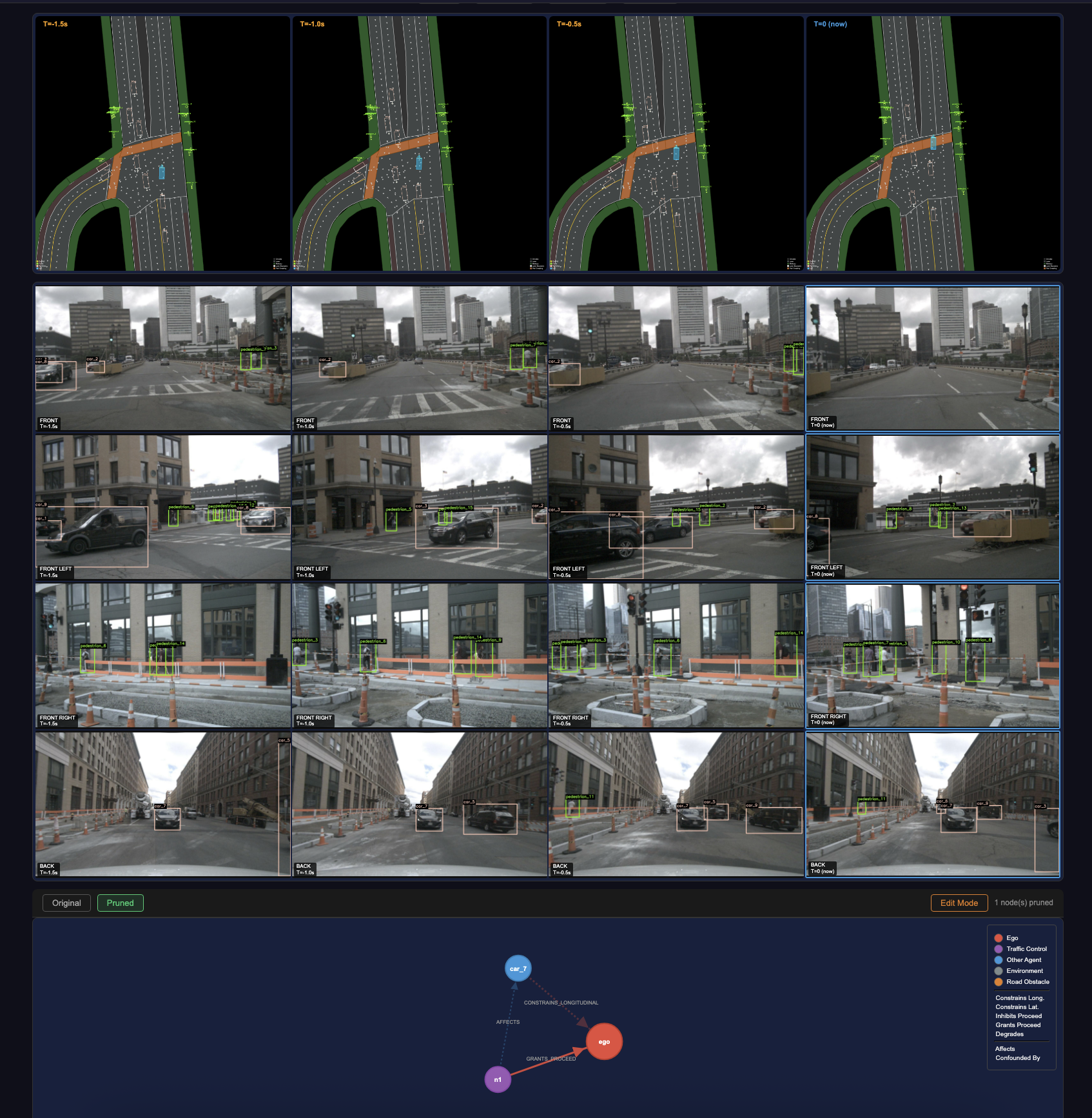}
\caption{Scene used for the attention-masking probe. The ego vehicle is accelerating straight through a green signal at an open intersection, with the multi-view camera frames and ego-history tokens encoding this factual configuration. This scene is used as the substrate for both the reasoning-on injection sweep (\cref{fig:mask_reason_on}) and the reasoning-off ablation sweep (\cref{fig:mask_reason_off}).}
\label{fig:mask_scene}
\end{figure}

The CF-trajectory ADE results in \cref{app:trajectory_eval} establish that counterfactual premises supplied through the reasoning channel fail to produce proportionate changes in predicted trajectories. To localise the cause of this failure architecturally, we run an attention-masking probe on four representative models, two driving VLAs (AlpaMayo-1.5, OpenREAD) and two general-purpose VLMs (Qwen3-VL-8B-thinking, Cosmos-Reason2-8B), on a controlled scene in which the ego vehicle is accelerating straight through a green traffic signal (\cref{fig:mask_scene}). The probe selectively masks each of the three input modalities the model receives, vision tokens $V$ (multi-view camera frames), ego-history tokens $E$ (past trajectory and state), and reasoning tokens $R$ (the chain-of-thought rationale). For the reasoning channel we additionally compare the un-masked baseline against a set of \emph{injected counterfactual rationales} $R'$ that re-describe the same scene under different alternative behaviours, and observe how the predicted trajectory $T'$ responds.

\begin{figure*}[h]
\centering
\includegraphics[width=\textwidth]{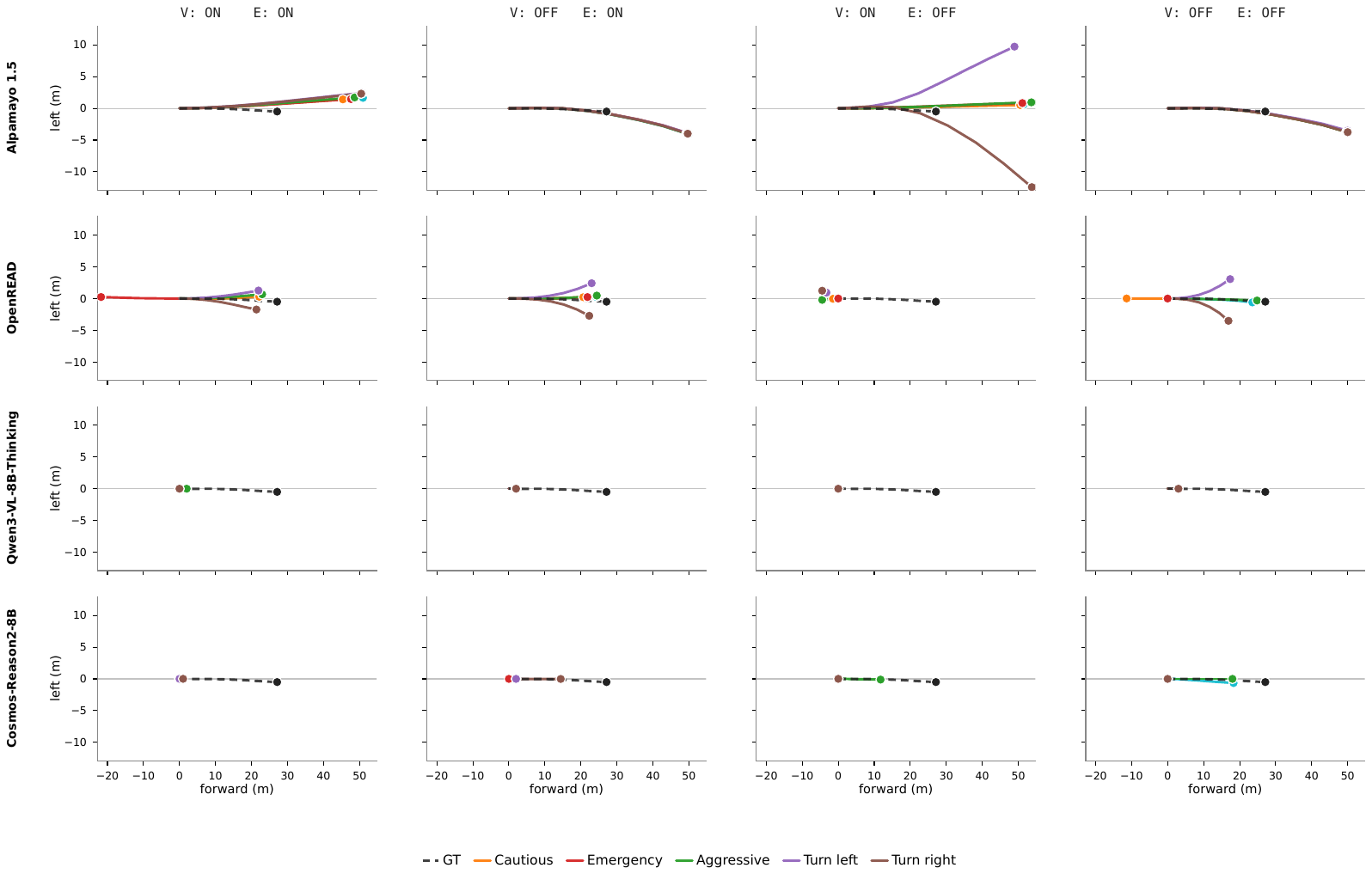}
\caption{Reasoning-on regime with rationale injection. Predicted trajectories for the un-masked baseline (teal) and the five injected counterfactual rationales \textsc{cautious}, \textsc{emergency}, \textsc{aggressive}, \textsc{turn-left}, \textsc{turn-right} (coloured) versus nuScenes ground truth (dashed black) under the four $(V, E)$ mask configurations across the four evaluated models. A faithfully-conditioned trajectory head should produce a visible fan of five distinct curves with \textsc{emergency} decelerating and \textsc{turn-left} and \textsc{turn-right} bending laterally. For AlpaMayo-1.5 the curves remain bunched in every configuration, including $V$:\textsc{off}, $E$:\textsc{off}; for OpenREAD a visible fan emerges only once vision is masked, and grows further once ego history is also removed. Qwen3-VL and Cosmos-Reason collapse to degenerate outputs rather than fanning, indicating that the rationale never structurally drives their trajectory head either.}
\label{fig:mask_reason_on}
\end{figure*}

\paragraph{Reasoning-on regime (rationale injection).} With $R$ retained, we replace the model's own chain-of-thought with each of five counterfactual rationales $R'$ that are mutually inconsistent with the factual scene: \textsc{cautious} (a pedestrian is approaching the crosswalk), \textsc{emergency} (the signal has turned red, brake immediately), \textsc{aggressive} (accelerate through), \textsc{turn-left}, and \textsc{turn-right}. Each rationale prescribes a qualitatively different trajectory. A model whose trajectory head genuinely consumes the rationale should produce a visible fan of five distinct $T'$ curves, with \textsc{emergency} decelerating, \textsc{turn-left} and \textsc{turn-right} bending laterally, and so on. Figure~\ref{fig:mask_reason_on} shows the outcome across the four $(V, E)$ mask configurations. With both vision and ego history available ($V$:\textsc{on}, $E$:\textsc{on}), the five counterfactual trajectories collapse onto each other for both AlpaMayo-1.5 and OpenREAD, none of the injected rationales meaningfully bend the trajectory away from the model's vision-derived prediction. The collapse persists when ego history is removed but vision remains. Only when vision is masked does OpenREAD begin to produce a visible spread among the counterfactual trajectories, and only when both vision and ego history are removed does the spread approach what a faithfully-conditioned rationale would imply. AlpaMayo-1.5 shows no measurable spread in any configuration, indicating that its trajectory head reads vision and ego history but never the rationale. Qwen3-VL and Cosmos-Reason exhibit a related but different failure mode: once vision is masked, their trajectories collapse to degenerate or near-zero outputs rather than fanning according to $R'$, suggesting that the rationale never enters the trajectory pathway in a structured way for these models either.

\begin{figure*}[h]
\centering
\includegraphics[width=\textwidth]{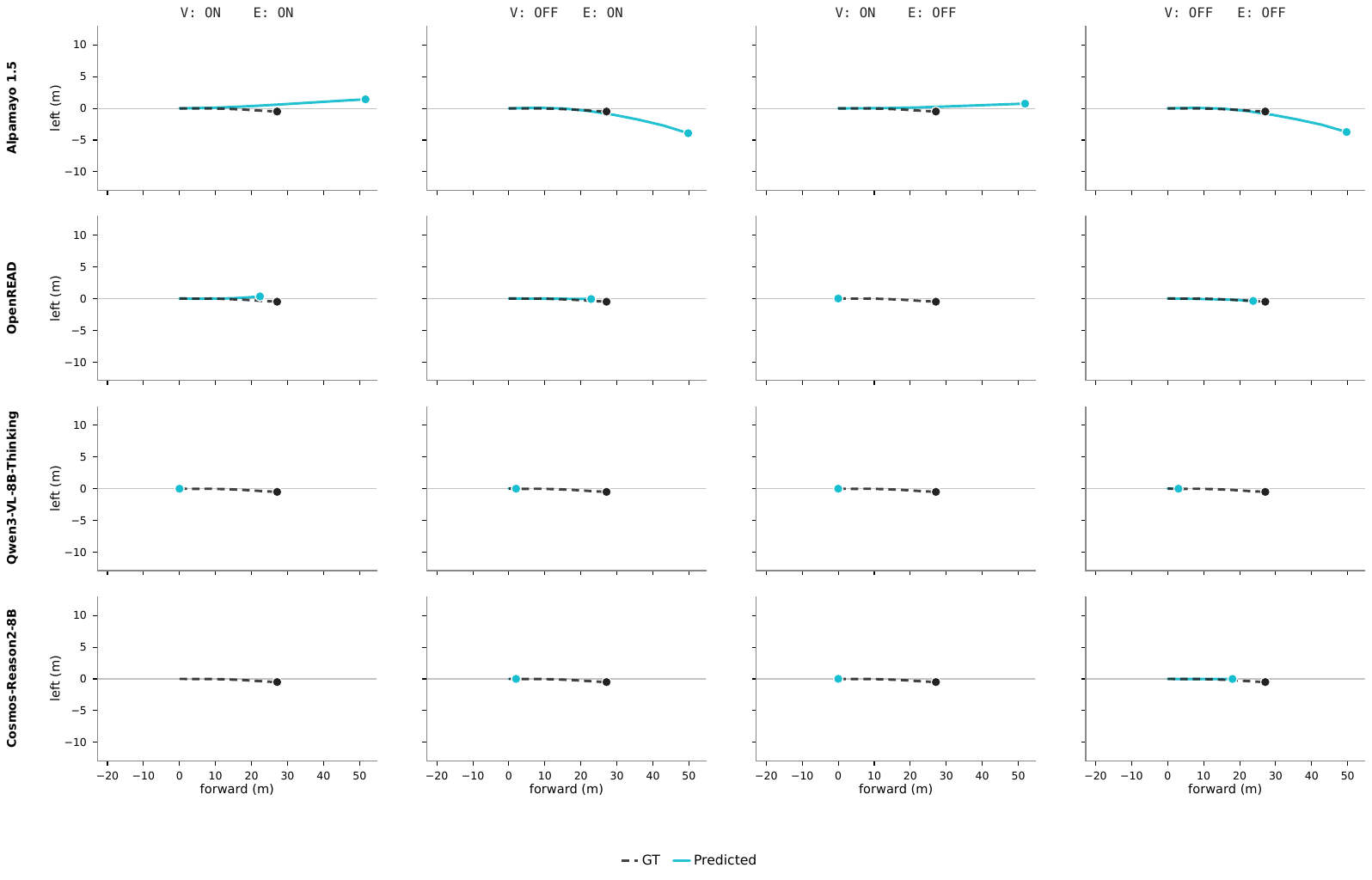}
\caption{Reasoning-off regime with the rationale ablated. Predicted trajectories (teal) versus nuScenes ground truth (dashed black) under the four $(V, E)$ mask configurations across the four evaluated models. Removing the rationale entirely while preserving vision and ego history leaves the prediction nearly aligned with ground truth for both driving VLAs, confirming that the rationale is not a load-bearing input for trajectory generation. Deviations escalate only when vision is also removed, and degenerate when all three modalities are masked.}
\label{fig:mask_reason_off}
\end{figure*}

\paragraph{Reasoning-off regime ($R$ ablated).} With $R$ masked entirely, we compare the resulting $T'$ against the nuScenes ground-truth trajectory $T$. Figure~\ref{fig:mask_reason_off} shows that removing the rationale while preserving vision and ego history ($V$:\textsc{on}, $E$:\textsc{on}, $R$:\textsc{off}) produces only minor deviations from $T$ for both driving VLAs, and for AlpaMayo-1.5 the prediction is essentially indistinguishable from the un-masked baseline. Deviations grow as further modalities are removed, with $V$:\textsc{off}, $E$:\textsc{on} producing large lateral and longitudinal errors, and the fully-masked configuration ($V$:\textsc{off}, $E$:\textsc{off}, $R$:\textsc{off}) collapsing to degenerate or hallucinated outputs that confirm the trajectory head is not falling back on a content-free prior.

Table~\ref{tab:mask_summary} compresses the qualitative pattern into a single view across the two driving VLAs and the two general-purpose VLMs.

\begin{table}[h!]
\centering
\small
\setlength{\tabcolsep}{4pt}
\renewcommand{\arraystretch}{1.15}
\caption{Qualitative summary of the attention-masking outcomes across the eight $(V, E, R)$ configurations for the two driving VLAs and the two general-purpose VLMs. Reasoning-on rows report whether the five injected counterfactual rationales $R'$ cause the predicted trajectory $T'$ to fan out into qualitatively distinct curves (more $\Rightarrow$ greater rationale influence). Reasoning-off rows report how far the predicted $T'$ deviates from ground-truth $T$ when the rationale is ablated (less $\Rightarrow$ closer to $T$). \emph{Hallucinates} marks configurations where the trajectory collapses to degenerate, near-zero, or unstructured output.}
\label{tab:mask_summary}
\resizebox{\textwidth}{!}{%
\begin{tabular}{ccc l l l l}
\toprule
\multicolumn{3}{c}{Inputs} & \multicolumn{2}{c}{Driving VLAs} & \multicolumn{2}{c}{General-purpose VLMs} \\
\cmidrule(lr){1-3} \cmidrule(lr){4-5} \cmidrule(lr){6-7}
$V$ & $E$ & $R$ & OpenREAD & AlpaMayo-1.5 & Qwen3-VL-8B-Thinking & Cosmos-Reason2-8B \\
\midrule
\multicolumn{7}{l}{\textit{Reasoning on with injected $R'$: spread of $T'$ across counterfactual rationales}}\\
\midrule
\checkmark & \checkmark & \checkmark & No effect    & No effect & No effect      & No effect \\
\checkmark & \ding{55}  & \checkmark & No effect    & No effect & No effect      & No effect \\
\ding{55}  & \checkmark & \checkmark & Some effect  & No effect & Hallucinates   & Hallucinates \\
\ding{55}  & \ding{55}  & \checkmark & Good effect  & No effect & Hallucinates   & Hallucinates \\
\midrule
\multicolumn{7}{l}{\textit{Reasoning off: deviation of $T'$ from ground-truth $T$}}\\
\midrule
\checkmark & \checkmark & \ding{55}  & Some deviation                & No deviation                  & Hallucinates  & Hallucinates \\
\checkmark & \ding{55}  & \ding{55}  & Some deviation                & Some deviation                & Hallucinates  & Hallucinates \\
\ding{55}  & \checkmark & \ding{55}  & Large deviation               & Large deviation               & Hallucinates  & Hallucinates \\
\ding{55}  & \ding{55}  & \ding{55}  & Large deviation, hallucinates & Large deviation, hallucinates & Hallucinates  & Hallucinates \\
\bottomrule
\end{tabular}%
}
\end{table}

\paragraph{Interpretation: vision and ego history dominate; reasoning is ornamental.} The two regimes triangulate the same conclusion for the driving VLAs. In the reasoning-on regime, injecting rationales that prescribe mutually inconsistent behaviours (emergency braking, lateral turning, aggressive acceleration) over the same scene produces nearly identical trajectories whenever vision or ego history is available, so $R$ is not a primary signal driving $T'$. In the reasoning-off regime, ablating $R$ while preserving $V$ and $E$ leaves the trajectory nearly intact for both VLAs, so $R$ is not a contributing signal either. The trajectory head is therefore reading $T'$ predominantly from $V$ and $E$, and the chain-of-thought rationale is functionally decorative with respect to action. AlpaMayo-1.5 is the more vision-and-ego-history-dominated of the two, with no measurable rationale influence in any configuration; OpenREAD shows a small residual coupling that surfaces only once vision has been removed. The general-purpose VLMs exhibit a distinct failure mode: their trajectories bunch under rationale injection in the same way the VLAs' do (so $R'$ does not steer the trajectory faithfully), but they collapse to hallucinated or near-zero outputs whenever vision is masked or the rationale is ablated, indicating that the trajectory pathway in these models is brittle to either modality being removed rather than smoothly redistributing reliance across the remaining inputs.

\paragraph{Connection to CF-trajectory ADE.} This explains the gap reported in \cref{app:trajectory_eval}. A counterfactual premise supplied to the model enters through the same reasoning channel that the masking probe shows to be inert with respect to trajectory whenever vision and ego history are present. Even when the premise is explicit, well-formed, and accompanied by the corresponding spatial context, the trajectory head continues to read $V$ and $E$ as the dominant signal, both of which still encode the factual scene rather than the counterfactual one. The resulting $T'$ is therefore closer to the original observed trajectory than to the ground-truth counterfactual trajectory, which is precisely the deviation the CF-ADE numbers capture. The masking experiment thus identifies the architectural locus of the failure: language reasoning is produced in parallel with, rather than upstream of, trajectory generation in current driving VLAs, so changes in the language or in the counterfactual premise do not translate into corresponding changes in action.

\subsection{Representational Analysis of VLM\,$\to$\,VLA Fine-tuning}
\label{app:representational_analysis}

The accuracy gaps in \cref{tab:vqa_results} show that Alpamayo-1.5-10B trails its parent Cosmos-Reason-2 by $35$--$47$\,pp on the lower rungs of the ladder, while OpenREAD trails Qwen3-VL-8B by at most $16$\,pp and on $R_0$ outperforms it. To understand where in the network this gap comes from, we run a layer-wise representational probe on the same query sets used for accuracy evaluation across all four rungs ($R_0$--$R_3$). For each parent--child pair, Cosmos-Reason-2\,$\to$\,Alpamayo-1.5-10B and Qwen3-VL-8B\,$\to$\,OpenREAD, we extract per-layer activations and compute three quantities at every layer: how many directions of variation the layer uses, how much the model's predicted next-token distribution changes from one layer to the next, and how far the child layer has drifted from the parent. Layer-by-layer comparison is principled because each VLA inherits the parent's transformer stack one-for-one. Let $H_\ell \in \mathbb{R}^{n \times d}$ denote the matrix of per-input activations at layer $\ell$ for $n$ inputs and embedding dimension $d$, with $H_\ell^c$ its mean-centred version.

\paragraph{Effective rank (ER).} ER \citep{roy2007effective} measures how many directions of the embedding space a layer's activations actually occupy. Let $\sigma_1 \ge \sigma_2 \ge \cdots \ge \sigma_r$ be the singular values of $H_\ell^c$ and define the variance fractions $p_i = \sigma_i^2 / \sum_j \sigma_j^2$. Then
\begin{equation}
\mathrm{ER}(\ell) \;=\; \exp\!\big(H(p)\big)
\;=\; \exp\!\Big(\!-\!\sum_i p_i \log p_i\Big),
\label{eq:er}
\end{equation}
where $H(p)$ is the Shannon entropy of the variance distribution. If all variance lies along one singular direction, $\mathrm{ER}=1$ and every input is encoded as a scalar position along a single line in embedding space. If variance is spread evenly over $k$ directions, $\mathrm{ER}=k$. ER thus reads as an effective dimensionality: the number of independent directions the layer is actually using to encode information.

\paragraph{Fisher--Rao length ($\widetilde{\mathcal{L}}$).} $\widetilde{\mathcal{L}}$ \citep{amari2016information} measures how much the model's predicted next-token distribution shifts from one layer to the next. We project each hidden state through the unembedding matrix $W_U$ to obtain a per-layer logit vector $z_\ell = h_\ell W_U^\top$, take a softmax to get a next-token distribution $p_\ell = \mathrm{softmax}(z_\ell)$, and measure how far $p_\ell$ moves to $p_{\ell+1}$ on the statistical manifold. To keep the comparison computable, we restrict to the union of top-$K$ tokens at the two layers ($K=2048$) and renormalise:
\begin{equation}
S \;=\; \mathrm{top}\text{-}K(p_\ell) \cup \mathrm{top}\text{-}K(p_{\ell+1}),
\qquad
\tilde p_\ell \;=\; p_\ell[S] \,/\, \textstyle\sum_{i \in S} p_\ell[i].
\end{equation}
The Fisher--Rao distance between the two distributions on this support is
\begin{equation}
\widetilde{\mathcal{L}}(\ell) \;=\; \frac{2}{\pi}\,\arccos\!\Big(\textstyle\sum_{i \in S} \sqrt{\tilde p_\ell[i]\,\tilde p_{\ell+1}[i]}\Big),
\label{eq:fr}
\end{equation}
normalised to $[0,1]$ by the $1/\pi$ factor. The inner sum measures the overlap between the two distributions, and $\arccos$ turns that overlap into a distance. Low $\widetilde{\mathcal{L}}$ means the layer is barely changing what the model thinks comes next; high $\widetilde{\mathcal{L}}$ means the layer is making a substantial update to the predicted token.

\paragraph{Cross-model CKA distance ($D_{\mathrm{CKA}}$).} $D_{\mathrm{CKA}}$ \citep{kornblith2019similarity} measures how far the child layer has drifted from the parent at the same depth. Given the parent's mean-centred activations $X^c \in \mathbb{R}^{n \times d_1}$ and the child's $Y^c \in \mathbb{R}^{n \times d_2}$ on the same $n$ inputs, linear CKA is
\begin{equation}
\mathrm{CKA}(X^c, Y^c)
\;=\;
\frac{\big\| (X^c)^\top Y^c \big\|_F^2}
     {\big\| (X^c)^\top X^c \big\|_F\;\big\| (Y^c)^\top Y^c \big\|_F},
\qquad
D_{\mathrm{CKA}}(\ell) \;=\; 1 - \mathrm{CKA}\big(X^c_\ell, Y^c_\ell\big).
\label{eq:cka}
\end{equation}
The numerator is the squared Frobenius norm of the cross-covariance between the two representations; the denominator normalises by each model's own representational norm. CKA is invariant to orthogonal transformations and isotropic scaling, so $D_{\mathrm{CKA}}$ is not fooled by trivial reparameterisations: a high value means the parent and child layers genuinely encode different information about the same inputs, not just the same information in a different basis.

\begin{figure*}[h]
\centering
\begin{minipage}[t]{0.485\textwidth}
  \centering
  \includegraphics[width=\linewidth]{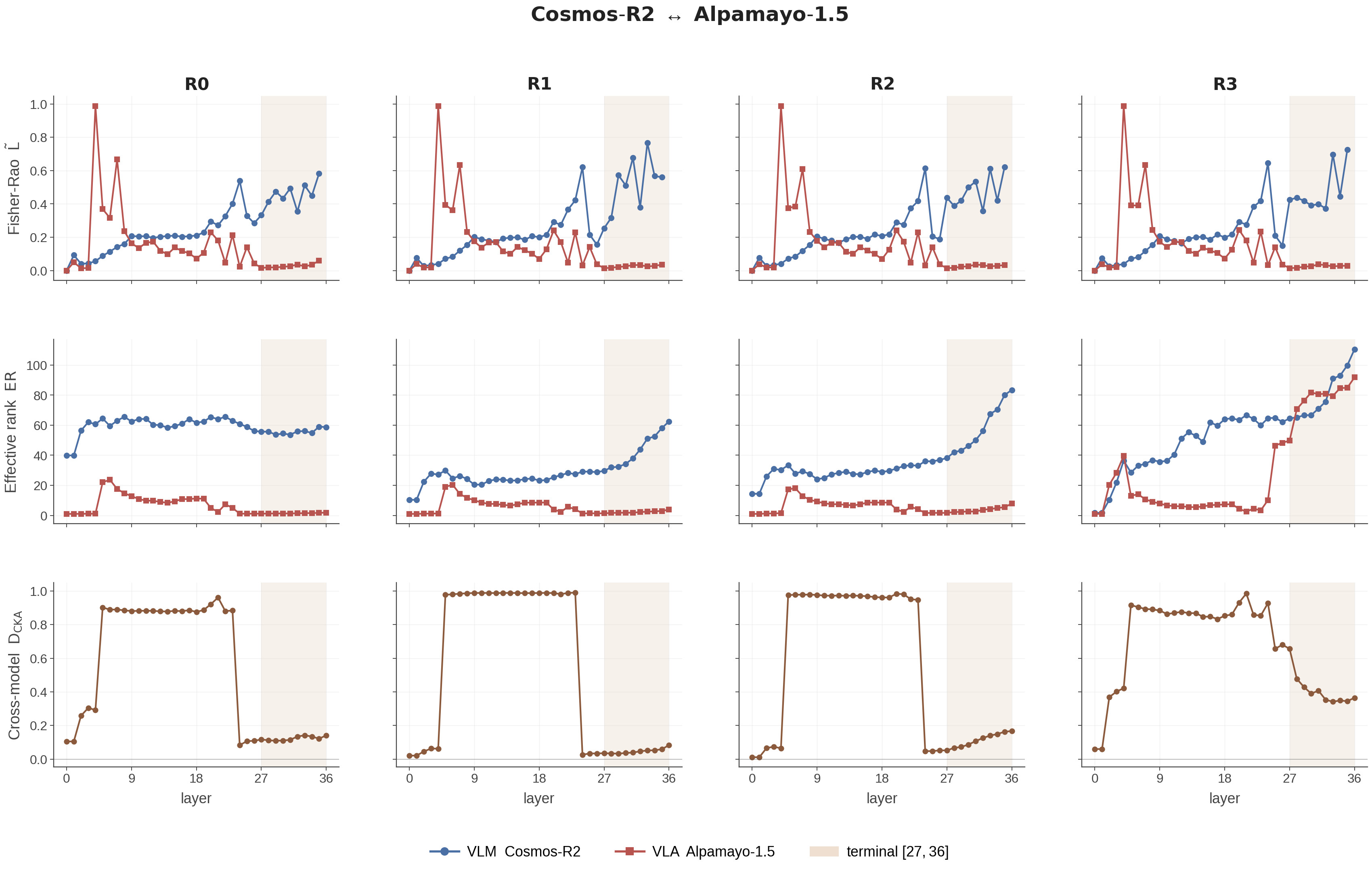}
  \subcaption{Cosmos-Reason-2 $\leftrightarrow$ Alpamayo-1.5-10B.}
  \label{fig:rep_cosmos_alpamayo}
\end{minipage}\hfill
\begin{minipage}[t]{0.485\textwidth}
  \centering
  \includegraphics[width=\linewidth]{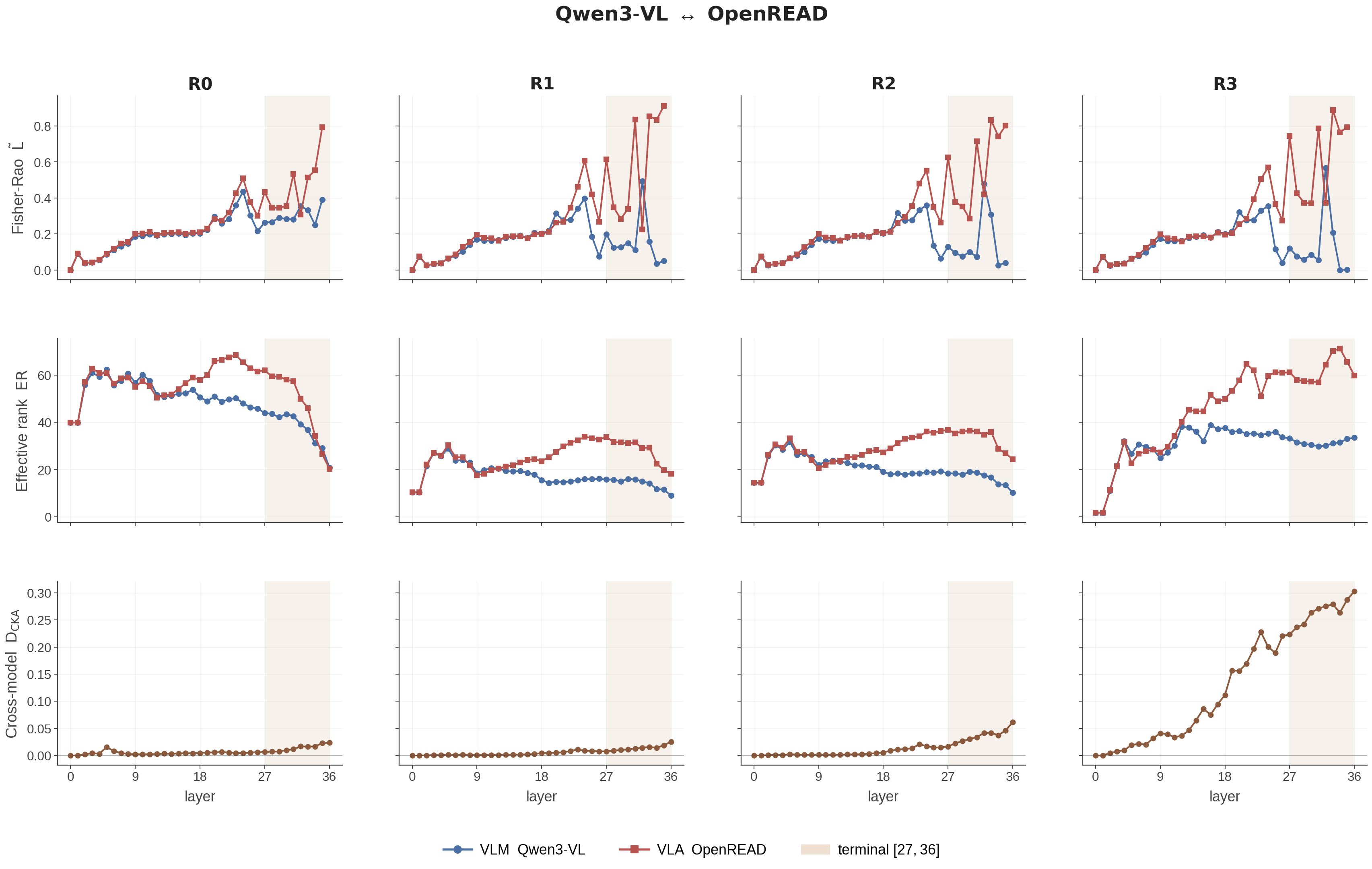}
  \subcaption{Qwen3-VL-8B $\leftrightarrow$ OpenREAD.}
  \label{fig:rep_qwen_openread}
\end{minipage}
\caption{Layer-wise representational metrics across $R_0$--$R_3$ for the two parent--child pairs. Rows in each $3\times 4$ panel: per-layer Fisher--Rao length between consecutive layers, effective rank, cross-model $D_{\mathrm{CKA}}$. Columns: $R_0$ (association), $R_1$, $R_2$ (intervention), $R_3$ (counterfactual). Shaded: terminal block, layers $27$--$36$. \textbf{(a)} On $R_0$--$R_2$ Alpamayo's terminal block is collapsed: ER and Fisher--Rao length sit at the floor, and $D_{\mathrm{CKA}}$ is small. The terminal layers have stopped contributing to the next-token prediction and lie on a near-$1$D subspace inside the parent's dominant directions. On $R_3$ ER climbs back to parent levels and $D_{\mathrm{CKA}}$ rises sharply, but Fisher--Rao length stays low: the terminal block has high-rank activations that no longer translate into changes in the predicted output. \textbf{(b)} OpenREAD's terminal effective rank and Fisher--Rao length are at or above the parent on every rung, and $D_{\mathrm{CKA}}$ grows monotonically with depth, with the magnitude of divergence increasing from $R_0$ to $R_3$. The contrast between (a) and (b) shows that the collapse is recipe-specific.}
\label{fig:rep_sidebyside}
\end{figure*}

Together the three quantities tell us what each layer has become: ER tells us its capacity, $\widetilde{\mathcal{L}}$ tells us whether it is still actively updating the model's prediction, and $D_{\mathrm{CKA}}$ tells us whether it still resembles the parent. We refer to the final ten layers as the \textbf{terminal block} (shaded in all panels of \cref{fig:rep_sidebyside}).

\paragraph{Finding 1: Alpamayo's last layers stop doing useful work on $R_0$--$R_2$.} On the lower three rungs, Alpamayo's terminal block has been compressed almost to nothing. Visually, this is the flat red lines hugging the bottom of the ER and Fisher--Rao panels in \cref{fig:rep_sidebyside}(a) on $R_0$, $R_1$, $R_2$. What the three metrics together tell us is that the layer has lost its ability to encode anything meaningful: every input gets squeezed onto roughly a single direction in the embedding space, and the layer is no longer updating the model's predicted next token from layer to layer. Cross-model CKA shows the line the child is using lies inside the parent's dominant subspace, so the child has not replaced the parent's terminal computation with something new, it has just kept a thin slice of it. Looking at the middle layers (5--23) on the same plot, $D_{\mathrm{CKA}}$ is high (around $0.9$), meaning fine-tuning has rewritten the middle of the network heavily. The combined picture is a familiar one: the model is doing the new driving-relevant work in its middle layers and using the terminal block as a narrow funnel that idles on the prediction the middle has already produced. Quantitatively, terminal-block ER means stay between $1.4$ and $3.7$ on these rungs, against the parent's $43$--$58$, with all values reported in \cref{tab:repr_summary}.

\paragraph{Finding 2: On $R_3$, the terminal block uses many directions but stops contributing to the output.} The rightmost column of \cref{fig:rep_sidebyside}(a) looks completely different. The Alpamayo ER curve climbs through the terminal block instead of hugging the floor, and $D_{\mathrm{CKA}}$ rises into the same range as the middle-layer plateau. So on counterfactual queries the terminal block has expanded back out into a high-dimensional embedding space and that space is no longer aligned with the parent's. But the Fisher--Rao panel tells the cautionary part of the story: the red line stays as flat as it was on $R_0$--$R_2$. The layer's hidden state is now varying across many directions, but those variations no longer translate into changes in the model's predicted next token. The terminal block is producing structurally rich activations that have stopped feeding into the output. Whether this reflects partially-recovered reasoning machinery that simply does not couple to the unembedding, or capacity recruitment without genuine computation, is not resolved by the plot, but the dissociation between embedding-space movement (high ER, high $D_{\mathrm{CKA}}$) and output-space movement (low $\widetilde{\mathcal{L}}$) is clear from it.

\paragraph{Finding 3: OpenREAD looks the opposite. The collapse is specific to Alpamayo.} \Cref{fig:rep_sidebyside}(b) shows a qualitatively different pattern from start to finish. Across all four rungs, OpenREAD's terminal red line sits at or above the blue parent line on both ER and Fisher--Rao, and the $D_{\mathrm{CKA}}$ curve rises smoothly with depth instead of dropping off. So OpenREAD's last layers are using lots of directions, those directions are still actively updating the model's predicted next token, and they have moved further from the parent the deeper you look. The amount of divergence in the terminal block also gets larger as the rung gets harder, as visible from the increasing height of the $D_{\mathrm{CKA}}$ curve from $R_0$ to $R_3$ in panel (b). The model has rebuilt its terminal block into something different from the parent rather than throwing it away. Because OpenREAD is also a VLA fine-tuned on driving data, the collapse seen in Alpamayo cannot be a generic effect of driving fine-tuning. It is specific to the Alpamayo training recipe.

\Cref{tab:repr_summary} summarises the per-rung terminal-block statistics for both pairs.

\begin{table}[h!]
\centering
\small
\setlength{\tabcolsep}{4pt}
\renewcommand{\arraystretch}{1.15}
\caption{Terminal-block (layers $27$--$36$) summary statistics for the two parent--child pairs across the four rungs of the benchmark. Each cell reports the mean over the ten terminal layers. ER is effective rank (higher $=$ more directions used); $\widetilde{\mathcal{L}}$ is Fisher--Rao length between consecutive layers (higher $=$ more change in next-token prediction); $D_{\mathrm{CKA}}$ is cross-model CKA distance (higher $=$ more divergence from the parent). Alpamayo's terminal block sits at the floor on $R_0$--$R_2$ and recovers on $R_3$; OpenREAD's terminal block exceeds the parent on every metric, on every rung.}
\label{tab:repr_summary}
\resizebox{\textwidth}{!}{%
\begin{tabular}{l ccc ccc ccc ccc}
\toprule
& \multicolumn{3}{c}{$R_0$} & \multicolumn{3}{c}{$R_1$} & \multicolumn{3}{c}{$R_2$} & \multicolumn{3}{c}{$R_3$} \\
\cmidrule(lr){2-4}\cmidrule(lr){5-7}\cmidrule(lr){8-10}\cmidrule(lr){11-13}
Model & ER & $\widetilde{\mathcal{L}}$ & $D_{\mathrm{CKA}}$ & ER & $\widetilde{\mathcal{L}}$ & $D_{\mathrm{CKA}}$ & ER & $\widetilde{\mathcal{L}}$ & $D_{\mathrm{CKA}}$ & ER & $\widetilde{\mathcal{L}}$ & $D_{\mathrm{CKA}}$ \\
\midrule
Cosmos-Reason-2 (VLM) & $55.8$ & $0.45$ & --- & $43.4$ & $0.51$ & --- & $57.7$ & $0.48$ & --- & $80.4$ & $0.48$ & --- \\
Alpamayo-1.5-10B (VLA) & $1.41$ & $0.029$ & $0.124$ & $2.25$ & $0.026$ & $0.047$ & $3.75$ & $0.026$ & $0.113$ & $78.1$ & $0.026$ & $0.411$ \\
\midrule
Qwen3-VL-8B (VLM) & $37.3$ & $0.30$ & --- & $14.0$ & $0.16$ & --- & $16.4$ & $0.15$ & --- & $31.5$ & $0.13$ & --- \\
OpenREAD (VLA) & $47.4$ & $0.46$ & $0.014$ & $27.8$ & $0.58$ & $0.014$ & $33.2$ & $0.57$ & $0.036$ & $62.3$ & $0.61$ & $0.265$ \\
\bottomrule
\end{tabular}%
}
\end{table}

\paragraph{Interpretation.} The two recipes leave qualitatively different fingerprints on the parent model. The Alpamayo recipe rewrites the middle layers and squeezes the terminal block into a low-rank projection of what the parent already had, with the terminal layers no longer contributing to the next-token prediction. The OpenREAD recipe does the opposite: the terminal block keeps using as many or more directions than the parent, those directions actively update the model's prediction, and the divergence from the parent grows smoothly with depth and with rung level. The two pairs are not two points on a continuum but two distinct outcomes: a low-rank terminal bottleneck in one, a high-rank reorganisation in the other.

\paragraph{Connection to accuracy.} Alpamayo trails Cosmos-Reason-2 by $42.7$, $47.0$, $34.6$\,pp on $R_0$, $R_1$, $R_2$, the rungs where its terminal block is collapsed. The gap shrinks to $18.3$\,pp on $R_3$, the rung where the terminal block recovers high-rank capacity (even though it does not yet feed into the output). OpenREAD shows neither collapse nor large gaps, and on $R_0$ the child outperforms the parent by $16.7$\,pp. Across both pairs, the rungs with degraded terminal capacity are also the rungs with the largest accuracy gaps.

\section{Annotation and Examples} 
\label{app:examples}
\subsection{Annotation Interface}
\label{app:annotation_interface}

\begin{figure}[h]
    \centering
    \includegraphics[width=\textwidth]{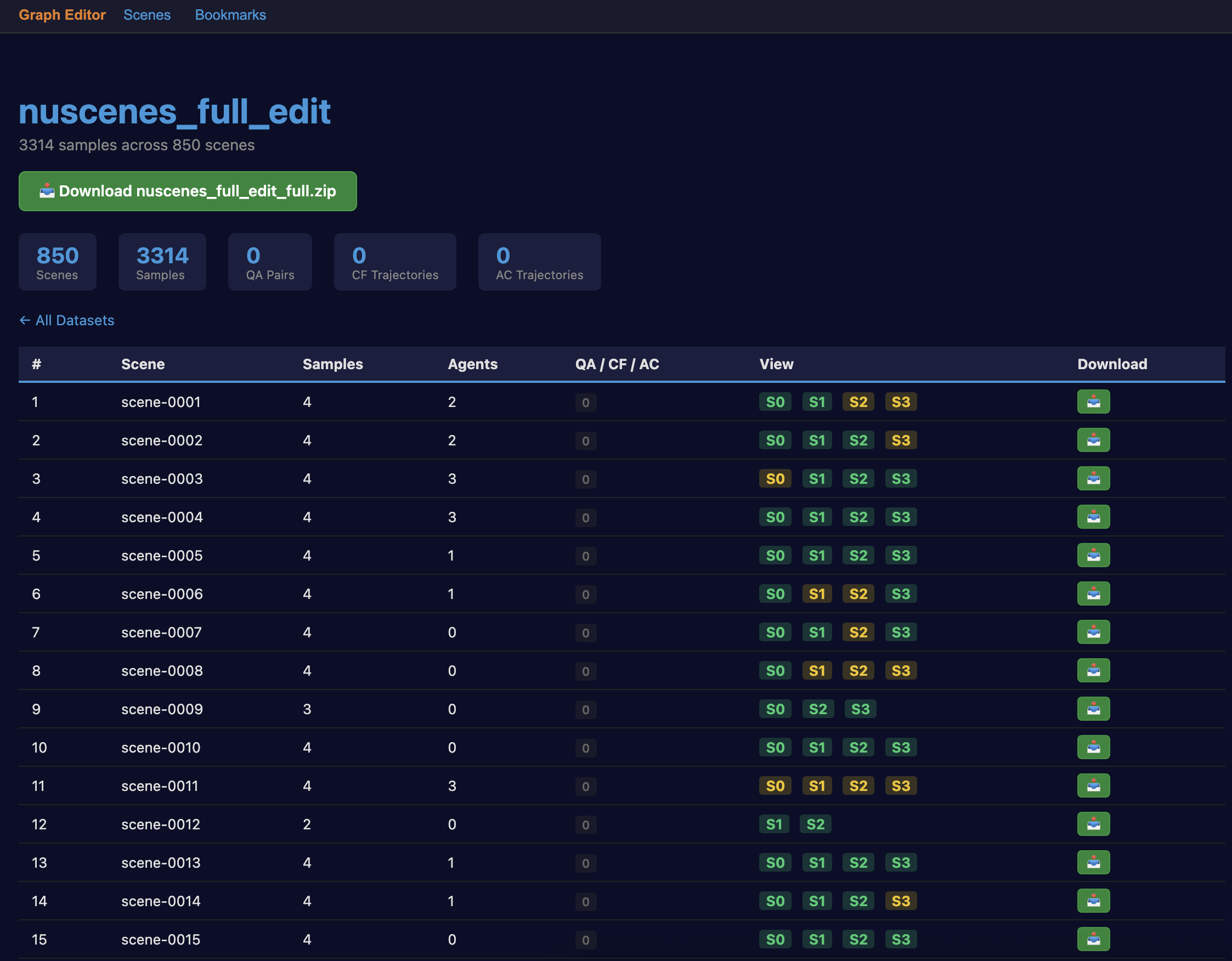}
    \caption{Dataset-wide review dashboard. The page lists all 850 nuScenes scenes alongside their per-sample review status. Each scene row carries four sample chips (S0--S3), with chip colour encoding review state (green for Good, amber for Needs Review or in-progress, grey for filtered). The header surfaces aggregate stat cards (scenes, samples, QA pairs, counterfactual trajectories, alternative-cause trajectories) and a one-click bulk download of the entire reviewed dataset.}
    \label{fig:edit_dashboard}
\end{figure}

\begin{figure}[h!]
    \centering
    \includegraphics[width=\textwidth]{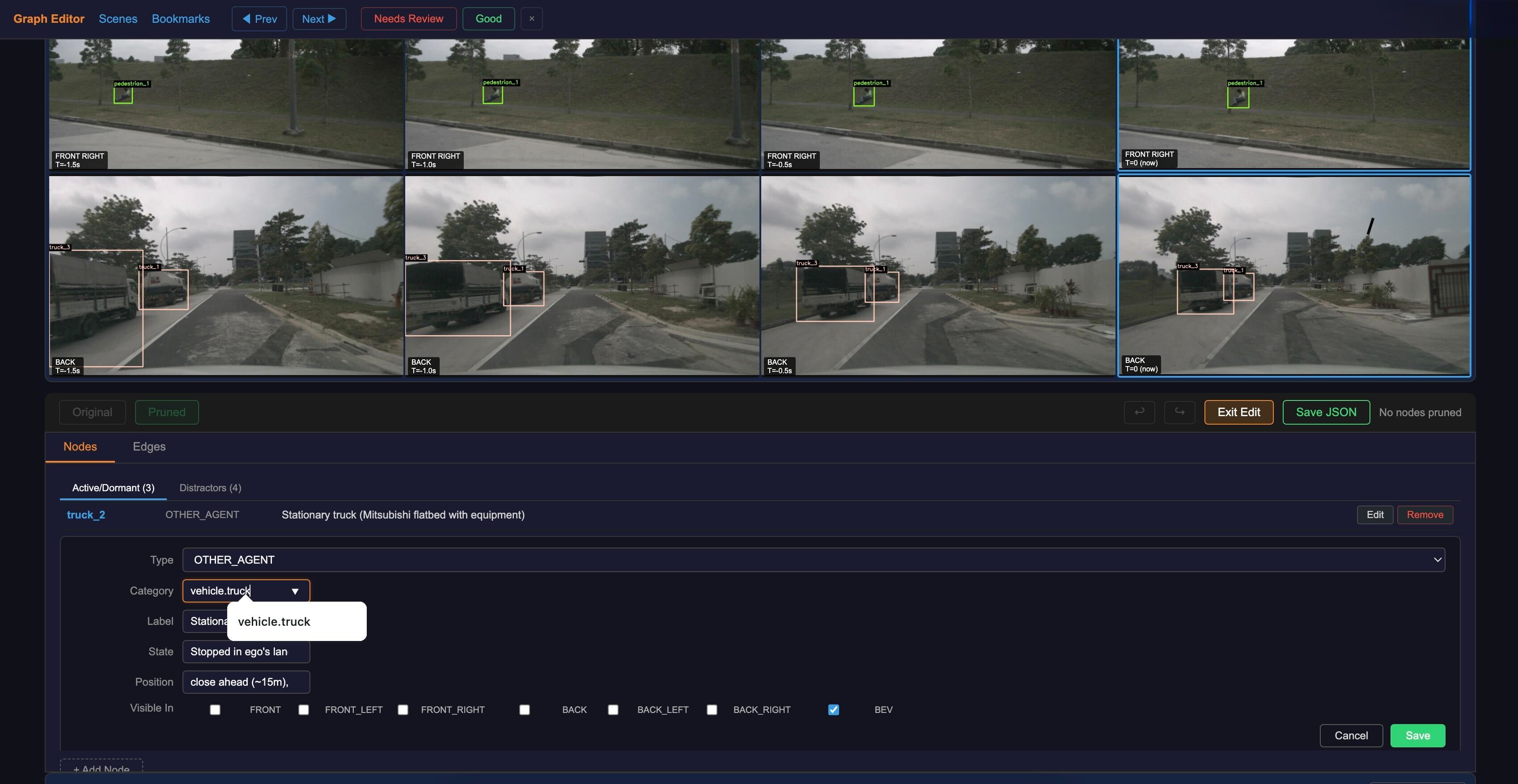}
    \caption{Per-sample editing interface in Edit Mode. Reviewers can change a node's Type, Category, Label, State, Position, and per-camera visibility, or remove it entirely; node-level edits are tracked through an undo / redo stack and committed as a versioned \texttt{graph.json}. The \textsc{Original} / \textsc{Pruned} toggle compares the extractor's raw output against the post-edit graph, the Active/Dormant and Distractors sub-tabs mirror the causal-status partition of \cref{app:causal_status}, and the \textsc{Needs Review} / \textsc{Good} buttons in the top bar set the sample's dashboard-level status.}
    \label{fig:edit_screen}
\end{figure}

\begin{figure}[h!]
    \centering
    \includegraphics[width=\textwidth]{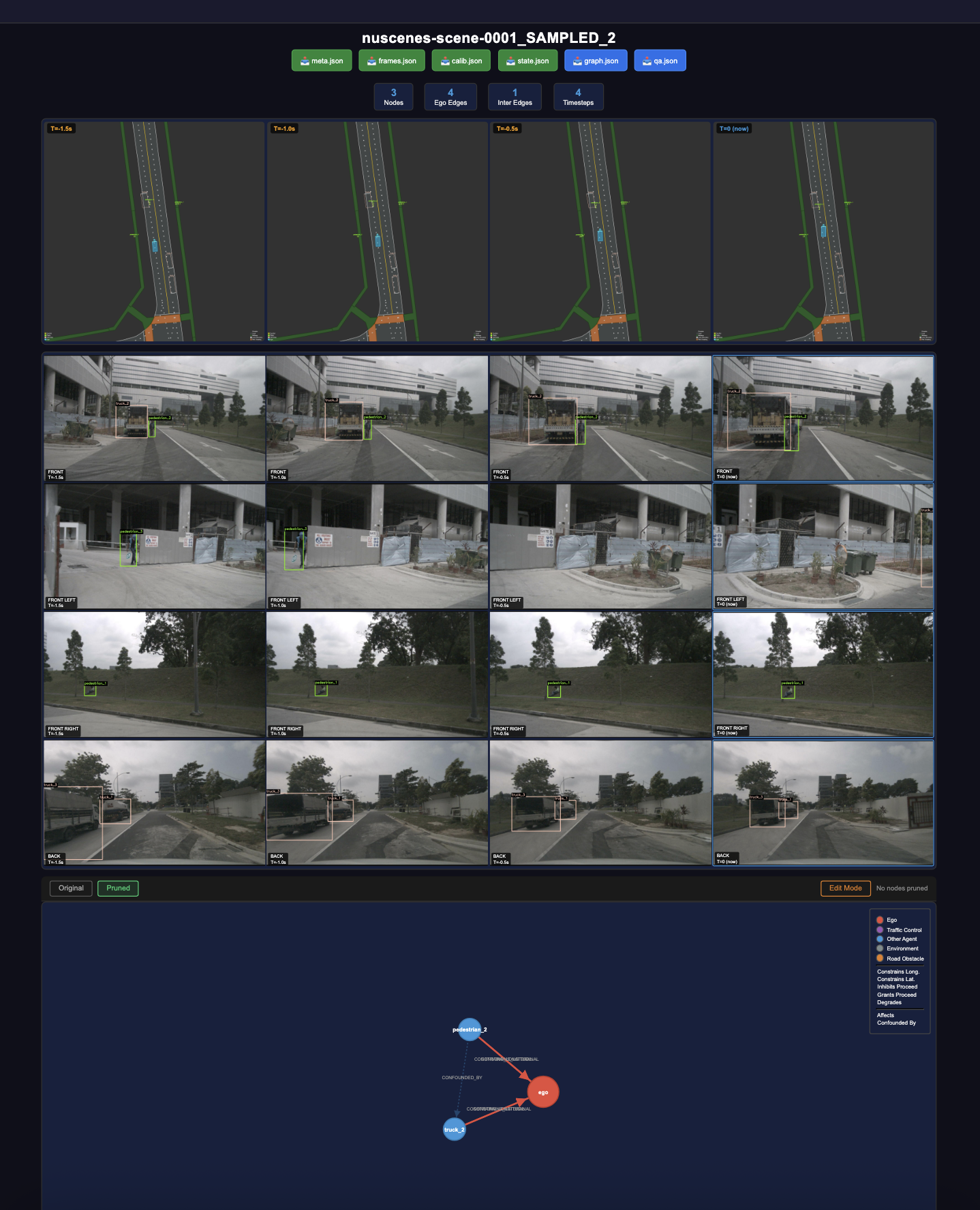}
    \caption{Per-sample read-only inspection view. The header lists the six JSON artefacts (\texttt{meta}, \texttt{frames}, \texttt{calib}, \texttt{state}, \texttt{graph}, \texttt{qa}) and node, ego-edge, inter-edge, and timestep counts for the sample. The middle panel renders the four-timestep BEV stack (top row) and the four ego-centric camera views (FRONT, FRONT\_LEFT, FRONT\_RIGHT, BACK) with all node bounding boxes overlaid. The bottom panel renders the extracted causal scene graph: solid arrows are directed ego edges colour-coded by effect type, and dashed bidirected arrows are CONFOUNDED\_BY links between non-ego nodes encoding shared unobserved causes.}
    \label{fig:sample_scene}
\end{figure}

Stage 2b (\cref{subsec:pipeline}) of the construction pipeline routes every extracted scene graph through an independent human reviewer before downstream QA generation. To make review tractable across the 3,314 samples drawn from the 850 nuScenes scenes, we built a purpose-built web tool that surfaces the extracted graph alongside the multi-view camera frames and BEV rasterisation that grounded it. The tool exposes three views described below: a dataset-wide progress dashboard (\cref{fig:edit_dashboard}), a per-sample editing screen (\cref{fig:edit_screen}), and a per-sample inspection page (\cref{fig:sample_scene}).

\paragraph{Dataset-wide progress tracking.} The top-level dashboard (\cref{fig:edit_dashboard}) lists every scene in the dataset together with its review status. Each row corresponds to a single nuScenes scene and reports the number of samples extracted from it (typically four, anchored at evenly spaced timesteps), the number of agent nodes the extractor inferred, and the per-sample QA, counterfactual-trajectory, and alternative-trajectory counts that populate once review completes. The View column shows one coloured chip per sample slot (S0--S3) whose colour encodes review state: green for samples confirmed Good, amber for samples flagged Needs Review or still in progress, and a greyed-out chip for samples that were filtered during the empty-scene pre-pass. The aggregate stat cards at the top of the page summarise scene, sample, QA, and trajectory counts across the entire dataset, while the per-row download icon allows reviewers to export an individual scene's reviewed graph for offline inspection.

\paragraph{Per-sample inspection.} Clicking a sample chip from the dashboard opens the read-only inspection view shown in Figure~\ref{fig:sample_scene}. The header surfaces the six JSON artefacts produced for each sample (\texttt{meta.json}, \texttt{frames.json}, \texttt{calib.json}, \texttt{state.json}, \texttt{graph.json}, \texttt{qa.json}) as separate downloads, alongside per-sample summary statistics (node count, ego-edge count, inter-node-edge count, and timestep count). The middle panel renders the four-timestep BEV rasterisation produced from the HD map and ego footprint (top row), followed by the four ego-centric camera views (FRONT, FRONT\_LEFT, FRONT\_RIGHT, BACK) at each of the four timesteps, with every node's 2D bounding box overlaid in its category-specific colour. The bottom panel renders the causal scene graph itself: solid coloured arrows are directed ego edges drawn in the legend's effect-type palette, and the bidirected dashed arrow encodes a CONFOUNDED\_BY link between two non-ego nodes that share an unobserved common cause inferred during extraction. The \textsc{Original} / \textsc{Pruned} toggle at the top of the panel flips between the extractor's raw output and the post-review graph, allowing the reviewer to audit which nodes were removed or modified.

\paragraph{Per-node graph editing.} Activating Edit Mode from the inspection view opens the editor shown in Figure~\ref{fig:edit_screen}. The upper panel displays the same multi-view camera stack with all node bounding boxes overlaid. The lower panel exposes the graph as two tabs, \textsc{Nodes} (further split into Active/Dormant and Distractors sub-tabs to mirror the two causal-status partitions of \cref{app:causal_status}) and \textsc{Edges}. Selecting a node opens an inline editor that lets the reviewer change its semantic Type (e.g.\ OTHER\_AGENT, TRAFFIC\_CONTROL, ROAD\_OBSTACLE), Category (e.g.\ \texttt{vehicle.truck}), free-text Label and State, position phrase, and the set of camera views in which the entity is visible. Each edit is captured as a discrete action that can be undone or redone via the toolbar, and the \textsc{Save JSON} button commits the result to the per-sample \texttt{graph.json} artefact. The header bar additionally exposes Prev / Next navigation across the scene's sample queue, and a pair of \textsc{Needs Review} / \textsc{Good} buttons that set the sample-level review status visible from the dashboard.

\section{Evaluation Prompts}
\label{app:Inference Prompts}

\lstdefinestyle{promptbox}{
  basicstyle=\fontsize{5.5pt}{6.5pt}\selectfont\ttfamily,
  breaklines=true,
  breakatwhitespace=false,
  columns=fullflexible,
  keepspaces=true,
  showstringspaces=false,
  frame=single,
  framerule=0.6pt,
  rulecolor=\color{black!70},
  backgroundcolor=\color{black!4},
  framesep=3pt,
  xleftmargin=4pt,
  xrightmargin=4pt,
  aboveskip=4pt,
  belowskip=4pt,
}

\label{sec:prompts}

This section reproduces the verbatim text payloads sent to each evaluated model across the four task types: VQA, baseline trajectory error (BTE), counterfactual trajectory error at Rung~2 (CTE-R2, intervention), and counterfactual trajectory error at Rung~3 (CTE-R3, counterfactual). All examples are drawn from \texttt{nuscenes-scene-0001} so the per-task framings are directly comparable across models. Both models receive the same 16 multi-camera frames (4 cameras $\times$ 4 timesteps from $T{=}{-}1.5$\,s to $T{=}0$); Alpamayo additionally receives \texttt{ego\_history\_xyz} and \texttt{ego\_history\_rot} tensors through its input transformer. Each model's system prompt (\cref{sec:prompts_alpamayo_sys,sec:prompts_cosmos_sys}) is sent as \texttt{role:~system} on every call.

\subsection{Alpamayo-1.5-10B}
\label{sec:prompts_alpamayo}

\subsubsection{System prompt}
\label{sec:prompts_alpamayo_sys}
Sent on every VQA and trajectory call.
\begin{lstlisting}[style=promptbox]
You are an expert autonomous driving assistant analyzing camera footage from a self-driving vehicle.
You are shown images from 4 cameras (front, front-left, front-right, back) across 4 consecutive timesteps (T-1.5s, T-1.0s, T-0.5s, T=0 current), plus the ego vehicle's trajectory history.
You will be asked causal reasoning questions about the driving scene. These questions cover:
- Dormant scene elements: objects that are visible but NOT currently affecting driving behavior
- Active causal elements: objects that ARE currently influencing driving decisions
- Counterfactual scenarios: what would happen if certain conditions changed
Ground your reasoning strictly in what you observe in the camera images and the vehicle's motion. Reason like an experienced driver explaining your thought process to a passenger.
For Binary (Yes/No) questions, respond with:  Answer: Yes / Reasoning: <your explanation in 2-4 sentences>
For Multiple Choice (A/B/C/D) questions, respond with:  Answer: <letter> / Reasoning: <your explanation in 3-5 sentences, including why the other options are wrong>
Use natural driving language. Do not reference internal graph terminology, coordinate values, or metric distances. Describe elements by their appearance and position relative to your vehicle.
\end{lstlisting}

\subsubsection{VQA}
\label{sec:prompts_alpamayo_vqa}
Question \texttt{CaI1}, sample \texttt{SAMPLED\_0}, MCQ format.
\begin{lstlisting}[style=promptbox]
Question: What is primarily allowing you to make this left turn right now?
A) The green traffic signal at the intersection, which permits me to proceed
B) The large tanker truck ahead in the opposing lane, which has left a gap for me to turn through
C) The blue car behind me, which is yielding to let me complete the turn
D) The unoccupied crosswalk on my turning path, which is giving me room to maneuver

Format: Answer: A, B, C, or D
\end{lstlisting}

\subsubsection{BTE}
\label{sec:prompts_alpamayo_bte}
Baseline trajectory prediction, sample \texttt{SAMPLED\_0}, 6 future waypoints over $[0.5, 3.0]$\,s.
\begin{lstlisting}[style=promptbox]
You are an autonomous driving system with access to multi-camera images from the past 1.5 seconds up to now (T=0).
Navigation command: TURN LEFT | Current ego speed: 3.99 m/s | Frame: (0, 0) = current position, +x = forward, +y = left
Task: Predict the ego vehicle's future trajectory waypoints.
Output timesteps (seconds): [0.5, 1.0, 1.5, 2.0, 2.5, 3.0]
Output ONLY the following JSON (replace x/y with your predictions):
{"waypoints": [{"t": 0.5, "x": 0.0, "y": 0.0}, {"t": 1.0, "x": 0.0, "y": 0.0}, {"t": 1.5, "x": 0.0, "y": 0.0}, {"t": 2.0, "x": 0.0, "y": 0.0}, {"t": 2.5, "x": 0.0, "y": 0.0}, {"t": 3.0, "x": 0.0, "y": 0.0}]}
\end{lstlisting}

\subsubsection{CTE-R2 (Intervention)}
\label{sec:prompts_alpamayo_cte_r2}
Question \texttt{traj\_CTE\_AB1}, sample \texttt{SAMPLED\_0}, future trajectory under a Rung-2 intervention. Emits the answer plus 6 future waypoints over $[0.5, 3.0]$\,s.
\begin{lstlisting}[style=promptbox]
You are an autonomous driving system with access to multi-camera images from the past 1.5 seconds up to now (T=0).
Navigation command: TURN LEFT | Current ego speed: 3.99 m/s | Frame: (0, 0) = current position, +x = forward, +y = left
=== Causal Reasoning Question ===
Suppose a person stepped off the far curb onto that crosswalk just as you reach this point in the turn. At your current speed and distance from the crossing, would you need to brake or adjust your steering?
Counterfactual scenario: A pedestrian steps off the far curb onto the crosswalk on ego's left-turn path, creating an occupied crosswalk that ego must yield to
Task 1 -- Answer the question (Yes or No).
Task 2 -- Given your counterfactual reasoning, predict the ego vehicle's trajectory under this scenario at timesteps: [0.5, 1.0, 1.5, 2.0, 2.5, 3.0] seconds.
Output ONLY the following JSON (fill in answer and x/y values):
{"answer": "<Yes or No>", "waypoints": [{"t": 0.5, "x": 0.0, "y": 0.0}, {"t": 1.0, "x": 0.0, "y": 0.0}, {"t": 1.5, "x": 0.0, "y": 0.0}, {"t": 2.0, "x": 0.0, "y": 0.0}, {"t": 2.5, "x": 0.0, "y": 0.0}, {"t": 3.0, "x": 0.0, "y": 0.0}]}
\end{lstlisting}

\subsubsection{CTE-R3 (Counterfactual)}
\label{sec:prompts_alpamayo_cte_r3}
Question \texttt{traj\_CTE\_WC1}, sample \texttt{SAMPLED\_1}, counterfactual-past trajectory plus answer, 3 waypoints over $[0.5, 1.5]$\,s.
\begin{lstlisting}[style=promptbox]
You are an autonomous driving system with access to multi-camera images from the past 1.5 seconds up to now (T=0).
Navigation command: GO STRAIGHT | Current ego speed: 5.86 m/s | Frame: (0, 0) = current position, +x = forward, +y = left
=== Causal Reasoning Question ===
You're cruising straight at moderate speed right now. If that parked truck ahead on the other side of the road had started drifting toward the center line by about half a lane, would you have been able to keep going without adjusting your steering?
Counterfactual scenario: Stationary canvas-covered truck in opposing lane drifts half a lane toward the center line, partially entering ego's lane space
Task 1 -- Answer the question (Yes or No).
Task 2 -- Given your counterfactual reasoning, predict the ego vehicle's trajectory under this scenario at timesteps: [0.5, 1.0, 1.5] seconds.
Output ONLY the following JSON (fill in answer and x/y values):
{"answer": "<Yes or No>", "waypoints": [{"t": 0.5, "x": 0.0, "y": 0.0}, {"t": 1.0, "x": 0.0, "y": 0.0}, {"t": 1.5, "x": 0.0, "y": 0.0}]}
\end{lstlisting}

\subsection{Cosmos-Reason-2 8B}
\label{sec:prompts_cosmos}

\subsubsection{System prompt}
\label{sec:prompts_cosmos_sys}
Sent on every VQA and trajectory call.
\begin{lstlisting}[style=promptbox]
You are an expert autonomous driving assistant with advanced spatial and temporal reasoning capabilities.
You are shown camera images from a self-driving vehicle across multiple timesteps and camera views (past to present). Use all provided images to reason carefully about the driving scene, the behavior of nearby agents, and causal relationships between events.
For Binary (Yes/No) questions: you MUST respond with 'Answer: Yes' or 'Answer: No' followed by a detailed explanation of your reasoning. Do not give the answer alone -- always explain why.
For Multiple Choice (A/B/C/D) questions: you MUST respond with 'Answer: <letter>' followed by a detailed explanation of your reasoning. Do not give the answer alone -- always explain why.
Always ground your reasoning in specific observations from the camera images.
\end{lstlisting}

\subsubsection{VQA}
\label{sec:prompts_cosmos_vqa}
Question \texttt{CaI1}, sample \texttt{SAMPLED\_0}, MCQ format.
\begin{lstlisting}[style=promptbox]
Question: What is primarily allowing you to make this left turn right now?
A) The green traffic signal at the intersection, which permits me to proceed
B) The large tanker truck ahead in the opposing lane, which has left a gap for me to turn through
C) The blue car behind me, which is yielding to let me complete the turn
D) The unoccupied crosswalk on my turning path, which is giving me room to maneuver

Format: Answer: A, B, C, or D
\end{lstlisting}

\subsubsection{BTE}
\label{sec:prompts_cosmos_bte}
Baseline trajectory prediction, sample \texttt{SAMPLED\_0}, 6 future waypoints over $[0.5, 3.0]$\,s.
\begin{lstlisting}[style=promptbox]
Navigation command: TURN LEFT | Current ego speed: 3.99 m/s | Frame: (0,0) = current position, +x = forward, +y = left
Task: Predict the ego vehicle's cumulative position at timesteps [0.5, 1.0, 1.5, 2.0, 2.5, 3.0] seconds.
- x = cumulative forward distance in metres (x >= 0, vehicle cannot go backward)
- y = lateral offset (positive = left, negative = right)
- Adjust for any braking, acceleration, or turning visible in the images.
Respond with ONLY a JSON object. No code blocks, no explanation, no extra fields:
{"waypoints": [{"t": 0.5, "x": float, "y": float}, {"t": 1.0, "x": float, "y": float}, {"t": 1.5, "x": float, "y": float}, {"t": 2.0, "x": float, "y": float}, {"t": 2.5, "x": float, "y": float}, {"t": 3.0, "x": float, "y": float}]}
\end{lstlisting}

\subsubsection{CTE-R2 (Intervention)}
\label{sec:prompts_cosmos_cte_r2}
Question \texttt{traj\_CTE\_AB1}, sample \texttt{SAMPLED\_0}, future trajectory under a Rung-2 intervention, 6 future waypoints over $[0.5, 3.0]$\,s.
\begin{lstlisting}[style=promptbox]
Navigation command: TURN LEFT | Current ego speed: 3.99 m/s | Frame: (0,0) = current position, +x = forward, +y = left
Suppose a person stepped off the far curb onto that crosswalk just as you reach this point in the turn. At your current speed and distance from the crossing, would you need to brake or adjust your steering?
Counterfactual scenario: A pedestrian steps off the far curb onto the crosswalk on ego's left-turn path, creating an occupied crosswalk that ego must yield to
Task 1 -- Answer the question (Yes or No).
Task 2 -- Predict the ego vehicle's trajectory at timesteps [0.5, 1.0, 1.5, 2.0, 2.5, 3.0] seconds under this counterfactual scenario.
- x = cumulative forward distance in metres (x >= 0, vehicle cannot go backward)
- y = lateral offset (positive = left, negative = right)
Respond with ONLY a JSON object. No code blocks, no explanation, no extra fields:
{"answer": "<Yes or No>", "waypoints": [{"t": 0.5, "x": float, "y": float}, {"t": 1.0, "x": float, "y": float}, {"t": 1.5, "x": float, "y": float}, {"t": 2.0, "x": float, "y": float}, {"t": 2.5, "x": float, "y": float}, {"t": 3.0, "x": float, "y": float}]}
\end{lstlisting}

\subsubsection{CTE-R3 (Counterfactual)}
\label{sec:prompts_cosmos_cte_r3}
Question \texttt{traj\_CTE\_WC1}, sample \texttt{SAMPLED\_1}, counterfactual-past trajectory anchored at $T{=}{-}1.5$\,s, 3 waypoints over $[0.5, 1.5]$\,s.
\begin{lstlisting}[style=promptbox]
Navigation command: GO STRAIGHT | Current ego speed: 5.86 m/s | Frame: (0,0) = current position, +x = forward, +y = left
You're cruising straight at moderate speed right now. If that parked truck ahead on the other side of the road had started drifting toward the center line by about half a lane, would you have been able to keep going without adjusting your steering?
Counterfactual scenario: Stationary canvas-covered truck in opposing lane drifts half a lane toward the center line, partially entering ego's lane space
Task 1 -- Answer the question (Yes or No).
Task 2 -- Predict the ego vehicle's trajectory at timesteps [0.5, 1.0, 1.5] seconds under this counterfactual scenario.
- x = cumulative forward distance in metres (x >= 0, vehicle cannot go backward)
- y = lateral offset (positive = left, negative = right)
Respond with ONLY a JSON object. No code blocks, no explanation, no extra fields:
{"answer": "<Yes or No>", "waypoints": [{"t": 0.5, "x": float, "y": float}, {"t": 1.0, "x": float, "y": float}, {"t": 1.5, "x": float, "y": float}]}
\end{lstlisting}
\vspace{-2ex}


\end{document}